\documentclass[10pt,twocolumn,letterpaper]{article}
\usepackage{cvpr}              
\usepackage{multirow, mathtools} 
\usepackage{verbatim}
\usepackage{pifont}

\definecolor{cvprblue}{rgb}{0.21,0.49,0.74}
\usepackage[pagebackref,breaklinks,colorlinks,allcolors=cvprblue]{hyperref}

\def\confName{CVPR}
\def\confYear{2025}

\title{ID-Patch: Robust ID Association for Group Photo Personalization}

\author{
  Yimeng Zhang$^{1,2,*}$ , Tiancheng Zhi$^{1}$, Jing Liu$^{1}$, Shen Sang$^{1}$,  \\ Liming Jiang$^{1}$, Qing Yan$^{1}$, Sijia Liu$^{2}$, Linjie Luo$^{1}$
 \\
 \\
  $^{1}$ ByteDance Inc., $^{2}$ Michigan State University
  }

\begin{document}

\twocolumn[{%
\renewcommand\twocolumn[1][]{#1}%
\maketitle
\begin{center}
    \centering
        \captionsetup{type=figure}
\includegraphics[scale=0.255]{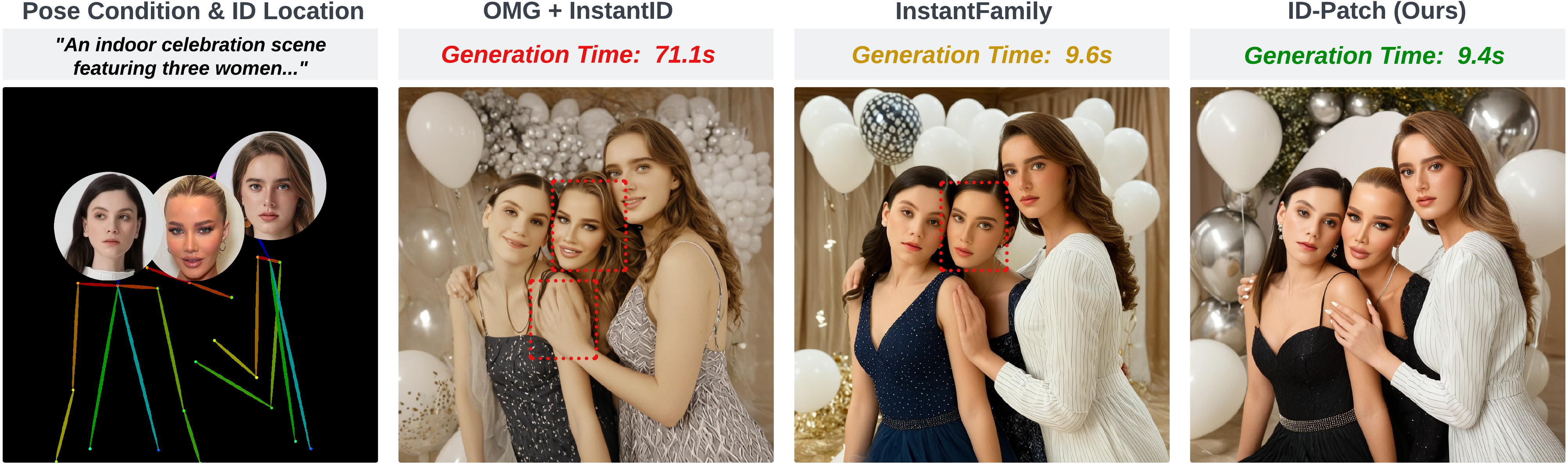}
        \caption{
    Comparison with state-of-the-art multi-identity generation methods. From left to right: the condition inputs followed by results generated using OMG~\cite{kong2024omg} (with InstantID~\cite{wang2024instantid}), InstantFamily~\cite{kim2024instantfamily}, and our proposed ID-Patch approach. Red dashed boxes highlight failures. OMG fails to preserve the hairstyles of the middle person, and creates artifacts for the right woman's hand, possibly because of the inconsistency between its generation stages  (explained in Fig.~\ref{fig:omg_fail}). InstantFamily suffers from ID leakage, resulting the incorrect ID of the middle person. Our approach preserves the detailed identity of each person. In addition, our approach is 7 times faster than OMG and has less computational overhead than InstantFamily.
         }
        \label{fig:teaser}
\end{center}%
}]
\footnotetext{$^{*}$Work done during internship at ByteDance.}

\begin{abstract}

The ability to synthesize personalized group photos and specify the positions of each identity offers immense creative potential. While such imagery can be visually appealing, it presents significant challenges for existing technologies. A persistent issue is identity (ID) leakage, where injected facial features interfere with one another, resulting in low face resemblance, incorrect positioning, and visual artifacts. Existing methods suffer from limitations such as the reliance on segmentation models, increased runtime, or a high probability of ID leakage. To address these challenges, we propose ID-Patch, a novel method that provides robust association between identities and 2D positions. Our approach generates an ID patch and ID embeddings from the same facial features: the ID patch is positioned on the conditional image for precise spatial control, while the ID embeddings integrate with text embeddings to ensure high resemblance. Experimental results demonstrate that ID-Patch surpasses baseline methods across metrics, such as face ID resemblance, ID-position association accuracy, and generation efficiency.
Project Page is: \url{https://byteaigc.github.io/ID-Patch/}

\end{abstract}    
\section{Introduction}
\label{sec:intro}

Personalized group photo generation has wide-ranging applications in entertainment, advertising, social media, and virtual reality. For example, users might want to create a group photo of friends at a virtual party, even if some could not attend in person. This technology enables the creation of engaging and memorable content that reflects real-life social interactions and relationships. However, this task poses significant challenges, particularly the risk of identity (ID) leakage, where the identity information of one individual inadvertently affects the representation of another at a different position. Such blending compromises the distinctiveness and accuracy of each person’s features. Besides, being able to specify the position of each identity in the image is an important function when users want to create a specific layout of people. Incorrect positioning of identities may lead to undesired results.

\begin{figure}[!t]
\begin{center}
\begin{subfigure}[b]{0.32\linewidth}
\includegraphics[width=\linewidth]{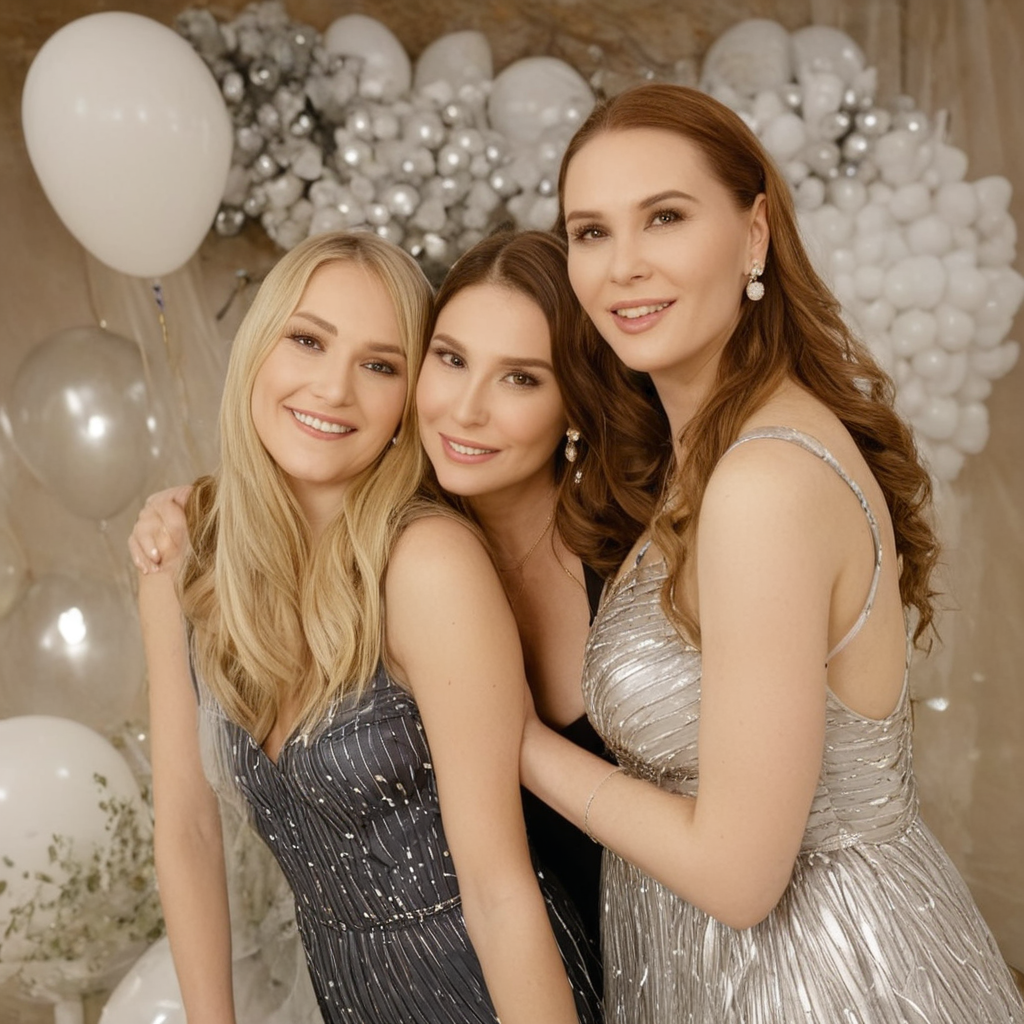} %
\caption{OMG first stage}
\end{subfigure}
\begin{subfigure}[b]{0.32\linewidth}
\includegraphics[width=\linewidth]{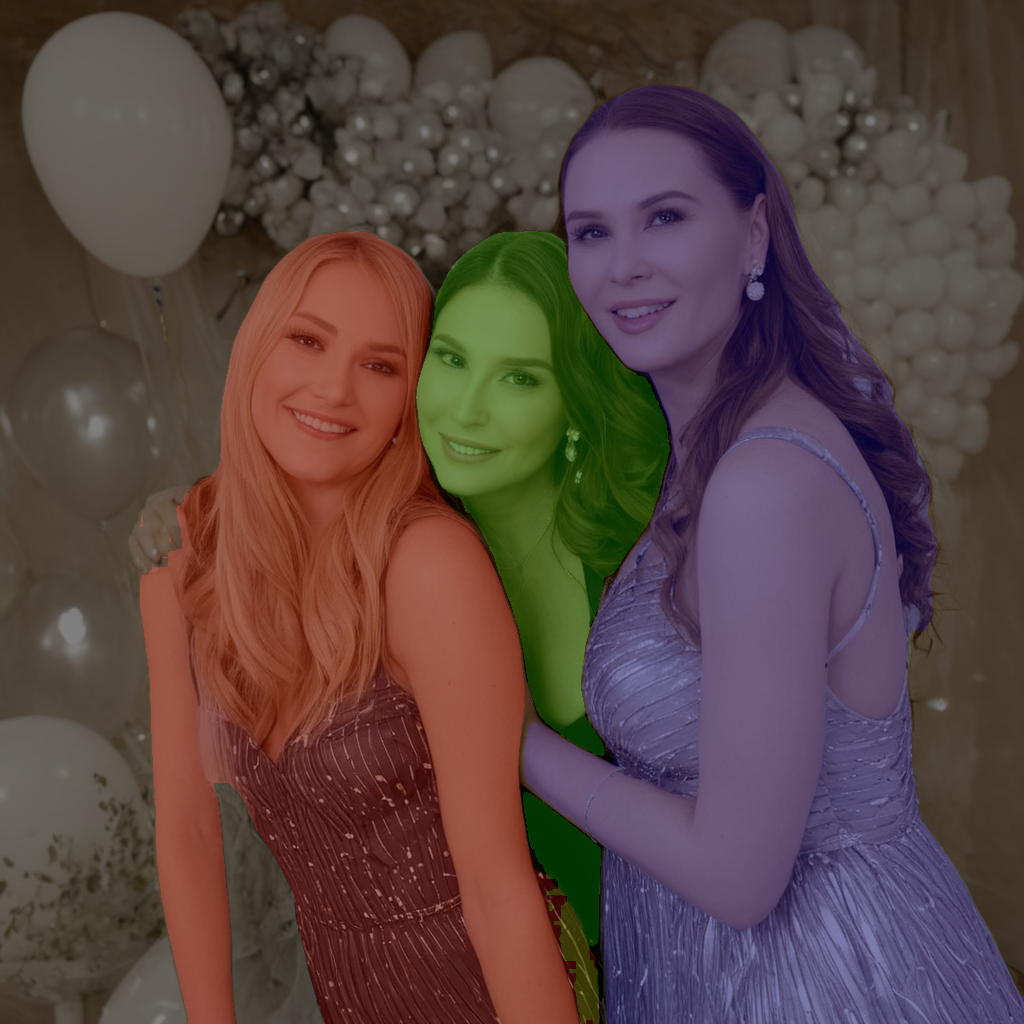} %
\caption{Human mask}
\end{subfigure}
\begin{subfigure}[b]{0.32\linewidth}
\includegraphics[width=\linewidth]{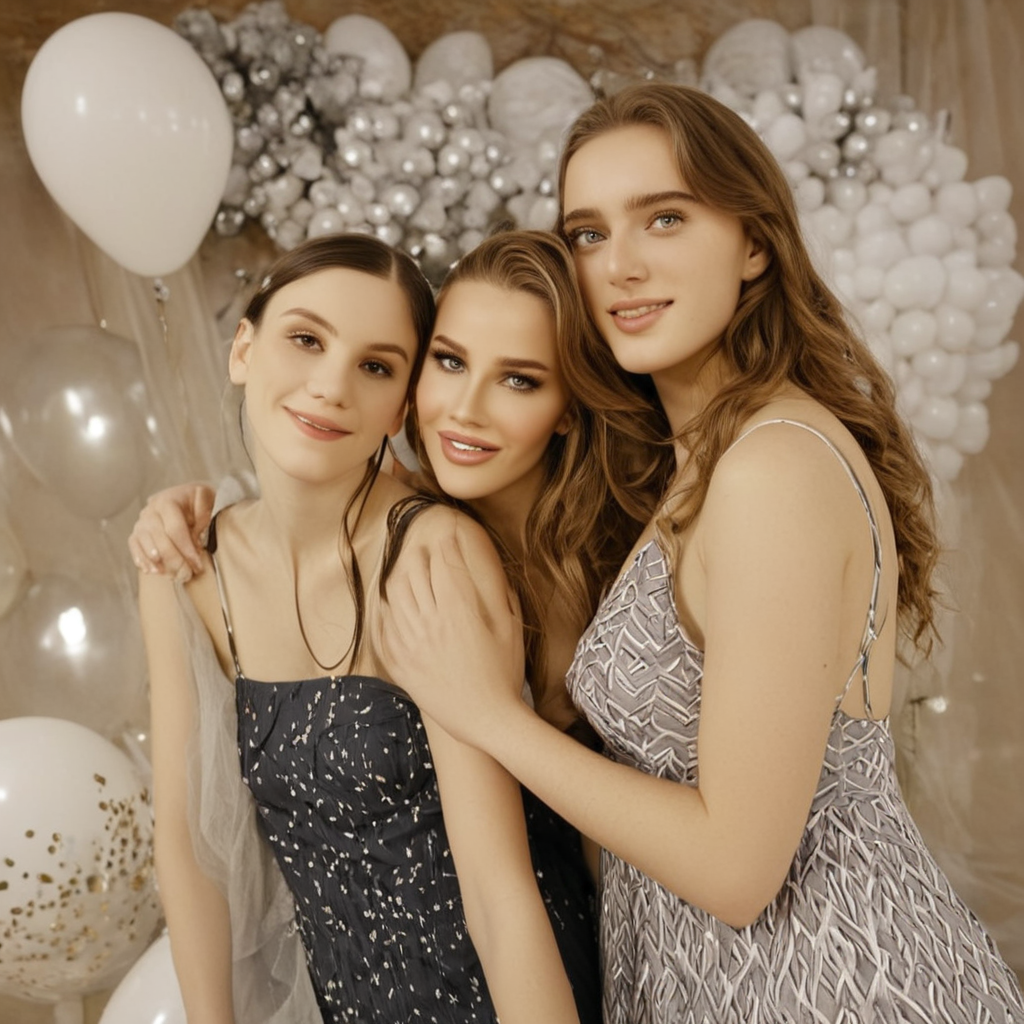} %
\caption{OMG final result}
\end{subfigure}
\vspace*{-3mm}
\caption{Why does OMG fail in Fig.\,\ref{fig:teaser}? OMG relies on a first stage result (no ID awareness) and its human segmentation to control layout. The second stage paints injected IDs within corresponding masks. However, the second pass may conflict with the first stage semantics, leading to deteriorated ID resemblance (long hair in the middle of image (c)) or artifacts (hand in the middle).}
\label{fig:omg_fail}
\end{center}
\vspace*{-5mm}
\end{figure}

\begin{figure}[!t]
\begin{center}
\begin{subfigure}[b]{0.32\linewidth}
\includegraphics[width=\linewidth]{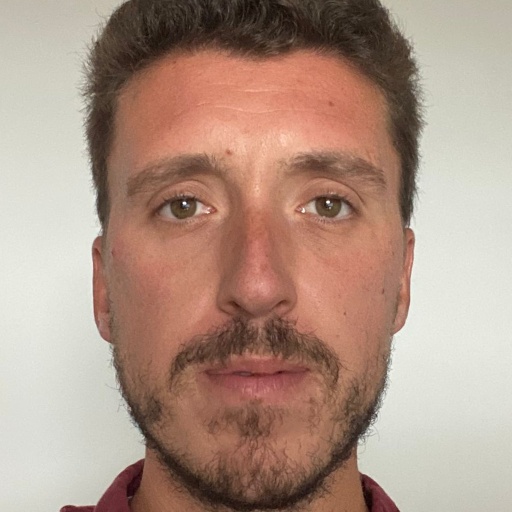} %
\caption{ID}
\end{subfigure}
\begin{subfigure}[b]{0.32\linewidth}
\includegraphics[width=\linewidth]{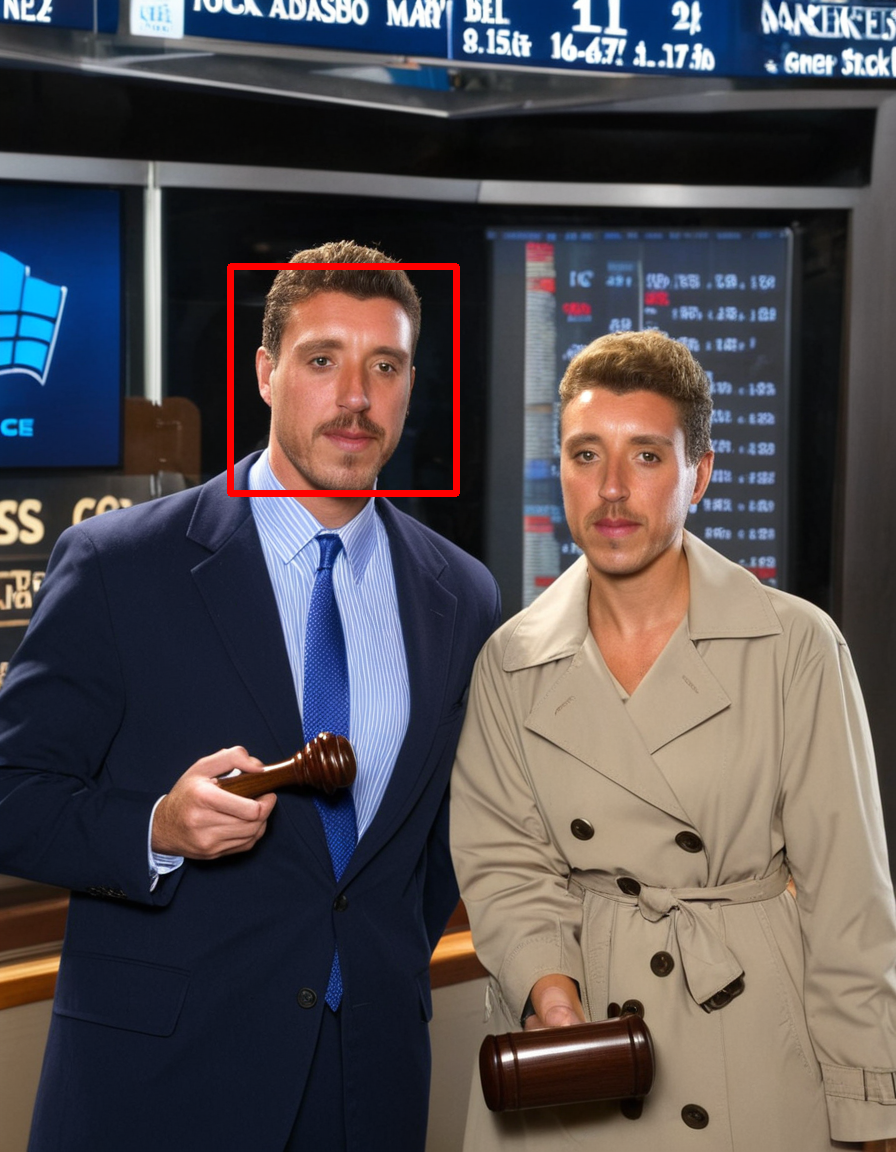} %
\caption{InstantFamily result}
\end{subfigure}
\begin{subfigure}[b]{0.32\linewidth}
\includegraphics[width=\linewidth]{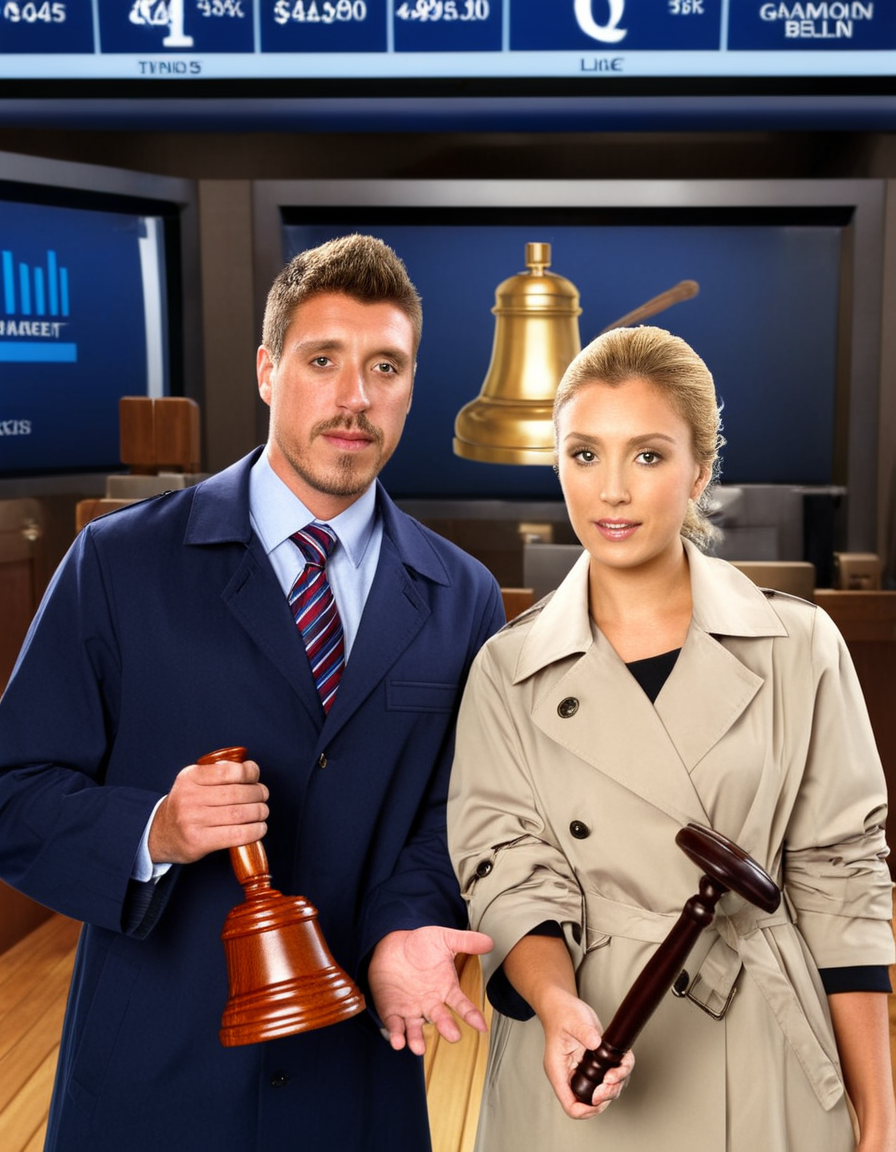} %
\caption{ID-Patch result}
\end{subfigure}
\vspace*{-3mm}
\caption{ID leakage of InstantFamily~\cite{kim2024instantfamily}. The text prompt describes a man and a woman standing together, and ID (a) is only injected into the left man. }

\label{fig:leak}
\end{center}
\vspace*{-5mm}
\end{figure}

State-of-the-art group photo personalization methods largely rely on diffusion models \cite{sohl2015deep, song2020denoising, ho2020denoising, dhariwal2021diffusion, balaji2022ediff, nichol2021glide,zhang2024defensive, wang2024edit, zhang2024unlearncanvas, sui2024disdet,xiao2024coap,pmlr-v202-xiao23e}, which deliver superior visual quality when paired with effective prompt engineering~\cite{esser2403scaling, saharia2022photorealistic, chen2023pixart, peebles2023scalable, ma2024sit,zhang2024generate}. To address the issue of ID leakage, OMG~\cite{kong2024omg} first generates an image without ID injection. Then, it estimates segmentation masks for the individuals in the image and subsequently injects the IDs into each segmented region. However, it depends on a segmentation model, which might fail in challenging cases. Besides, the content generated in the two stages may conflict with each other, leading to reduced visual fidelity. See \textbf{Fig.\,\ref{fig:teaser}} and \textbf{Fig.\,\ref{fig:omg_fail}} for example. Moreover, OMG requires running the denoising process separately for each identity, resulting in a runtime that increases linearly with the number of people. 
Alternatively, InstantFamily~\cite{kim2024instantfamily} employs a single-pass generation strategy by manipulating cross-attention masks. This approach uses approximate head region masks for each ID to direct pixel attention exclusively to specific ID embeddings derived from facial features. However, it struggles to prevent ID leakage due to two key issues: (1) imprecise masks or close proximity of faces, which can cause overlaps, and (2) unintended information propagation through self-attention and convolutional layers. Please see \textbf{Fig.\,\ref{fig:teaser}} and \textbf{Fig.\,\ref{fig:leak}} for more details.

In response to these limitations, we propose a novel method named ID-Patch, designed for efficient and robust ID-position association in group photo generation. Our idea is to let the model learn to associate input IDs with their spatially designated locations. Specifically, we project facial features onto a small RGB image patch called ID patch and token embeddings called ID embeddings. The ID patches are placed onto the conditioning image of a ControlNet~\cite{zhang2023adding} according to face locations to ensure precise position control of IDs within the generated images. ID embeddings are integrated with text embeddings to enhance facial details. Our ID-Patch method seamlessly integrates with various types of spatial conditions, such as poses, canny edges or depth map, enhancing the robustness and flexibility of our method.

\noindent We summarize our key \textbf{contributions} as follows: \begin{itemize} \item ID-Patch Method: We introduce a novel approach for multi-ID image generation that directly links face ID features with their locations using visual patches, ensuring accurate resemblance and position control. \item Efficiency Improvement: Our method enhances the efficiency of multi-ID image generation by simplifying the process to adding ID patches and concatenating ID embeddings, reducing computational overhead. \item Simplified Control: We eliminate the need for auxiliary segmentation models, requiring only a single point for ID position control, thereby increasing robustness and reducing complexity. \item Superior Performance: Our approach achieves higher fidelity in ID resemblance and better position control, especially in complex images with multiple identities, as demonstrated by our experimental results. \end{itemize}

\section{Related Work}
\label{sec:related}

\begin{figure*}[t]
    \centering
    \includegraphics[scale=0.52]{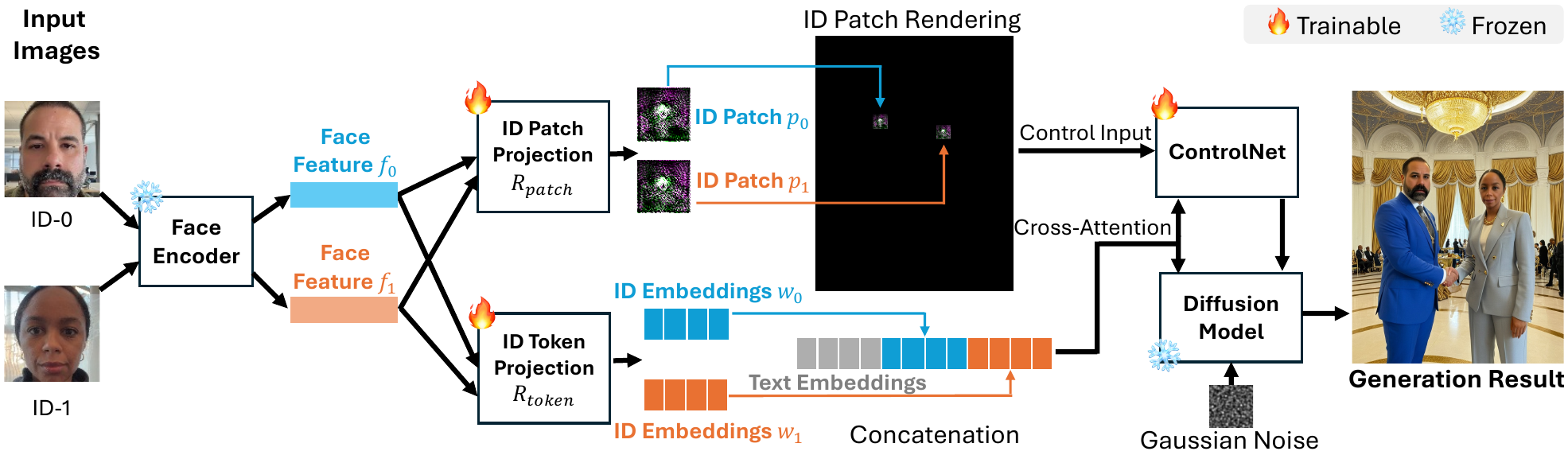}
    \vspace*{-3mm}
    \caption{Our pipeline. Given a text prompt (e.g., two people shaking hands), $N$ face images and locations, we generate an image with $N$ IDs. We extract face features for each ID and project them into ID patches and ID embeddings. ID patches are rendered on a black canvas (or added on top of a pose image) according to face locations and sent into a ControlNet to control the positions of generated faces. ID embeddings are appended to text embeddings to provide detailed face information to the diffusion model and ControlNet via cross-attention.}
    \label{fig:pipeline}
     \vspace*{-3mm}
\end{figure*}

\noindent\textbf{Single-ID Personalization.}
The personalization of diffusion models has emerged as a popular research topic. Fine-tuning or optimization based methods~\cite{gal2022image,wen2024hard,ruiz2023dreambooth,gal2023encoder,ruiz2024hyperdreambooth} have the potential to reach high quality results in terms of ID resemblance, but are time-consuming. Inference-only methods~\cite{wei2023elite,ye2023ip,li2024photomaker,wang2024instantid,he2024imagine,peng2024portraitbooth,zhou2023enhancing,yuan2023inserting} eliminate the finetuning process usually by encoding the concept into token embeddings and injecting them into the diffusion models via cross-attention. However, most of these works focus on single concept injection rather than group photo generation.

\noindent\textbf{Multi-ID Personalization.}
The main challenge of personalized group photo generation is ID blending, i.e., the identity information of person A may affect the generation of person B at another position.
To resolve this issue, OMG \cite{kong2024omg} performs separate denoising processes. First, a non-customized image is generated and masks of each person are extracted. Second, masked regions are painted using separate customized models and stitched together. This approach is slow and conflicts between stages may cause identity loss. Additionally, it relies on a segmentation module, which may fail in challenging cases. To avoid running separate customized models, recent works~\cite{xiao2024fastcomposer,wang2024moa,he2024uniportrait} let the network to learn to attend to specific concept tokens by supervising the attention mask using segmentation or ID routing. However, these methods do not allow the specification of the accurate locations of each ID. Although they can be combined with pose conditions, describing the assignment of IDs to positions in text may be hard. InstantFamily~\cite{kim2024instantfamily} provides an opportunity to specify the approximate location of each ID by manipulating the attention masks during both training and inference. However, no constraints have been imposed to avoid the ID leakage through self-attention and convolutional layers, leading to ID blending, omitting or swapping.

\noindent\textbf{Multiple General Concepts.}
Besides human generation, there are series of work focusing on the customization of multiple general concepts~\cite{kumari2023multi,avrahami2023break,tewel2023key,avrahami2023spatext,ding2024freecustom,gu2024mix,han2023svdiff,liu2023cones,xiao2024fastcomposer} and grounded image generation~\cite{li2023gligen,avrahami2023spatext,Wang_2024_CVPR} associating objects with positions. GLIGEN~\cite{li2023gligen} integrates grounding tokens, which align textual prompts with specific image regions. SpaText~\cite{avrahami2023spatext} and InstanceDiffusion~\cite{Wang_2024_CVPR} allows users to define the placement and attributes of objects. However, these concepts are usually semantically distinct while it is hard to describe the human identities in text and subtle details matter.

\section{Method}
\label{sec:method}
In this section, we first outline some preliminaries about text-to-image diffusion models and ControlNets. Subsequently, we detail the problem formulation for multi-ID image generation and the design of our modules, including the ID patch and ID embeddings, as well as the training and inference schemes.

\subsection{Preliminaries}
\label{ssec:prelim}

\noindent\textbf{Latent diffusion models (LDMs).}
LDMs \cite{rombach2022high} are image generation models that can incorporate conditioning signals into the generation process. Specifically, during training, an encoder $\mathcal{E}$ of variational autoencoder (VAE) \cite{kingma2013auto}  encodes an image $x$ into a latent space, and a neural network $\epsilon_\theta(z_t, t, c)$ performs the denoising process within this latent space. It attempts to predict the original Gaussian noise $\epsilon$ from a noisy latent variable $z_t$, where $t$ represents the diffusion timestep, and $c$ is the conditioning signal, injected via cross-attention. The loss function for an LDM is defined as:
\begin{equation}
    L_\mathrm{LDM} = \mathbb{E}_{\mathcal{E}(x),\epsilon \sim \mathcal{N}(0, 1),t}\left[\lVert\epsilon - \epsilon_\theta(z_t, t, c)\rVert_2^2\right]
\end{equation}

\noindent\textbf{ControlNet.} The incorporation of ControlNet \cite{zhang2023adding} into diffusion models marks a significant advancement in controlled image synthesis. ControlNet is designed to steer the image generation process by applying spatial conditioning signals such as rasterized pose images, depth maps, or canny edges. 
It introduces a control network that processes these signals and adds their features to the base model's intermediate results. The control network is a trainable copy of the base model's encoder layers and is initialized with zero convolution. This effective integration achieves high quality controlled generation. 

\subsection{Multi-ID Image Generation}
\label{sec:pd}
This paper focuses on efficiently and effectively addressing the problem of multi-ID image generation utilizing conditioning signals that include text prompts, identity features, and face locations. The inputs to our model comprise text prompt embeddings $c_t$, facial features $c_f = (f_0, f_1, \dots, f_{N-1})$ for $N$ distinct identities, and their corresponding facial locations $c_l = (l_0, l_1, \dots, l_{N-1})$. The facial location $l_i = (x_i, y_i)$ identifies the xy-coordinates of the nose tip for each identity within the intended image. The goal is to synthesize an image following the text description with each given face located in the designated place. 

\begin{figure}[t]
\begin{center}
\begin{subfigure}[b]{0.23\linewidth}
\includegraphics[width=\linewidth]{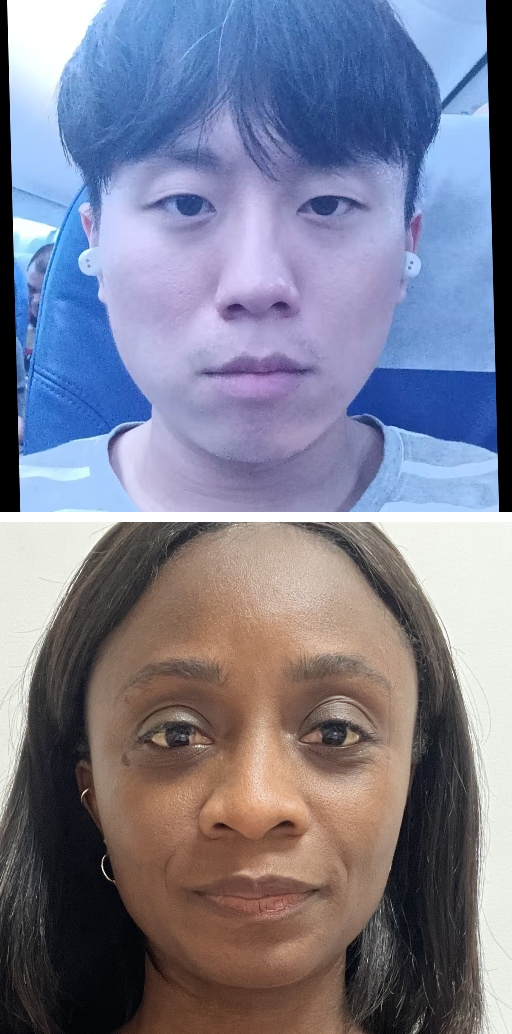} %
\caption{ID}
\end{subfigure}
\begin{subfigure}[b]{0.3625\linewidth}
\includegraphics[width=\linewidth]{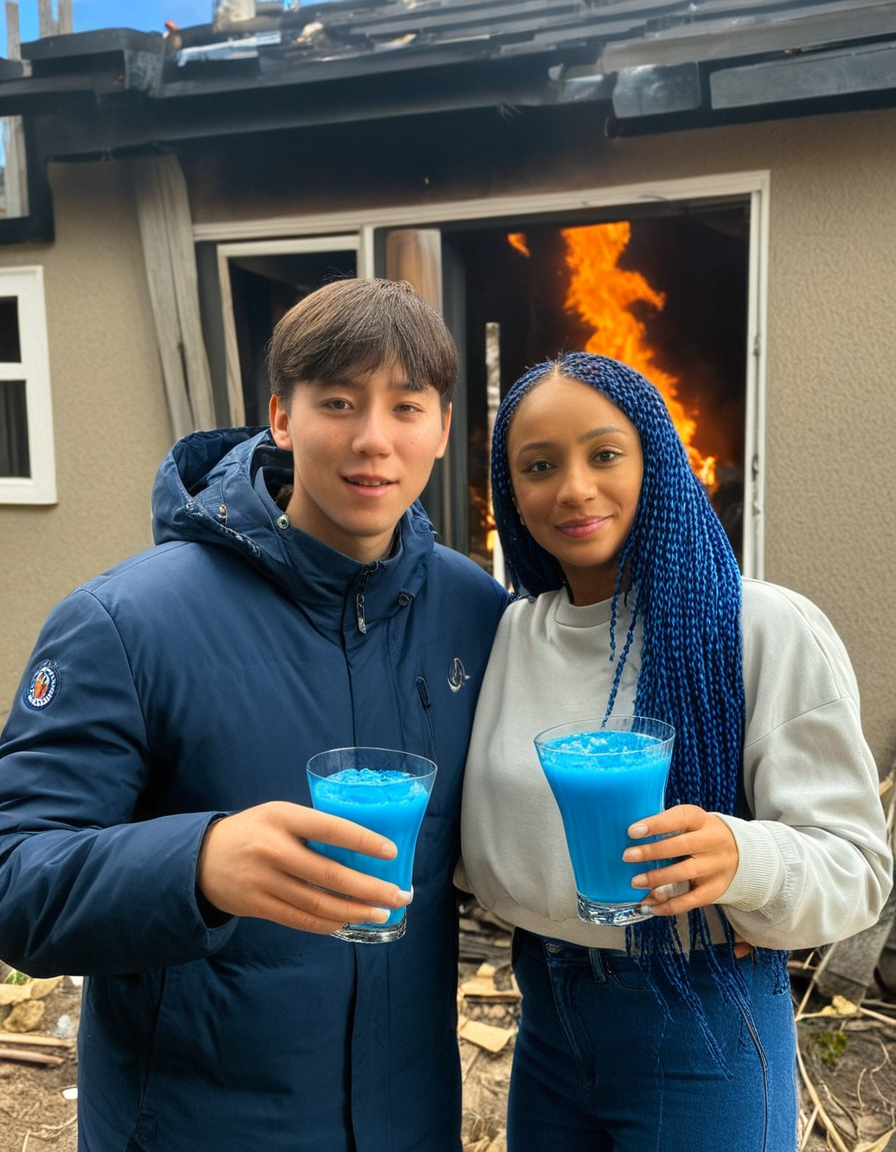} %
\caption{w/o ID embeddings}
\end{subfigure}
\begin{subfigure}[b]{0.3625\linewidth}
\includegraphics[width=\linewidth]{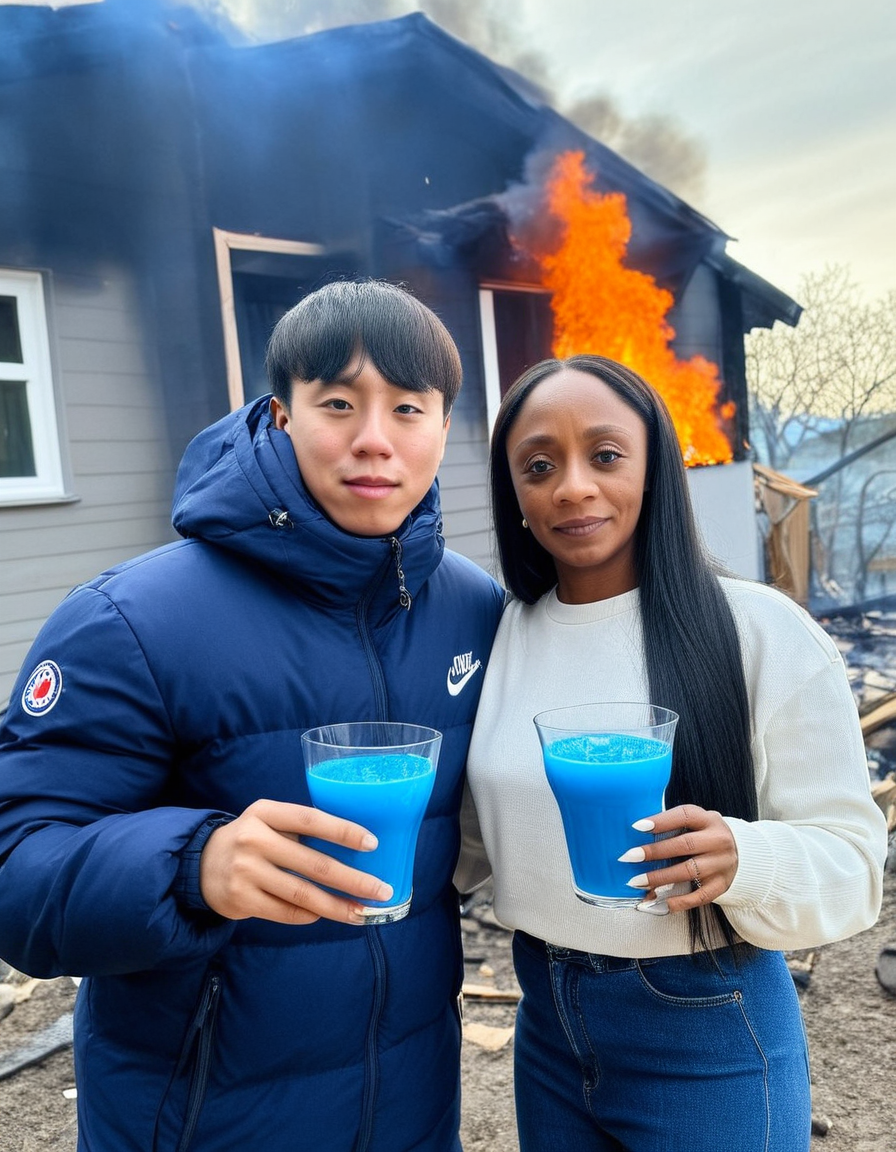} %
\caption{with ID embeddings}
\end{subfigure}
\caption{Effectiveness of ID embedding. Without ID embeddings, one can distinguish between the two people but the resemblance is low. Incorporating ID embeddings significantly improves face resemblance.}
\label{fig:id_token}
\end{center}
\vspace*{-5mm}
\end{figure}

\subsubsection{ID Patches and ID Embeddings}

As discussed in Sec.~\ref{ssec:prelim}, ControlNet effectively integrates spatial conditional signals into diffusion models. Building on this capability, we introduce a method to encode ID location information using condition images, subsequently processed by ControlNet to guide ID localization within the generated image. Our approach features ID conditions manifesting as both patches and embeddings. Specifically, ID patches, precisely aligned with facial locations, are integrated via ControlNet to ensure accurate identity localization, while ID embeddings are utilized via cross-attention mechanisms in the LDM to enhance the detailed representation of facial features, as illustrated in \textbf{Fig.\,\ref{fig:pipeline}}.

Both ID patches and ID embeddings are derived from facial features, enabling a synergistic association between them, with both projection processes facilitated by the Perceiver Resampler \cite{ye2023ip}. For ID patches, we convert the facial feature $f_i$ into an identity-specific patch $p_i \in \mathbb{R}^{P \times P \times 3}$, functioning similarly to a QR code, uniquely identifying each person and easily distinguishable from others. These ID patches are placed on a black canvas at specified facial locations to create an image $I$, which then serves as the conditioning input for ControlNet, thereby providing precise identity localization information.

As depicted in \textbf{Fig.\,\ref{fig:id_token}}, while ID patches alone are generally sufficient for distinguishing between identities, they may not always achieve optimal results in generating images with high ID resemblance. To address this limitation, and inspired by InstantFamily \cite{kim2024instantfamily}, we also incorporate ID embeddings. These embeddings are designed to inject detailed facial appearance information into the diffusion model, thus enhancing the resemblance of the generated identities to their real-life counterparts. Specifically, ID embeddings $w_i \in \mathbb{R}^{2048 \times M}$, share the same token dimension 2048 as the SDXL text embeddings. As shown in \textbf{Fig.\,\ref{fig:pipeline}}, these embeddings are appended to text embeddings $c_t$ to form an extended text embedding $c_t' = [c_t, w_0, w_1, \dots, w_{N-1}]$, which is then fed into both the UNet and ControlNet through cross-attention mechanisms.

\begin{figure}[t]
\begin{center}
\begin{subfigure}[b]{0.1\linewidth}
\includegraphics[width=\linewidth]{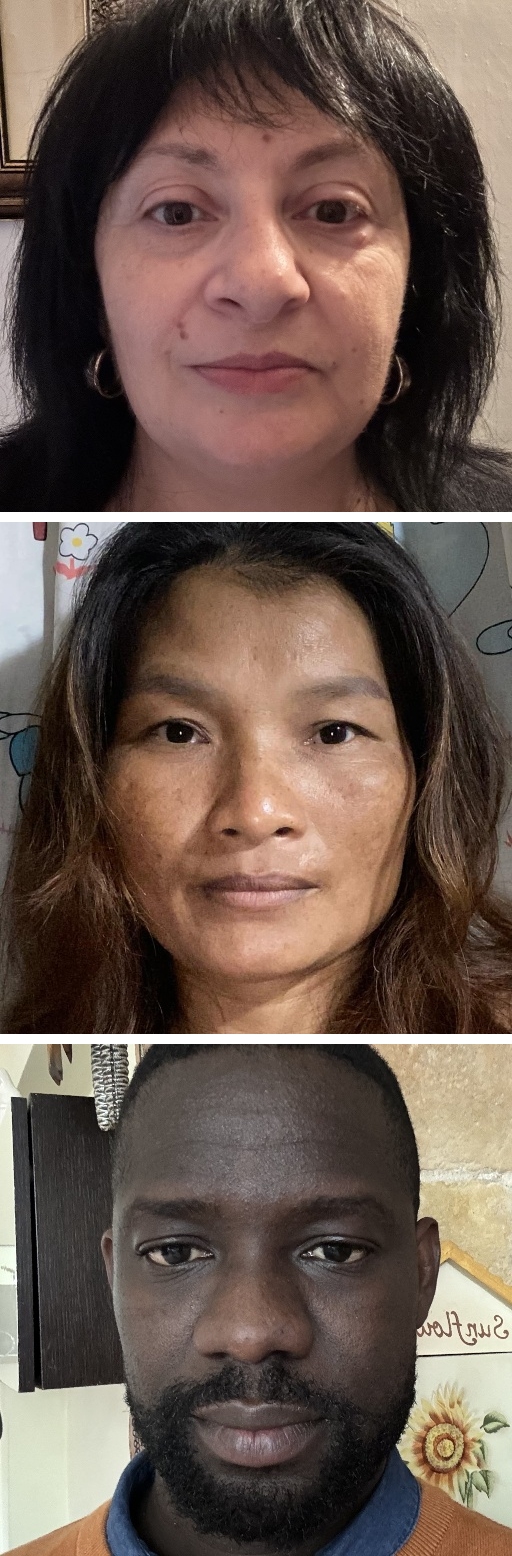} %
\caption{ID}
\end{subfigure}
\begin{subfigure}[b]{0.29\linewidth}
\includegraphics[width=\linewidth]{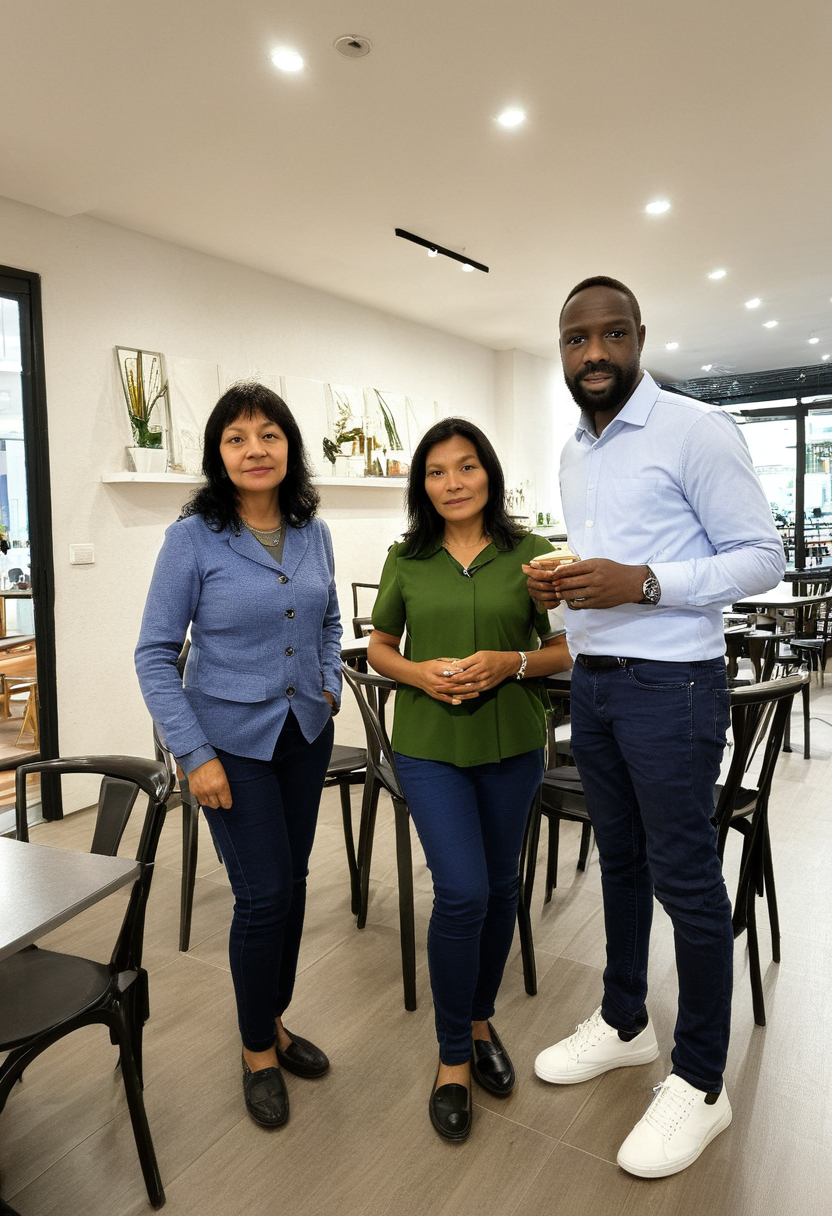} %
\caption{Result w/o pose}
\end{subfigure}
\begin{subfigure}[b]{0.29\linewidth}
\includegraphics[width=\linewidth]{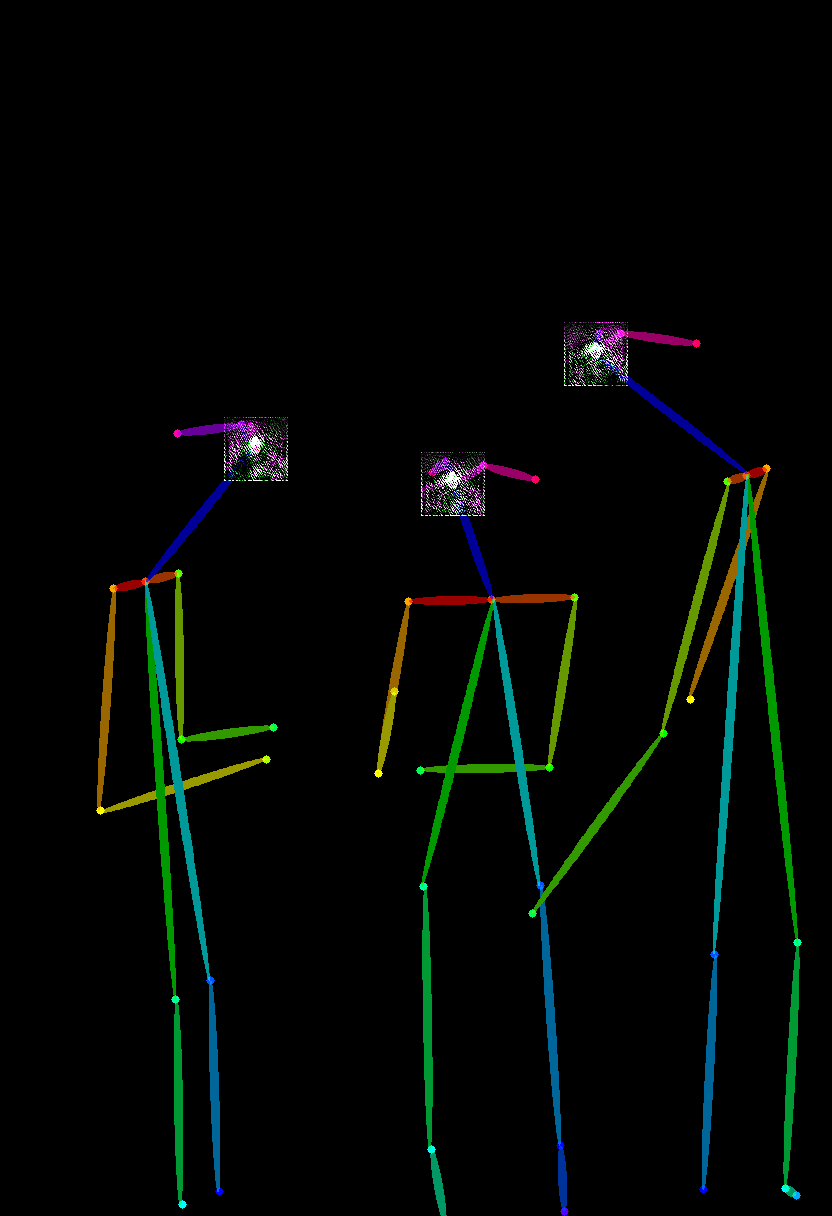} %
\caption{ID patch on pose}
\end{subfigure}
\begin{subfigure}[b]{0.29\linewidth}
\includegraphics[width=\linewidth]{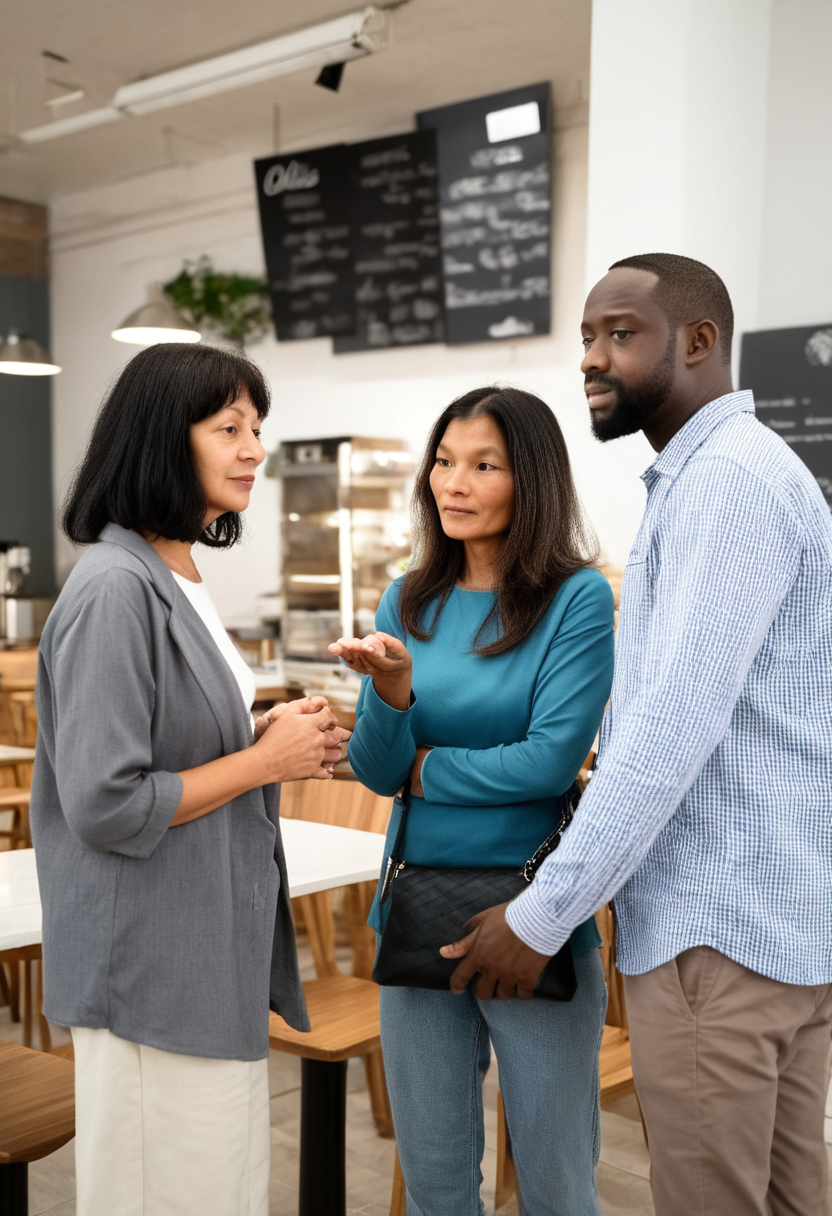} %
\caption{Result with pose}
\end{subfigure}
\caption{ID-Patch combined with pose conditions. Provided with user ID images in (a), our method can generate results with only the specifications of nose tip locations as shown in (b). Merging ID patches with pose images (c) enhances spatial control over the generated results as seen in (d), without incurring any computational overhead.}
\label{fig:idpatch_on_pose}
\end{center}
\vspace*{-6mm}
\end{figure}

Since human pose conditions are often utilized to enhance group photo generation, we can combine our ID-Patch approach with the pose conditions for multi-id image generation. As illustrated in \textbf{Fig.\,\ref{fig:idpatch_on_pose}}, the ID patches can be overlaid onto a pose-conditioned image, creating a composite condition image that encodes both identity positioning and pose information. It is important to mention that while our method can incorporate pose information, it does not inherently require it; pose is just an optional enhancement, one can also choose other conditions like canny edge, depth map for specific applications. The adaptability of our ID-Patch approach presents enhanced control over generated results yet without incurring the extra overhead of an additional ControlNet branch.

\subsubsection{Two-Stage Training}
\label{ssec:twostage}
\begin{figure}[t]
\begin{center}
\begin{subfigure}[b]{0.32\linewidth}
\includegraphics[width=\linewidth]{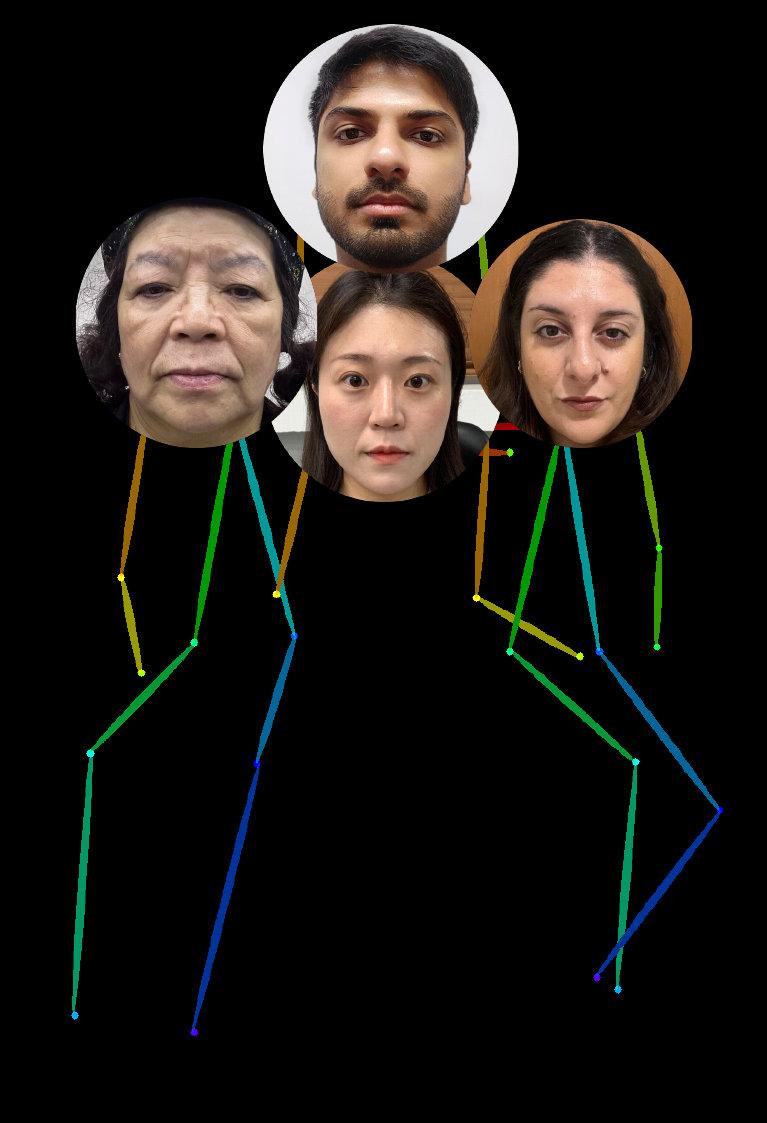} %
\caption{Pose + ID}
\end{subfigure}
\begin{subfigure}[b]{0.32\linewidth}
\includegraphics[width=\linewidth]{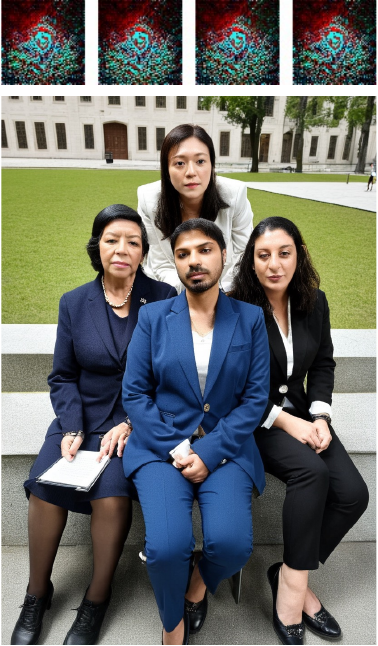} %
\caption{Single-stage train}
\end{subfigure}
\begin{subfigure}[b]{0.32\linewidth}
\includegraphics[width=\linewidth]{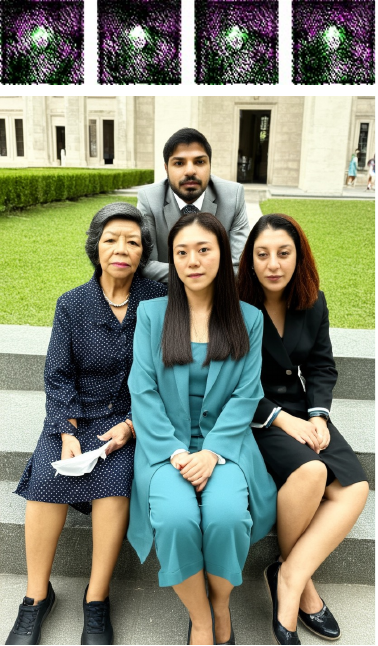} %
\caption{Two-stage train}
\end{subfigure}
\caption{Two-stage training to improve positioning robustness. Given pose and ID conditions in (a), single-stage training cannot fully prevent the incorrect face positioning issue. For example, in (b), the man is mistakenly placed at the central bottom position, generating an inharmonious result. (c) Two-stage training is introduced to solve this problem. As can be seen on the top row of this figure, the ID patches from our two-stage training approach are visually more distinguishable from each other, compared to the ones from single-stage training. Our experimental results prove that this is critical in solving the ID leak issues.}
\label{fig:single_stage}
\end{center}
\vspace*{-8mm}
\end{figure}

The training of our model employs a conventional diffusion loss function, defined as follows: \begin{equation} L = \mathbb{E}_{\mathcal{E}(x),\epsilon \sim \mathcal{N}(0, 1),t}[\lVert\epsilon - \epsilon_\theta(z_t, t, I, c'_t)\rVert_2^2] \end{equation} Initially, one might consider training the models end-to-end from scratch. However, our experiments reveal that such a method does not consistently ensure accurate ID localization, as depicted in \textbf{Fig.\,\ref{fig:single_stage}}. This shortcoming likely stems from the network’s tendency to preferentially inject ID information into ID embeddings, rather than ID patches, resulting in non-distinctive ID patches that could potentially confuse ControlNet. In the single-stage model, as shown in \textbf{Fig.\,\ref{fig:single_stage}}, the ID patches exhibit only subtle differences between different individuals.
To resolve this, we introduce a two-stage training approach. In the initial stage, ID embeddings are omitted, compelling the network to prioritize the incorporation of ID information into ID patches, thereby enhancing their distinctiveness. In the subsequent stage, ID embeddings are introduced and the model is fine-tuned to further improve the resemblance of the generated images.

\section{Experiments}
\label{sec:experiment}
\subsection{Settings}

\noindent\textbf{Implementation details.} 
 Facial features $f_i$ we use are 512-dimensional vectors derived from ArcFace~\cite{deng2019arcface}. During training, we utilize the SDXL diffusion model \cite{podellsdxl}.
 During the inference phase, we employ the further finetuned SDXL, namely Juggernaut-X-v10\footnote{https://huggingface.co/RunDiffusion/Juggernaut-X-v10}, as base model to enhance portrait quality.
The inference process involves 50 timesteps. 
We set the ID token injection ratio to 0.8 to regulate the influence of identity features, where the initial 20\% of timesteps operate without ID token embeddings to focus on the general style, followed by the remaining 80\% timesteps which incorporate ID token embeddings to refine the identity-specific aspects of the images. 
The default size of ID-patch is $P\times P = 64 \times 64$ and the default length of ID embeddings is $M = 16$. The network is trained using 8 NVIDIA H100 GPUs with batch size=4, learning rate=1e-05, weight decay = 0.01, using AdaFactor optimizer~\cite{shazeer2018adafactor}. We train it for 720k iterations for stage 1, and stage 2 usually converges around 1 million iterations. We select the best model based on a 20 people validation set.

\noindent\textbf{Dataset.}
Our training set consists of 17 million single-person and 1.95 million multi-person images. These images are obtained from purchased data and publicly available sources.
For evaluation, we utilize two specialized datasets:
1) \textit{Style Dataset} comprises 40 styles with pose and text prompt definitions, ranging from one to eight faces. 
2) \textit{Identity Dataset} includes 600 unique identities representing a variety of races and genders. 
Each style is paired with 20 randomly sampled unique identity combinations for quantitative evaluation.

\noindent\textbf{Baselines.} Our analysis includes two state-of-the-art multi-ID generation methods as baselines:  1) \textbf{OMG}~\cite{kong2024omg}: This method incorporates multiple identity features into separate segmented regions through latent stitching. It can be paired with a single ID injection method InstantID~\cite{wang2024instantid} for multi-ID generation. 2) \textbf{InstantFamily}~\cite{kim2024instantfamily}: This approach employs coarse attention masks within cross-attention layers, a mandatory process that ensures distinct identity features are injected into different image instances effectively.  Because there is no official publicly available code for InstantFamily~\cite{kim2024instantfamily}, for fairness, we implemented it based on the SDXL base model~\cite{podellsdxl}, and trained it using the same training data as our model.

\begin{table}

\centering
\caption{Performance comparison across four dimensions: identity resemblance, ID-position association accuracy, text alignment, and generation time. For qualitative metrics to evaluate similarity or accuracy, higher is better. Our approach significantly outperforms baseline methods in terms of resemblance, association, and generation time, with similar text alignment scores.}
\vspace*{-3mm}
\scalebox{0.8}{
\begin{tabular}{c|cccc}
    \toprule[1pt]
    \midrule
       Methods &ID $\uparrow$  & Association $\uparrow$ & Text $\uparrow$ & 
       Time (s) $\downarrow$\\
    \midrule
    OMG + InstantID & 0.644& 0.926& 0.271&   84.83\\
    InstantFamily & 0.683 & 0.862 & \textbf{0.278}&  10.02 \\
    ID-Patch (Ours) &\textbf{0.751} &  \textbf{0.958} & 0.273 & \textbf{9.69} \\
  \midrule
  \bottomrule[1pt]
\end{tabular}
}
\vspace*{-2mm}

\label{tab:compare}
\end{table}

\begin{figure*}[t]
\vspace*{-4mm}
\begin{center}
\begin{subfigure}[b]{0.28\linewidth}
\includegraphics[width=\linewidth]{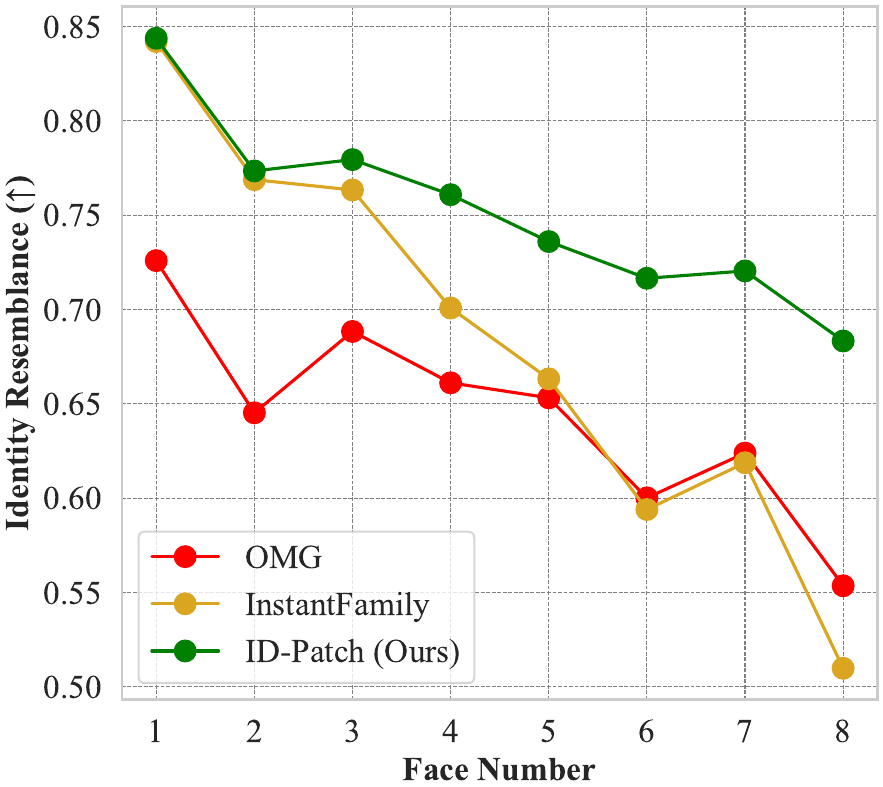} %
\caption{Identity Resemblance}
\end{subfigure}
\quad\quad
\begin{subfigure}[b]{0.28\linewidth}
\includegraphics[width=\linewidth]{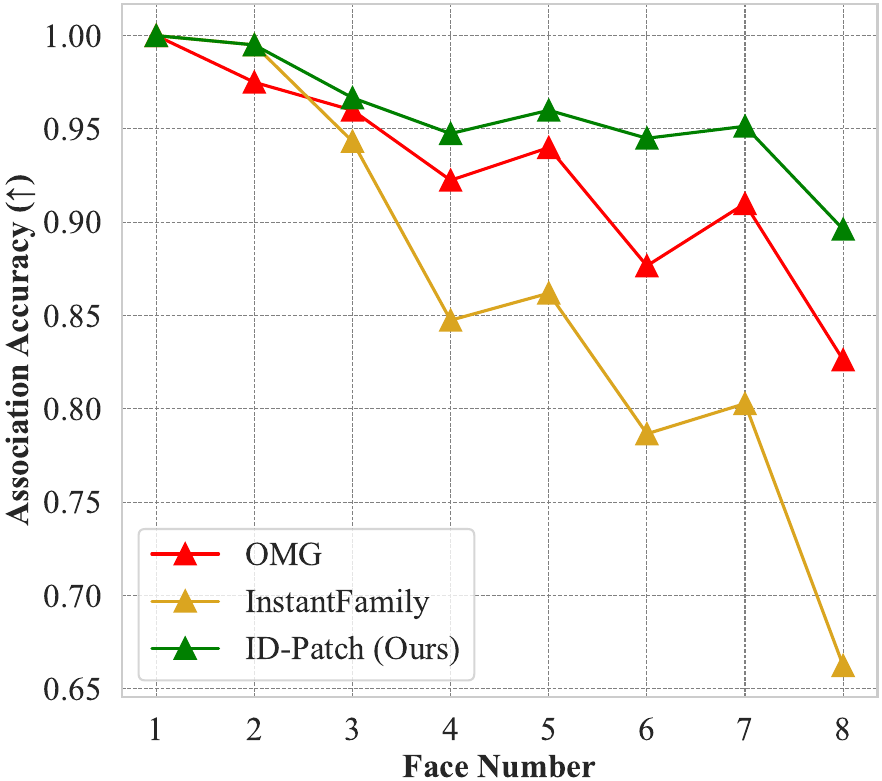} %
\caption{Association Accuracy}
\end{subfigure}
\quad\quad
\begin{subfigure}[b]{0.28\linewidth}
\includegraphics[width=\linewidth]{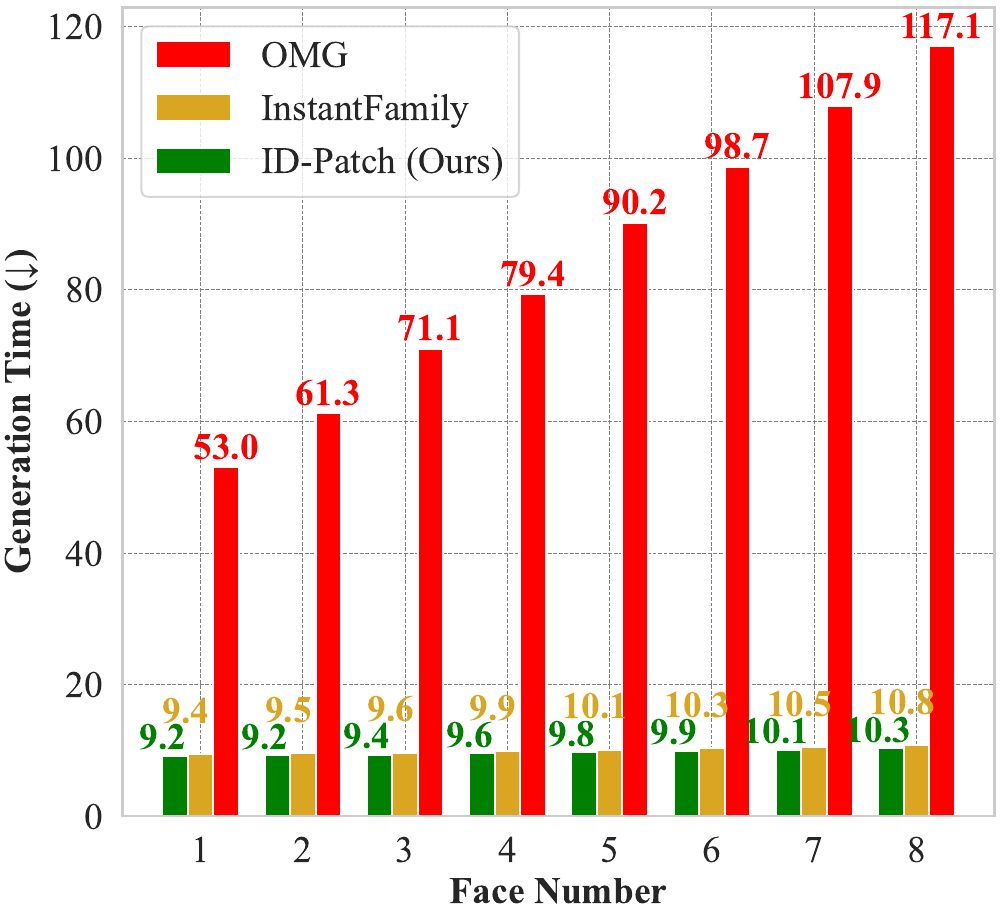} %
\caption{Generation Time}
\end{subfigure}
\caption{Performance evaluation of different model generations from three different aspects: identity resemblance, location accuracy, and generation time. As face number increases, the ID and location metric scores of OMG and Instantfamily drop much more significantly than our approach. Our approach also achieves near-constant generation time, in contrast to OMG's linearly increasing running time. 
}
\label{fig:compare_n_face_all}
\end{center}
\vspace*{-5mm}
\end{figure*}

 \begin{figure*}[t]
  \centering
  \resizebox{0.99\textwidth}{!}{
  \begin{tabular}{c|c|c|c}
  \toprule[1pt]
  \midrule
  \multicolumn{1}{c|}{\scriptsize{\textbf{ID + Pose}}}  & \multicolumn{1}{c|}{\scriptsize{\textbf{OMG}}}
 & \multicolumn{1}{c|}{\scriptsize{\textbf{InstantFamily}}}
 & \multicolumn{1}{c}{\scriptsize{\textbf{ID-Patch (Ours)}}}
    \\
\midrule

    \begin{minipage}{0.18\textwidth}\centering\includegraphics[width=\linewidth]{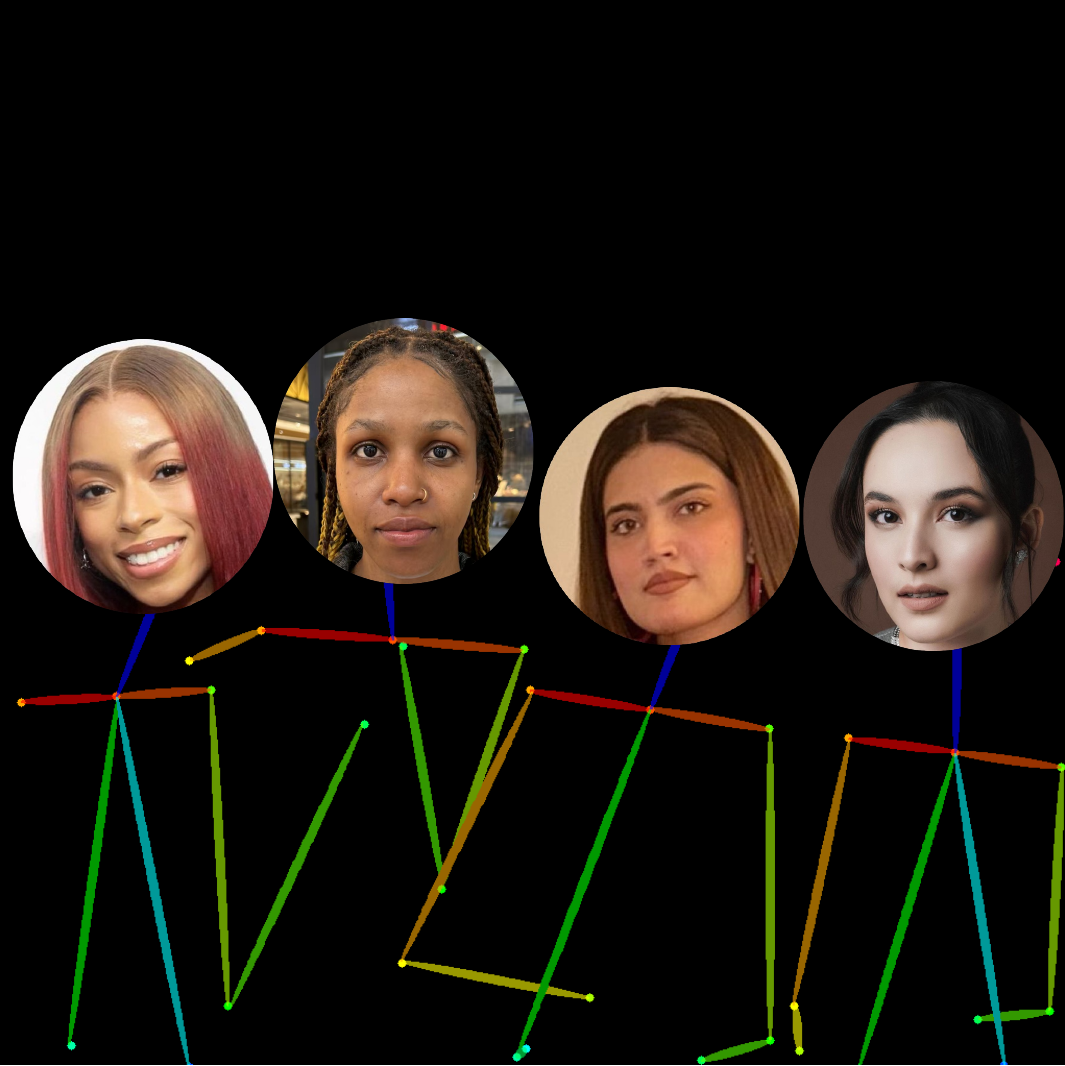}\end{minipage} 
    \vspace*{1mm} 
     &
    \begin{minipage}{0.18\textwidth}\centering\includegraphics[width=
    \linewidth]{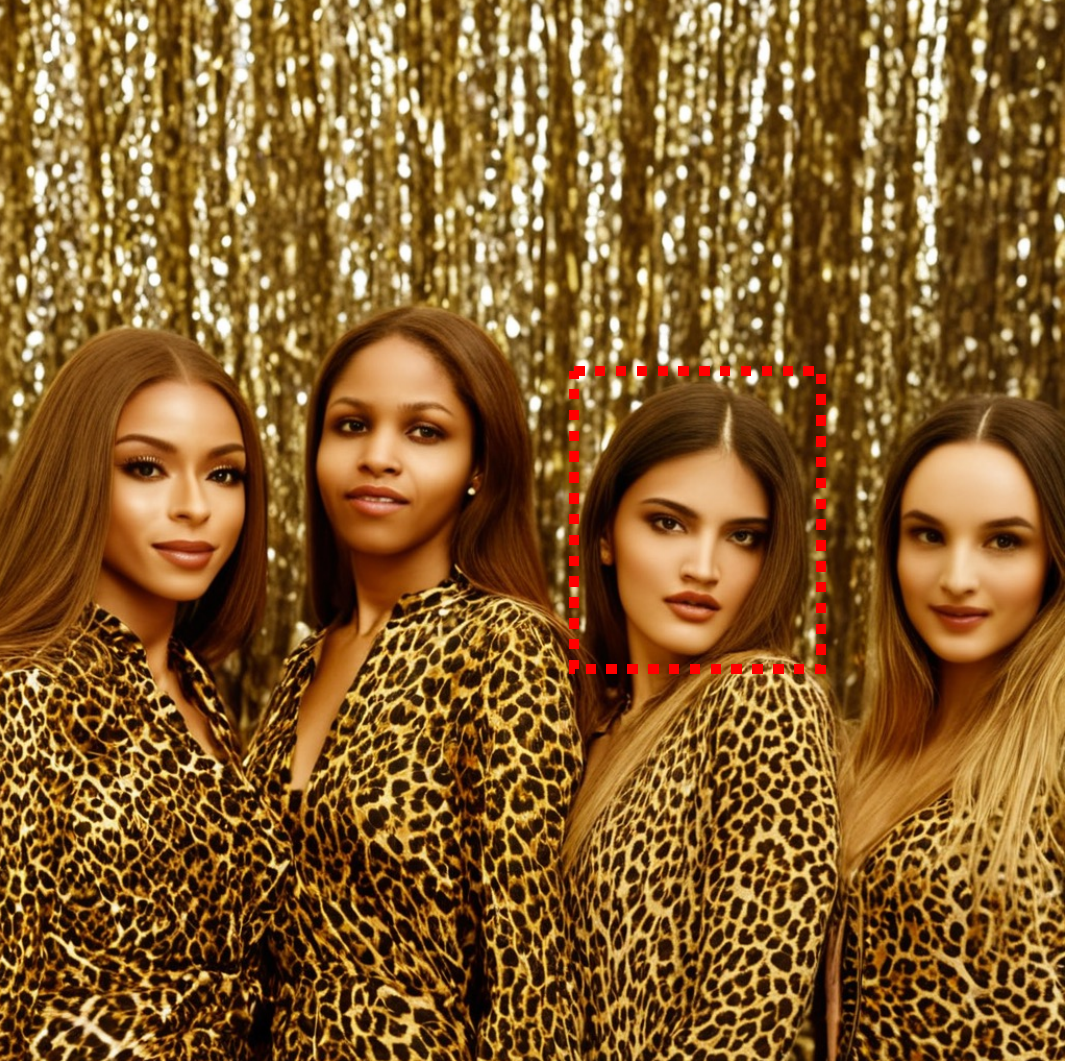}\end{minipage}
    &
    \begin{minipage}{0.18\textwidth}\centering\includegraphics[width=\linewidth]{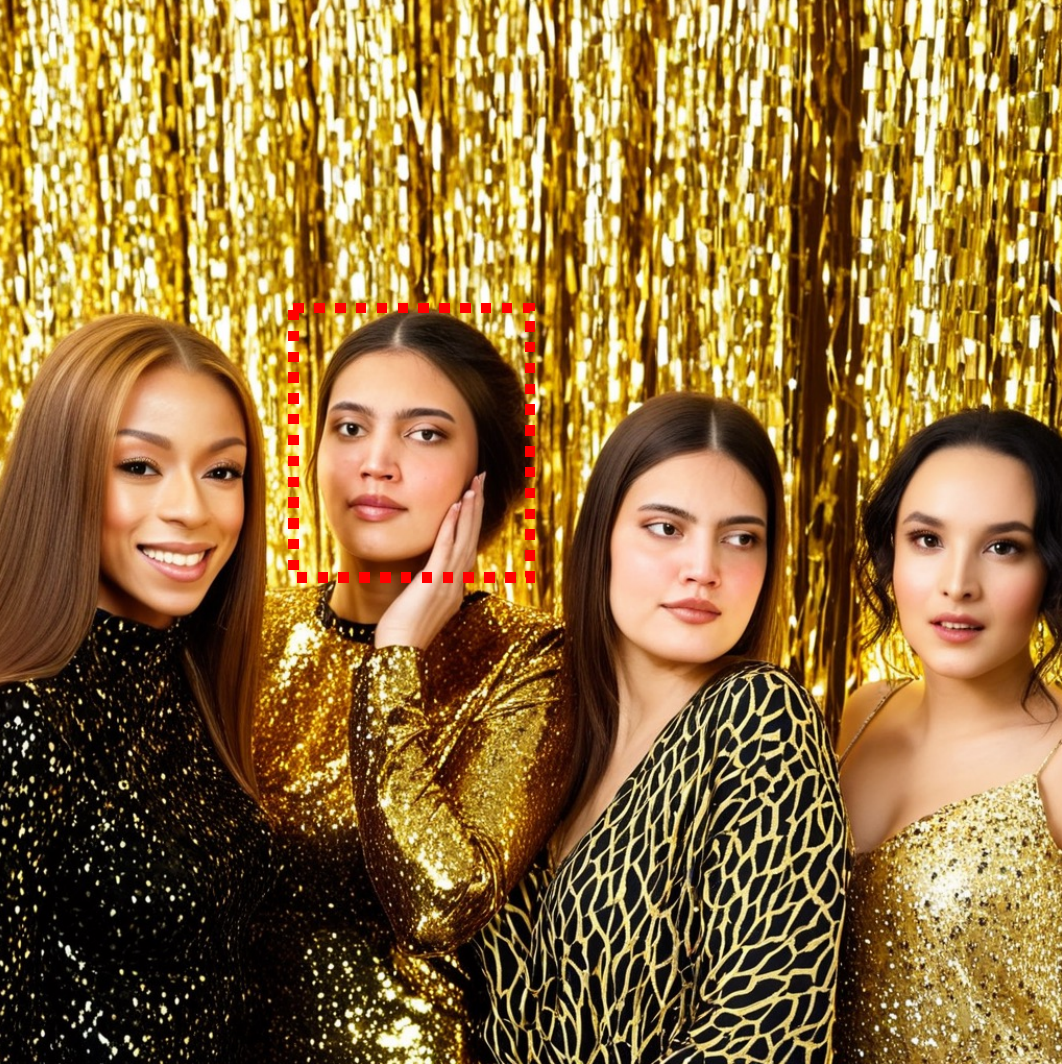}\end{minipage}
    &
    \begin{minipage}{0.18\textwidth}\centering\includegraphics[width=\linewidth]{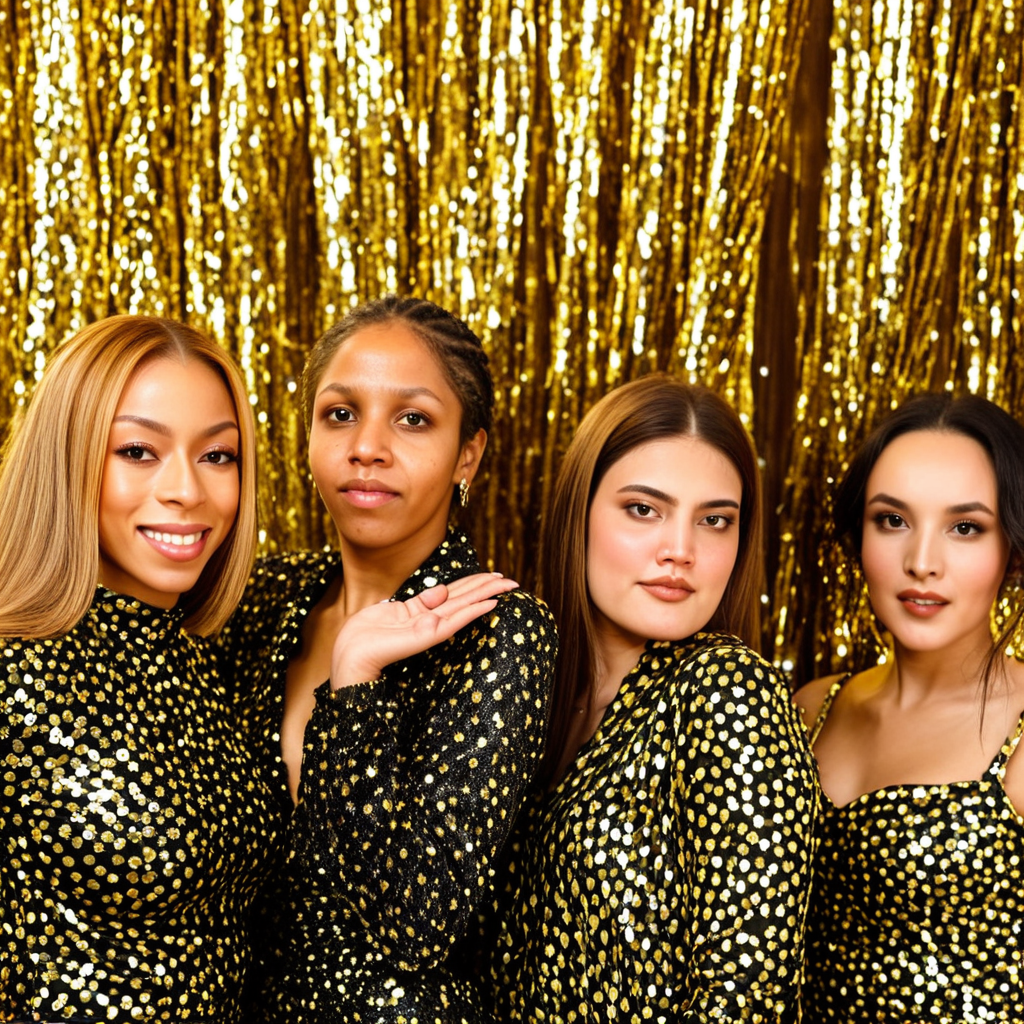}\end{minipage}
    \\
\midrule

    \begin{minipage}{0.18\textwidth}\centering\includegraphics[width=\linewidth]{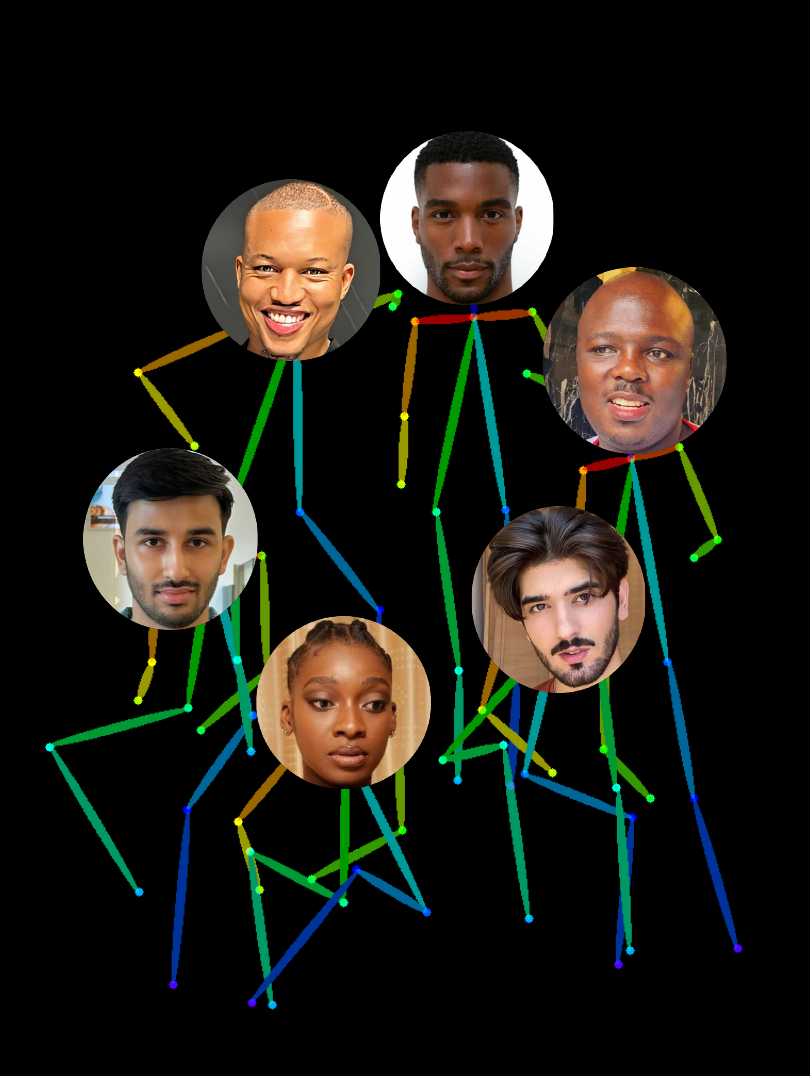}\end{minipage} 
    \vspace*{1mm} 
     &
    \begin{minipage}{0.18\textwidth}\centering\includegraphics[width=\linewidth]{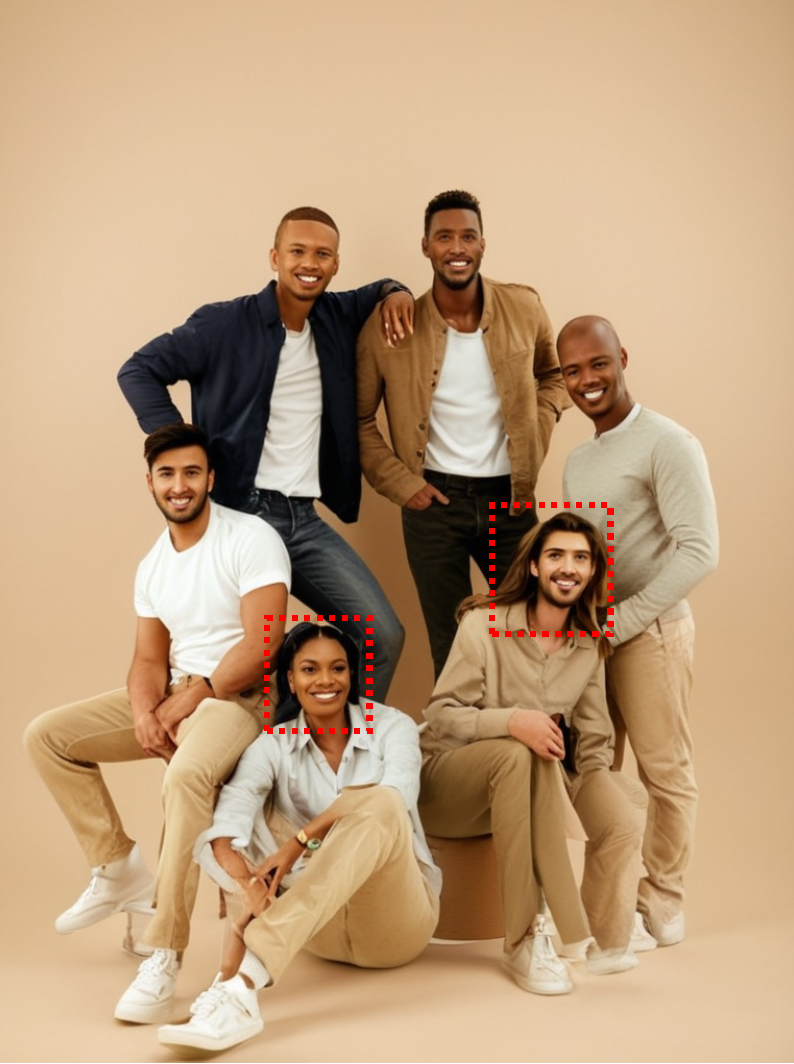}\end{minipage}
    &
    \begin{minipage}{0.18\textwidth}\centering\includegraphics[width=\linewidth]{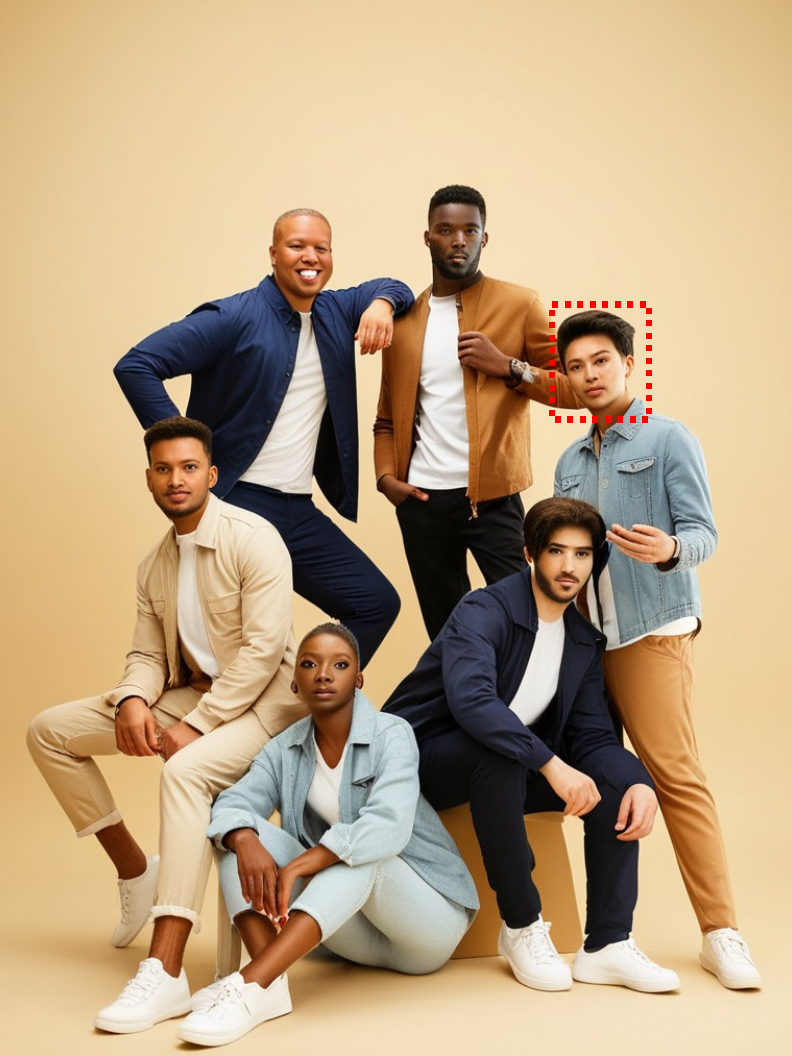}\end{minipage}
    &
    \begin{minipage}{0.18\textwidth}\centering\includegraphics[width=\linewidth]{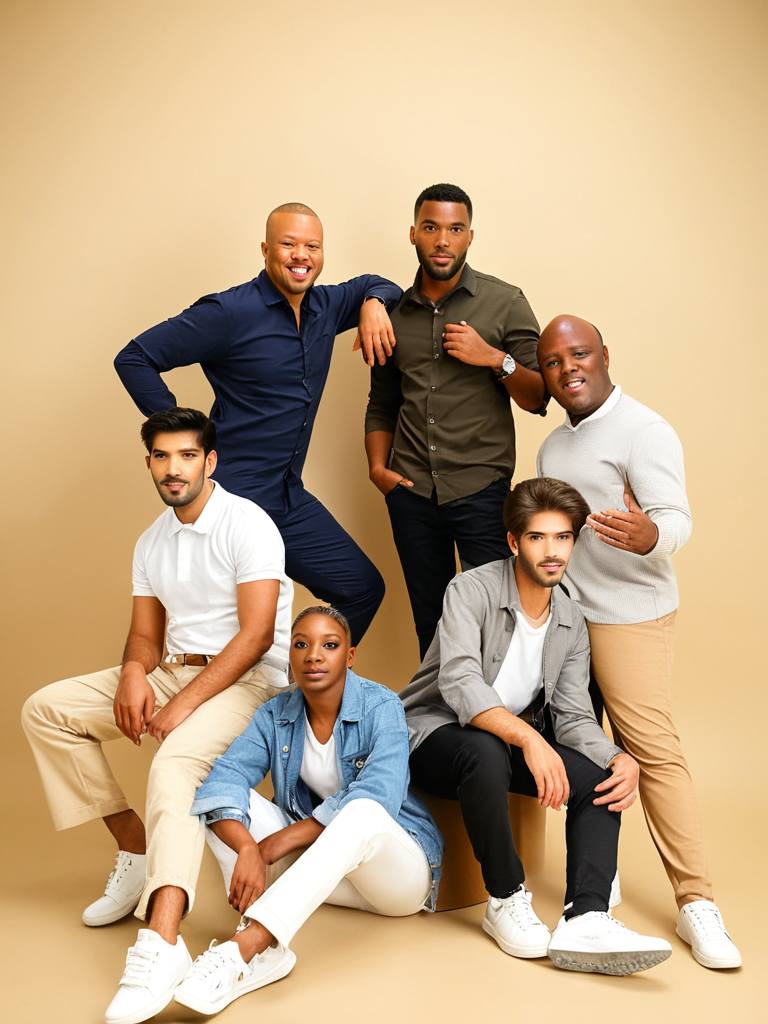}\end{minipage}
    \\
  \midrule
  \bottomrule[1pt]
\end{tabular} 

  }
  \caption{Comparison with baselines on pose-conditioned generation, where red dashed boxes highlight instances with low identity resemblance. In row 1, OMG fails to preserve the face shape of the third woman (from left to right) because its first stage result conflicts with this ID's face shape. InstantFamily incorrectly generates the second person because of the ID leakage from the third woman. In row 2, OMG does not generate correct hair styles and accurate facial features for the two people in the red boxes, while InstantFamily generates the wrong ID in the red box.}
  \label{fig:compare}
  \vspace*{-3mm}
\end{figure*}

\noindent\textbf{Evaluation Metrics.} We evaluate the methods in a pose-conditioned setting, because pose is a requirement of InstantFamily, and without pose condition, OMG is not able to generate faces at specified locations for its first stage. Besides, pose condition makes the head size consistent, which is easier for resemblance evaluation and reduces the impact of head sizes on generation quality. For comprehensive evaluation, we consider four critical dimensions: 
1) \underline{Identity} resemblance: we measure the identity resemblance by calculating the cosine similarity (CosSim) between the facial features of faces in the generated images and the reference faces. For fairness, we use FaceNet \cite{schroff2015facenet}, a face model different from the ArcFace model used for training.
2) \underline{Association} accuracy: we assess the accuracy of identity-position association in generated images by ensuring that the generated faces are most similar to their assigned counterparts, compared to all other input faces used in the generation process. In other words, we calculate the similarity between the $i$-th generated face and the $j$-th input face. Let $s(i) = \arg \max_{0 \leq j \leq N-1} \text{CosSim}(\Tilde{f}_{i}^{\text{\text{gen}}}, \Tilde{f}_j) $ be the face index among all input faces that is most similar to the $i$-th generated face, where $\Tilde{f}_{i}^{\text{gen}}$, $\Tilde{f}_j$ denote the FaceNet features of the $i$-th generated face and the $j$-th input face respectively, and $N$ is the number of faces in this image. When $s(i)=i$, the $i$-th generated face has the correct association. The identity-position association accuracy is defined as: 
\begin{equation}\label{location_distance}
    \frac{1}{N} \sum_{i=0}^{N-1} \mathbf{1}{\left\{ i = s(i) \right\}}
\end{equation}
This measurement assesses the correct placement of each identity within the image. 
3) \underline{Text} alignment: this metric evaluates the alignment between the textual context of the image generation prompt and the visual output. Cosine similarity is used to compare textual features extracted by the CLIP text encoder with visual features extracted by the CLIP image encoder from the generated images~\cite{radford2021learning}. 
4) Generation \underline{time}: we measure the efficiency by recording the time (in seconds) required to generate an image on a NVIDIA A100 GPU, excluding the time taken for model loading and image I/O.

 \begin{figure*}[t]
\vspace{-3.5mm}
  \centering
  \resizebox{\textwidth}{!}{
  \begin{tabular}{c|c|c|c|c}
  \toprule[1pt]
\midrule
      \multirow{1}[4]{*}[0.8cm]{
\rotatebox{90}{ 
 \scriptsize{\textbf{ID + ID Location}} }
    } 
    &
    \begin{minipage}{0.14\textwidth}\centering\includegraphics[width=\linewidth]{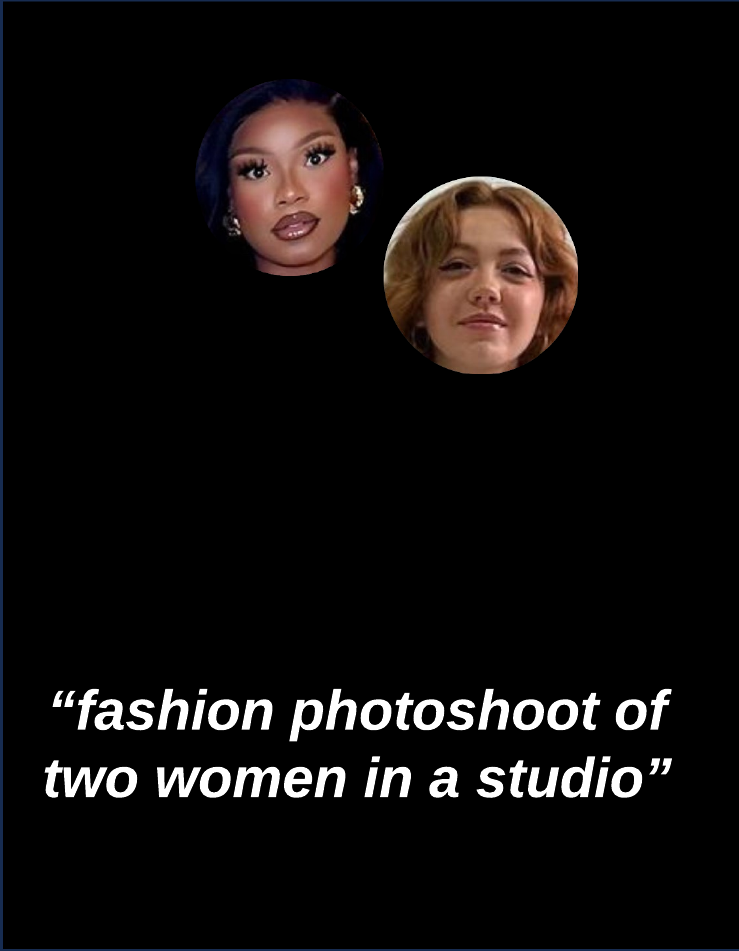}\end{minipage} 
    \vspace*{1mm} 
     &
    \begin{minipage}{0.18\textwidth}\centering\includegraphics[width=\linewidth]{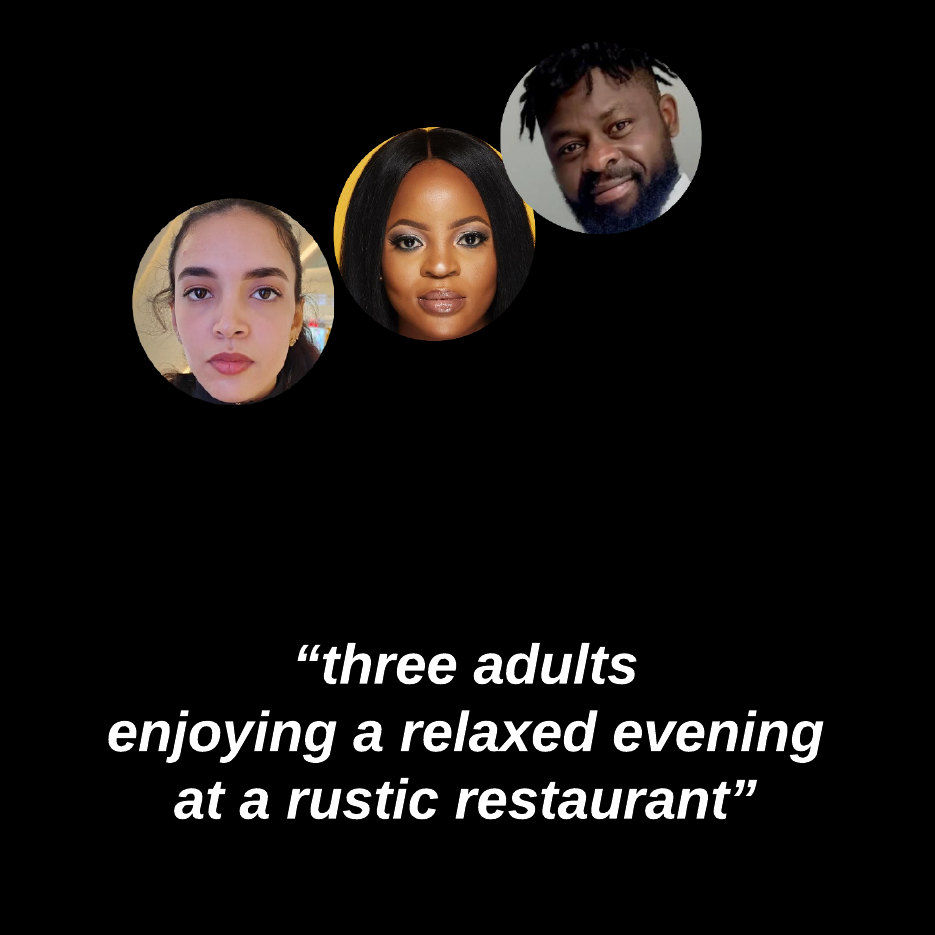}\end{minipage}
    &
    \begin{minipage}{0.205\textwidth}\centering\includegraphics[width=\linewidth]{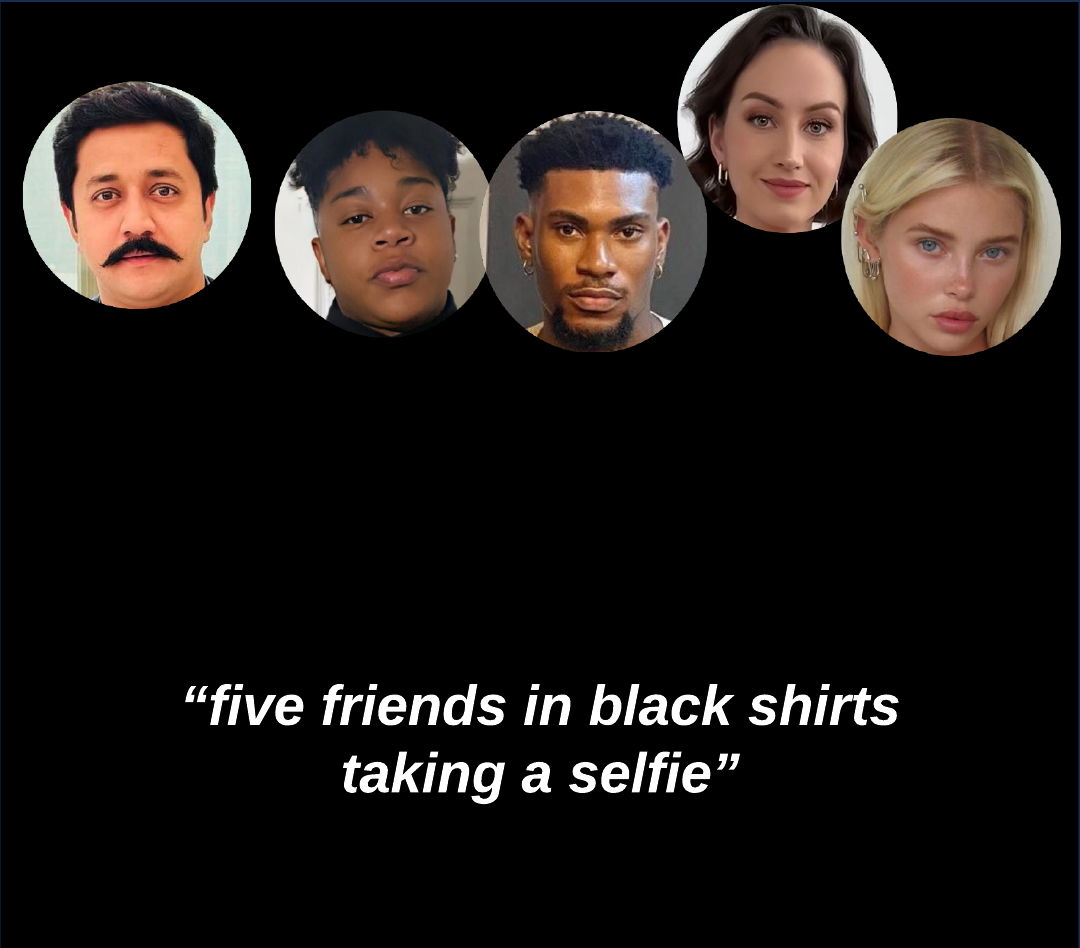}\end{minipage}
    &
    \begin{minipage}{0.18\textwidth}\centering\includegraphics[width=\linewidth]{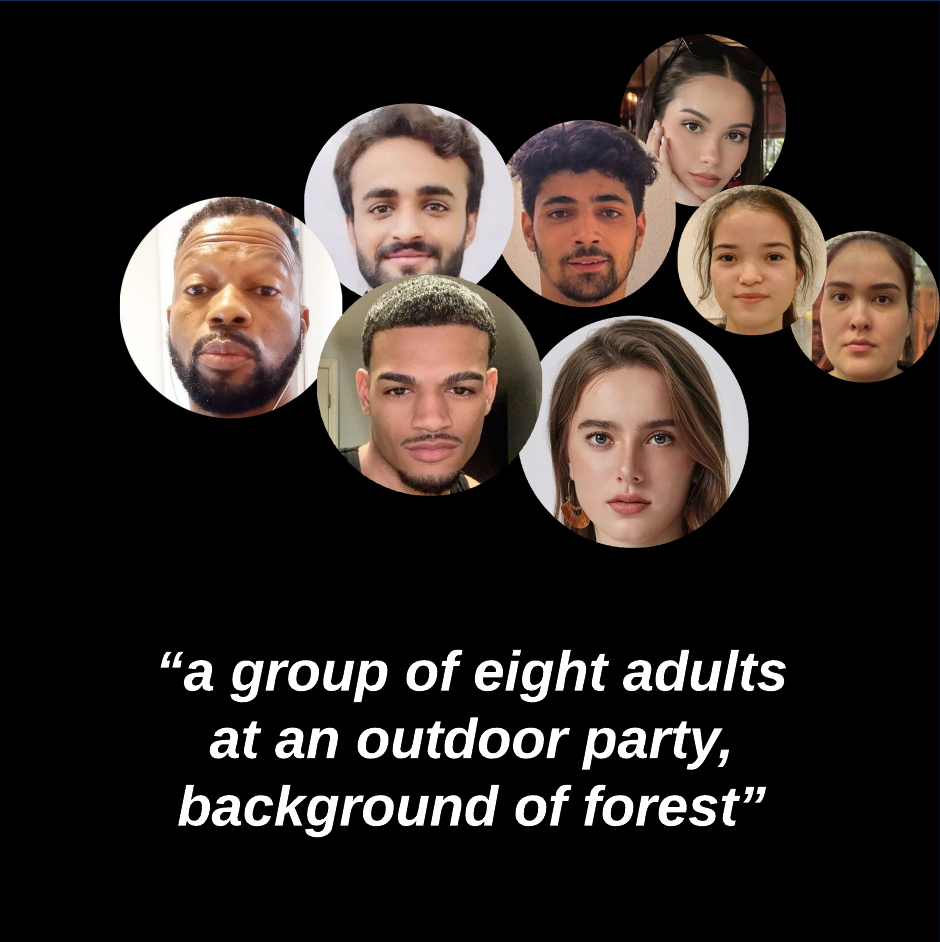}\end{minipage}
    \\
\midrule
\multirow{1}[4]{*}[0.8cm]{
\rotatebox{90}{ 
 \scriptsize{\textbf{Generation Result}} }
    } 
    &
    \begin{minipage}{0.14\textwidth}\centering\includegraphics[width=\linewidth]{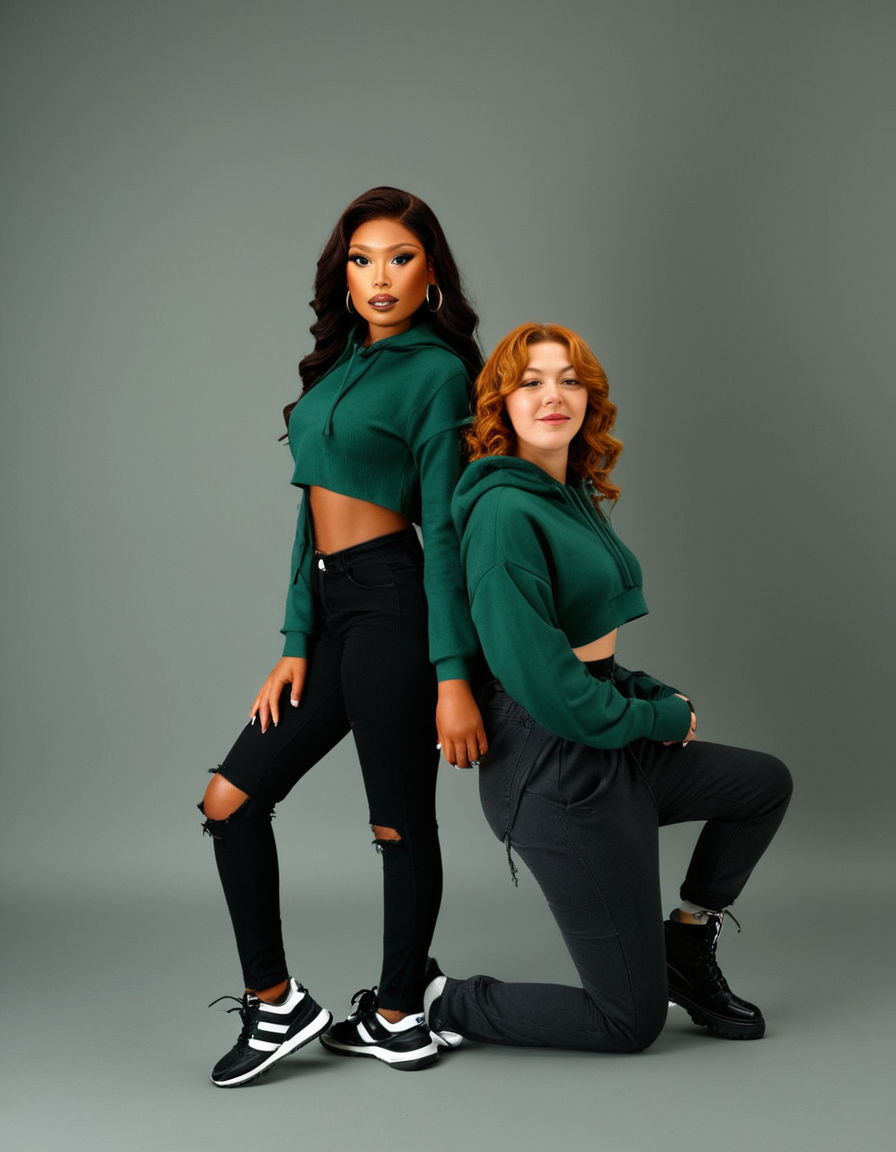}\end{minipage} 
    \vspace*{1mm} 
     &
    \begin{minipage}{0.18\textwidth}\centering\includegraphics[width=\linewidth]{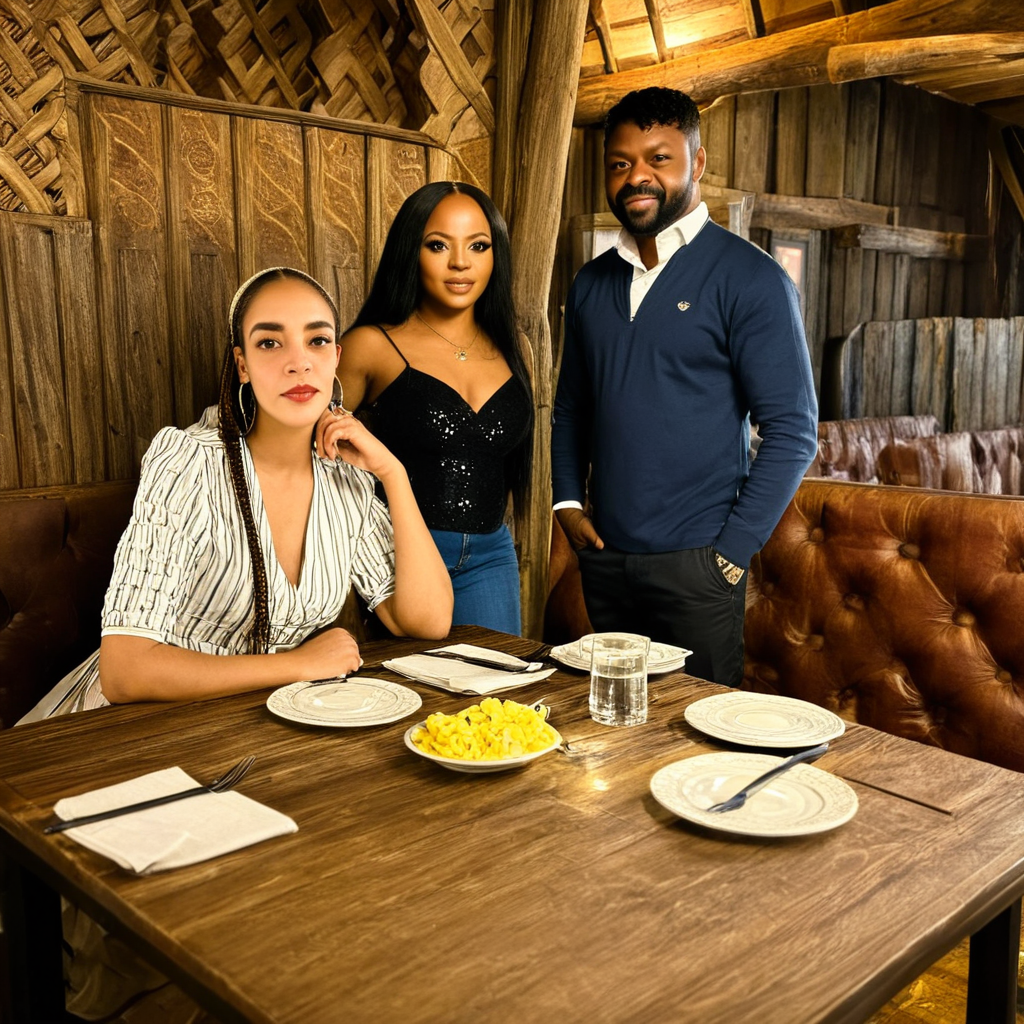}\end{minipage}
    &
    \begin{minipage}{0.205\textwidth}\centering\includegraphics[width=\linewidth]{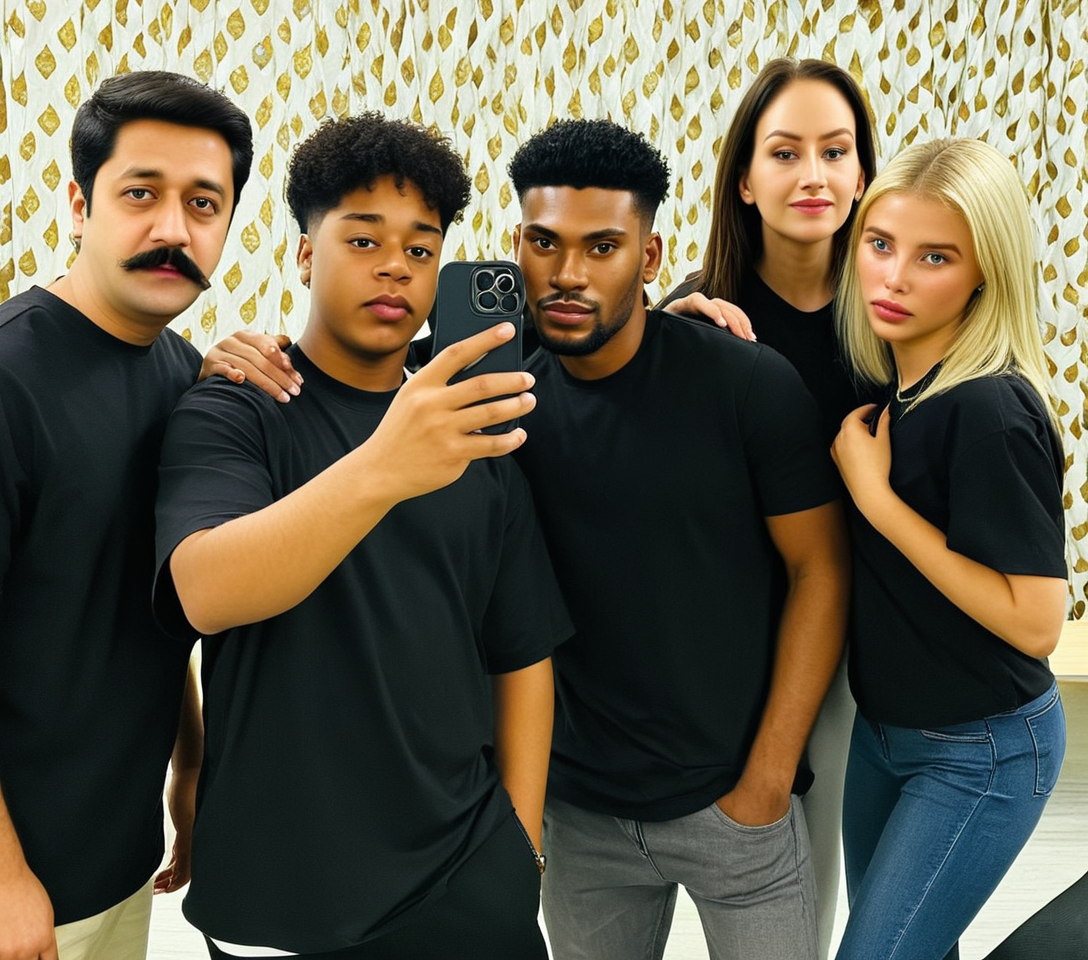}\end{minipage}
    &
    \begin{minipage}{0.18\textwidth}\centering\includegraphics[width=\linewidth]{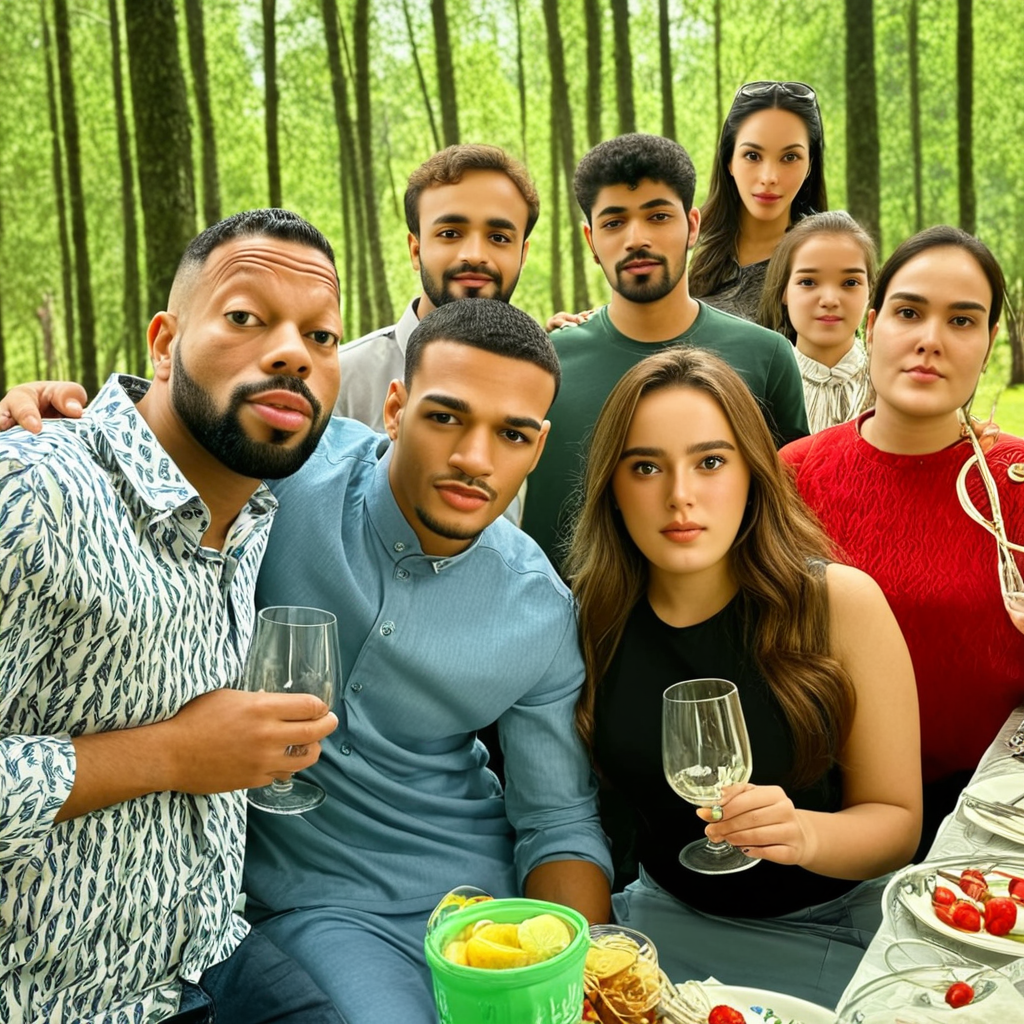}\end{minipage}
    \\
\midrule
  \bottomrule[1pt]
\end{tabular} 

  }
  \vspace{-1.5mm}
  \caption{ID-Patch pose-free generation. Here the condition image is only ID patches rendered onto a black canvas without any pose condition. The generation is conditioned on individual face locations while corresponding head sizes and body poses are inferred implicitly. Our method can accommodate a large number of people with diverse ethnic backgrounds, yet generating visually appealing group photo results. See ``Evaluation Metrics" for reasons why we compare with baseline approaches only in pose-conditioned settings.
  }
  \label{fig:nopose}
  \vspace*{-5mm}
\end{figure*}

\subsection{Comparison with Baselines}
As listed in \textbf{Tab.\,\ref{tab:compare}}, ID-Patch excels in identity and association metrics with scores of 0.751 and 0.958, respectively, showcasing its robust capability to maintain identity consistency and achieve precise placement in generated images.
There is no significant difference in text alignment scores between compared methods.
Notably, ID-Patch achieves the fastest image generation time.

Further details on the performance with varying face numbers are shown in \textbf{Fig.\,\ref{fig:compare_n_face_all}}. 
As we observed no deterioration in text alignment with an increase in the number of faces across all methods, we have relegated the corresponding figure to the appendix.
As the number of faces increases, both OMG and InstantFamily experience significant performance drops due to ID leakage. Although ID-Patch also shows a decline, it is less severe, mainly because the increase in face numbers reduces the average face area, adversely affecting the quality of smaller faces produced by the SDXL models. Moreover, the presence of more faces slightly extends the image generation time due to the increased demand for facial feature extraction. InstantFamily is slightly slower than our approach due to the overhead of masked attention. For OMG, injecting identity features into designated segmented areas causes a substantial increase in generation time; it takes about 2 minutes to generate an image with eight people, whereas ID-Patch requires only about 10 seconds. See \textbf{Fig.\,\ref{fig:compare}} for visual comparisons.

\subsection{Ablation Study}
\begin{table}

\centering
\caption{Ablation study results for our proposed ID-Patch variants, designed to assess the impact of different modules (patch projection and token projection) and training/inferencing schemes on the performance of multi-ID image generation. The full method achieves the best ID resemblance and ID-position association, indicating the necessity of each module.}
\vspace*{-3mm}
\scalebox{0.8}{
\begin{tabular}{c|ccc}
    \toprule[1pt]
    \midrule
       Variants &ID $\uparrow$ & Association $\uparrow$ & Text $\uparrow$\\
    \midrule
    No patch projection & 0.368& 0.345&0.271   \\
    No token projection & 0.565&0.895 & \textbf{0.284}  \\
    Single stage training &0.660 &0.784 &0.271 \\
     ID token injection ratio = 1.0 &0.750&0.957&0.270 \\
    \midrule
    FULL &\textbf{0.751} &  \textbf{0.958} & 0.273  \\
    \midrule
    \bottomrule[1pt]
\end{tabular}
}
\vspace{-3mm}

\label{tab:ablation}
\end{table}

We conduct an ablation study on variants of the proposed method in \textbf{Tab.\,\ref{tab:ablation}}. 
The performance notably degrades when patch projection is omitted, leading to poor location alignment as the algorithm attempts to infer ID locations based on indirect cues such as gender and pose. Similarly, omitting token projection reduces resemblance by eliminating essential facial details, quantitatively validates the finding of \textbf{Fig.\,\ref{fig:id_token}}. Single-stage training results in indistinguishable ID patches (as shown in \textbf{Fig.\,\ref{fig:single_stage})}, which impairs effective positioning. 
The default setting with an ID token injection ratio of 0.8 improves text alignment compared to injecting the ID token at every timestep, while maintaining the same level of identity resemblance. This improvement may arise because ID embeddings have little effect on facial detail during the early denoising steps, but become more influential in the later stages. 
It is important to note that even without the inference trick (i.e., using the ID token injection ratio), our proposed method still outperforms all baseline approaches.

\subsection{Pose-Free Generation}
Our ID-Patch model can be operated effectively without the need for pose conditioning, a flexibility that permits its integration with other conditions for image generation. This capability enables the use of ID-Patch in versatile configurations, whether plug-and-play or co-trained, which we will explore in subsequent subsections.

In the pose-free mode, the ControlNet uses a conditioning image that solely features ID patches against a black background. The placement of each ID patch at the tip of the nose ensures accurate preservation of facial locations during image synthesis, while head sizes and poses are variably generated based on the initial noise. As illustrated in \textbf{Fig.~\ref{fig:nopose}}, this methodology allows our model to produce aesthetically pleasing group photographs of different number of individuals across a variety of settings, demonstrating the robustness and adaptability of our approach.

\begin{figure}[t]
\begin{center}
\begin{subfigure}[b]{0.2\linewidth}
\includegraphics[width=\linewidth]{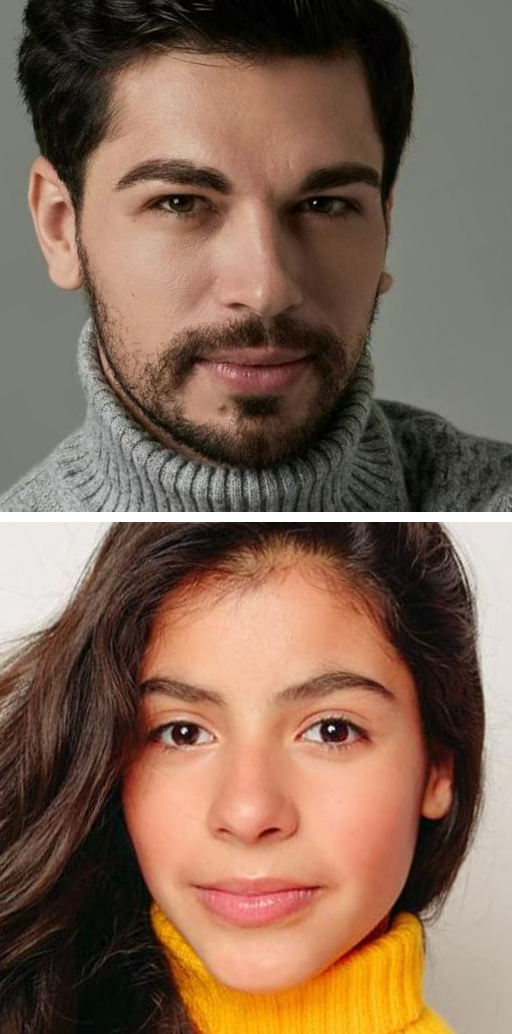} %
\caption{ID}
\end{subfigure}
\begin{subfigure}[b]{0.39\linewidth}
\includegraphics[width=\linewidth]{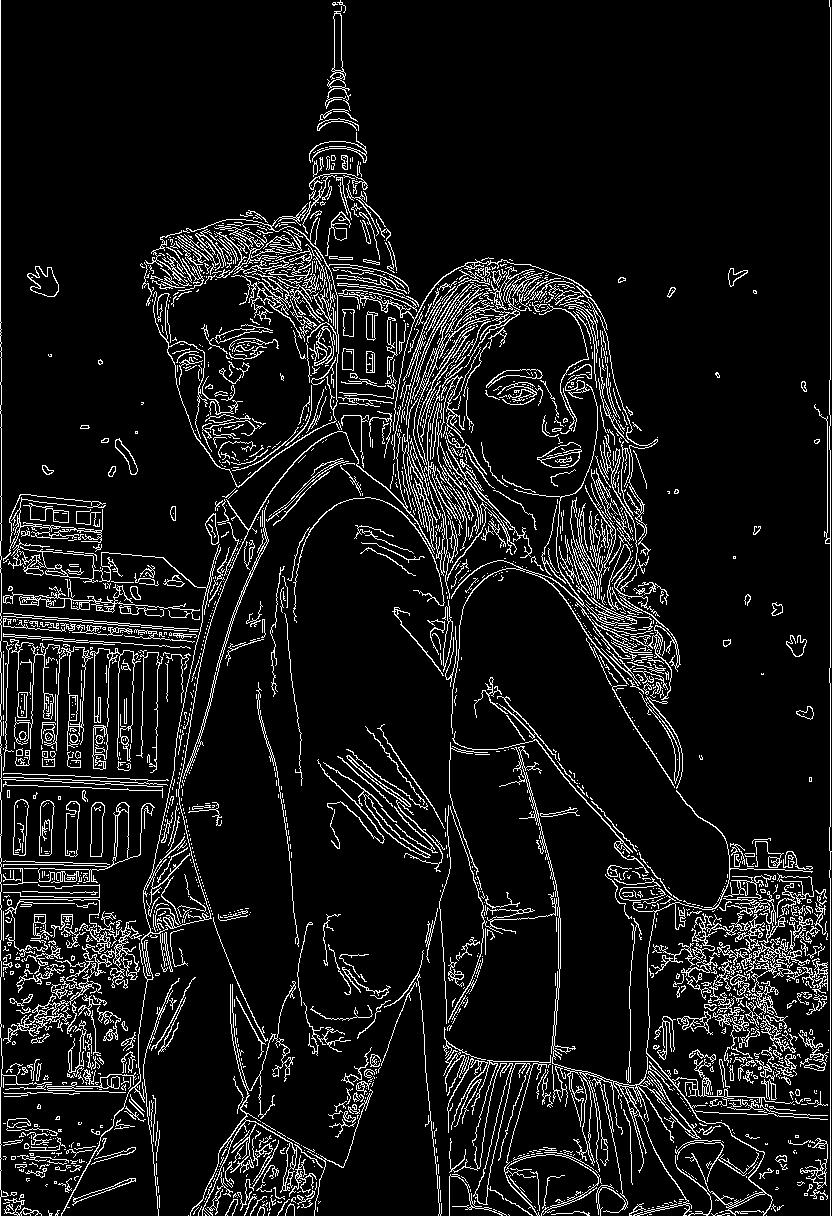} %
\caption{Canny edge condition}
\end{subfigure}
\begin{subfigure}[b]{0.39\linewidth}
\includegraphics[width=\linewidth]{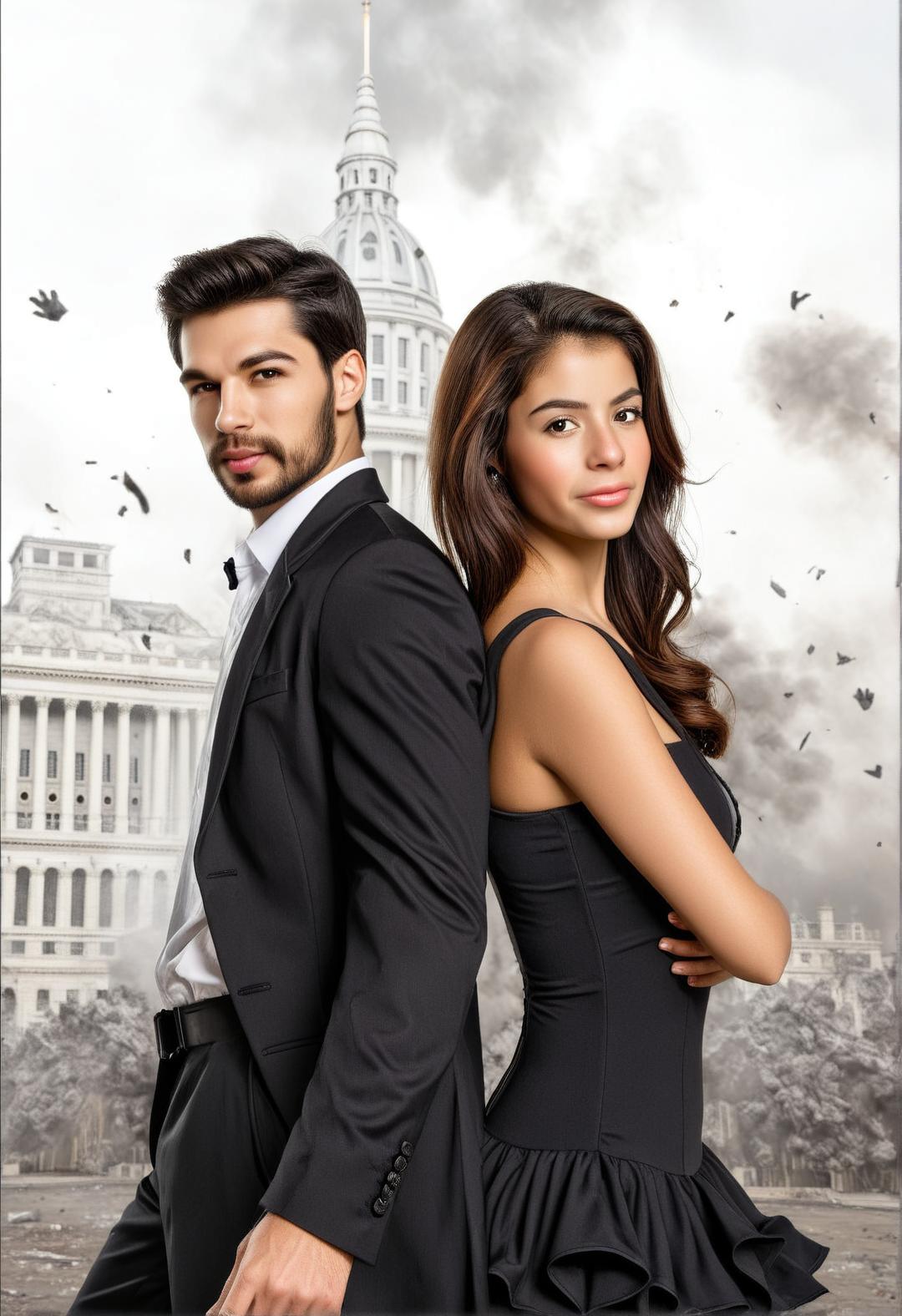} %
\caption{ID-Patch result}
\end{subfigure}
\caption{Combination of pose-free ID-Patch ControlNet and pretrained Canny edge ControlNet. We can generate background and clothing details, with high quality ID injection.}
\label{fig:plug_and_play}
\end{center}
\vspace*{-7mm}
\end{figure}

The pose-free version of ID-Patch is designed to be compatible with other ControlNets (\textbf{plug-and-play}). As illustrated in \textbf{Fig.\,\ref{fig:plug_and_play}}, it is possible to combine our ID-Patch ControlNet the Canny edge ControlNet to achieve more precise control over the final image outcomes.

\subsection{Limitations}

\begin{figure}[!t]
\begin{center}
\begin{subfigure}[b]{0.42\linewidth}
\includegraphics[width=\linewidth]{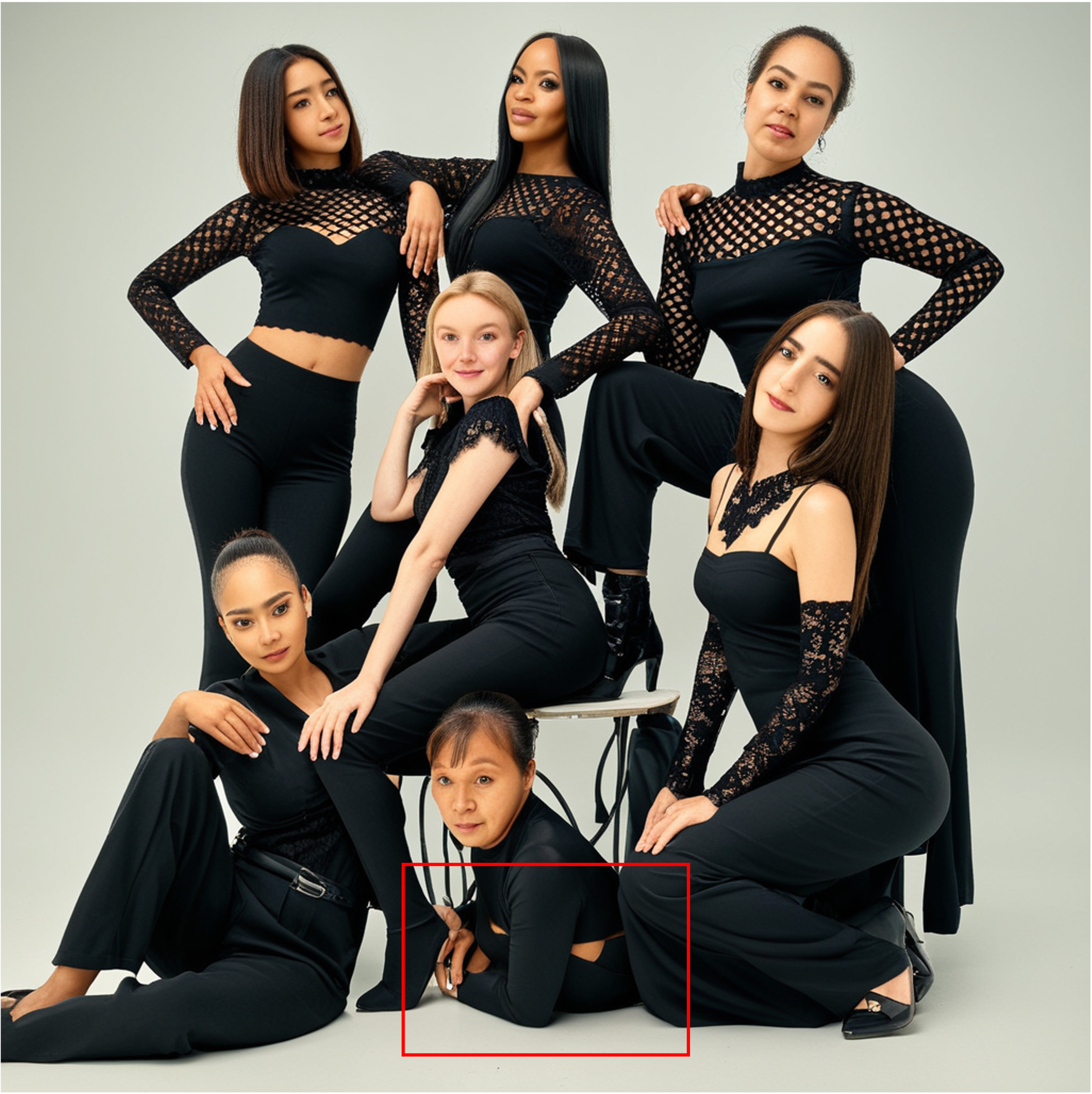} %
\caption{Abnormal anatomy}
\end{subfigure}
\begin{subfigure}[b]{0.56\linewidth}
\includegraphics[width=\linewidth]{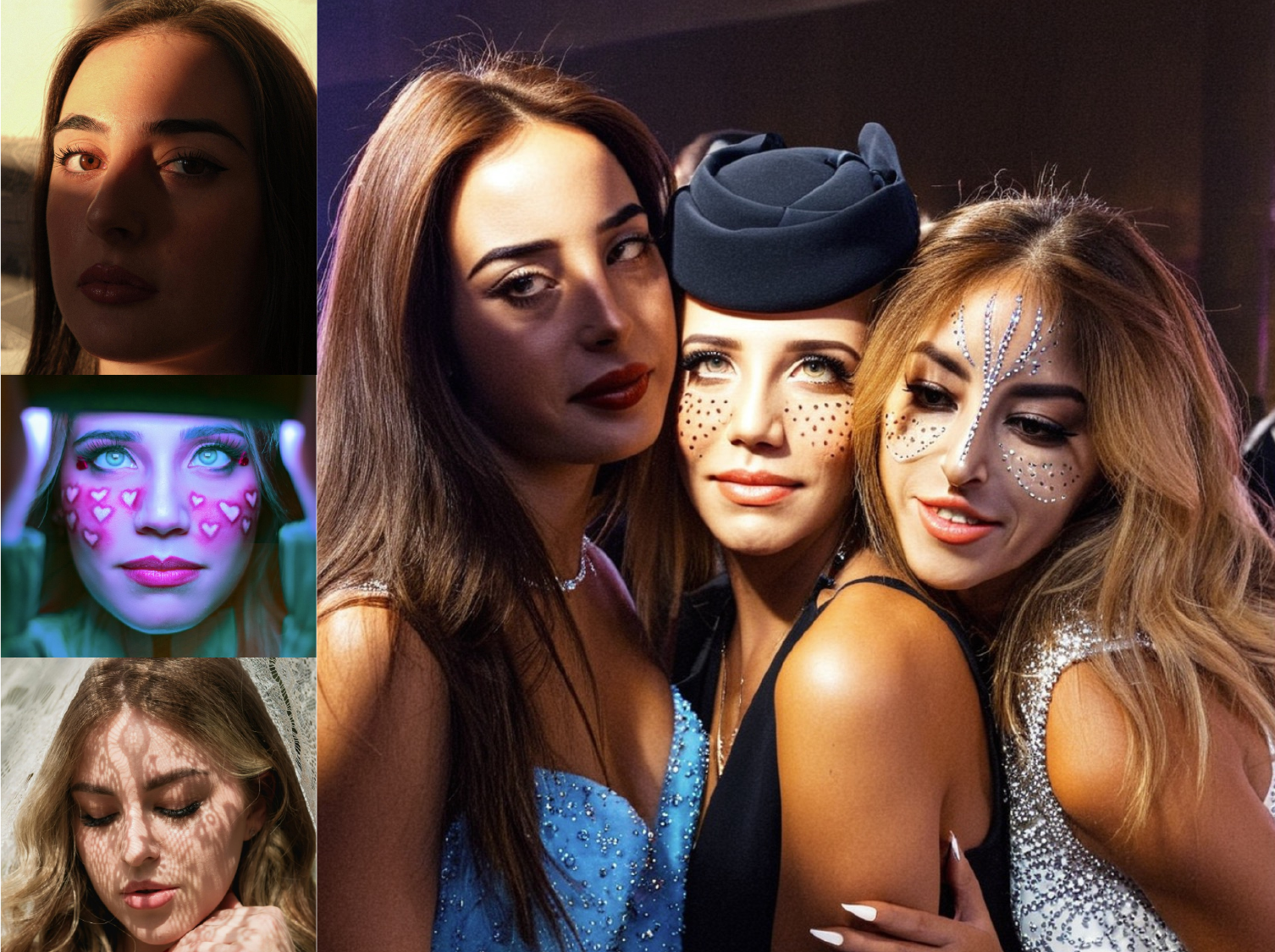} %
\caption{Over-fitting input}
\end{subfigure}
\vspace{-3mm}
\caption{Limitations. (a) Abnormal anatomy can be found in complex scenes, e.g. the forearms in the red box. (b) When input faces contain strong shadows, our method struggles to fully disentangle user's ID in generated result.}
\label{fig:fail}
\end{center}
\vspace*{-8mm}
\end{figure}

As illustrated in \textbf{Fig.\,\ref{fig:fail}}, our method encounters two primary limitations. First, our implementation is based on the SDXL model, which occasionally produces inaccuracies in body and finger anatomy. Transitioning to a more advanced base model could potentially rectify this issue. Second, facial features in the current model may incorporate lighting and expression details that are not fully separated from identity information, potentially compromising the generation quality. Enriching our training dataset with multiple images of the same identities, showcasing varied expressions and lighting conditions, might help alleviate this problem.

Although the resemblance has been improved with our approach, there still exists a performance gap between generating a single person and multiple people (\textbf{Fig.\,\ref{fig:compare_n_face_all} (a)}). This means generating personalized group photo remains a challenging problem and requires further study.

\section{Conclusion}
\label{sec:conclusion}
In conclusion, our novel method, ID-Patch, markedly enhances identity resemblance and positioning in group photo generation. By embedding each identity feature within a distinct patch and utilizing ControlNet to accurately place each identity at the designated spatial locations, we successfully reduce ID leakage. Our approach seamlessly integrates with additional conditioning signals such as pose control. Experimental results underscore the superiority of ID-Patch, particularly in complex scenarios involving more than three identities. This work paves the way for future explorations in multi-ID image generation. Potential future research directions include leveraging multiple images of the same individual from different angles to further enhance identity resemblance and simultaneously controlling location and facial expressions using the patch technique.

\noindent\textbf{Societal Impact:} 
Our work studies ID injection from a technical perspective and is not intended for malicious use. However, inappropriate usage might generate undesired fake images. Users should be aware of the ethical implications and use it responsibly.

{
    \small
    \bibliographystyle{ieeenat_fullname}
    \bibliography{bib/main, bib/diffusion}
}

\clearpage
\setcounter{page}{1}
\maketitlesupplementary

\setcounter{section}{0}


\setcounter{section}{0}
\setcounter{figure}{0}
\makeatletter 
\renewcommand{\thefigure}{A\arabic{figure}}
\renewcommand{\theHfigure}{A\arabic{figure}}
\renewcommand{\thetable}{A\arabic{table}}
\renewcommand{\theHtable}{A\arabic{table}}

\makeatother
\setcounter{table}{0}

\setcounter{mylemma}{0}
\renewcommand{\themylemma}{A\arabic{mylemma}}
\setcounter{equation}{0}
\renewcommand{\theequation}{A\arabic{equation}}

\begin{figure}[t]
    \centering
    \includegraphics[width=\linewidth]{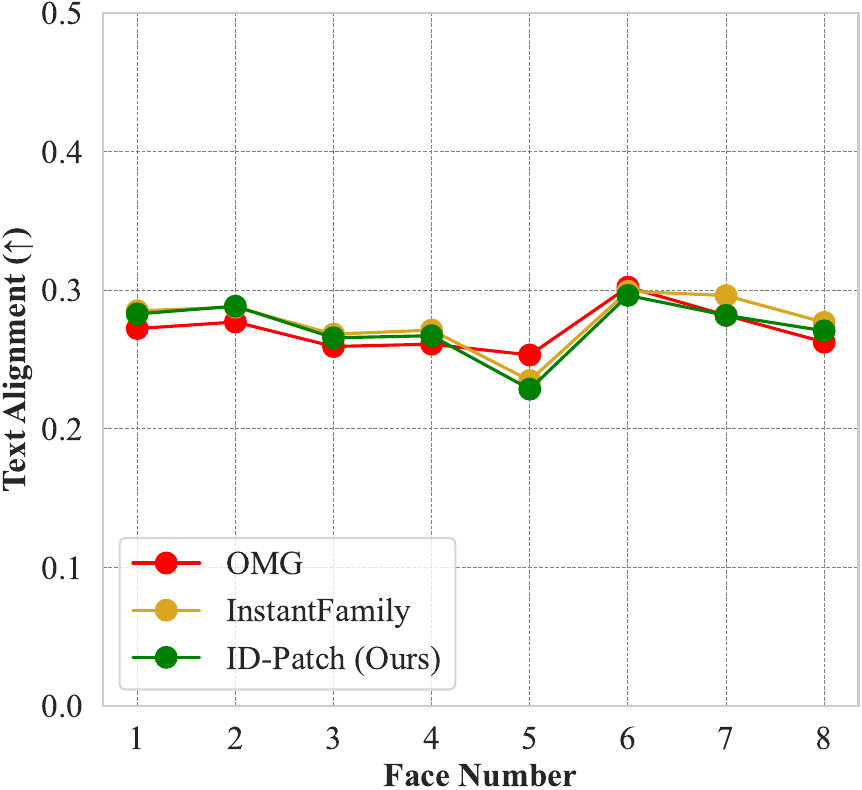}
    \caption{Performance evaluation of different model generations from the perspective of the text alignment score. There is no significant trend or difference between different number of faces. All methods achieve similar scores.}
    \label{fig:text_n_face_all}
\end{figure}

\section{Additional Experiment Setups}
\label{appx: additional_exp}

\subsection{Training Dataset} 

Our training set comprises a robust collection of images, totaling 17 million single-person and 1.95 million multi-person instances. These images are sourced from both purchased data and publicly available sources. Each image has been carefully cropped and resized according to predefined specifications and subsequently organized into distinct data buckets based on shape categories. BLIP-2~\cite{li2023blip} is employed to obtain image captions.

Face localization in images is performed using the MTCNN framework \cite{zhang2016joint}. For generating the pose conditions, we employ the HRNet-DEKR model \cite{geng2021bottom} for pose estimation. To enhance the accuracy of this estimation, we compute the average distance between facial keypoints detected by MTCNN (considered as ground truth) and those estimated by HRNet-DEKR. Poses with a large keypoint distance are filtered out to ensure precision.

\section{Additional Quantitive Results}
\label{appx: additional_details}
Details regarding the text alignment scores with varying numbers of faces are illustrated in \textbf{Fig.\,\ref{fig:text_n_face_all}}. It is evident that all three methods exhibit comparable performance in terms of text alignment. Beside, we do not observe a significant change of the text alignment score across different number of faces.

\section{Additional Visualizations}
\label{appx: additional_visual}
\begin{figure*}[t]
  \centering
  \resizebox{0.9\textwidth}{!}{
  \begin{tabular}{c|c|c}
  \toprule[1pt]
\midrule
      \multirow{1}[4]{*}[0.8cm]{
\rotatebox{0}{ 
 \scriptsize{\textbf{ID + ID Location}} }
    } 
    &
    \begin{minipage}{0.14\textwidth}\centering\includegraphics[width=\linewidth]{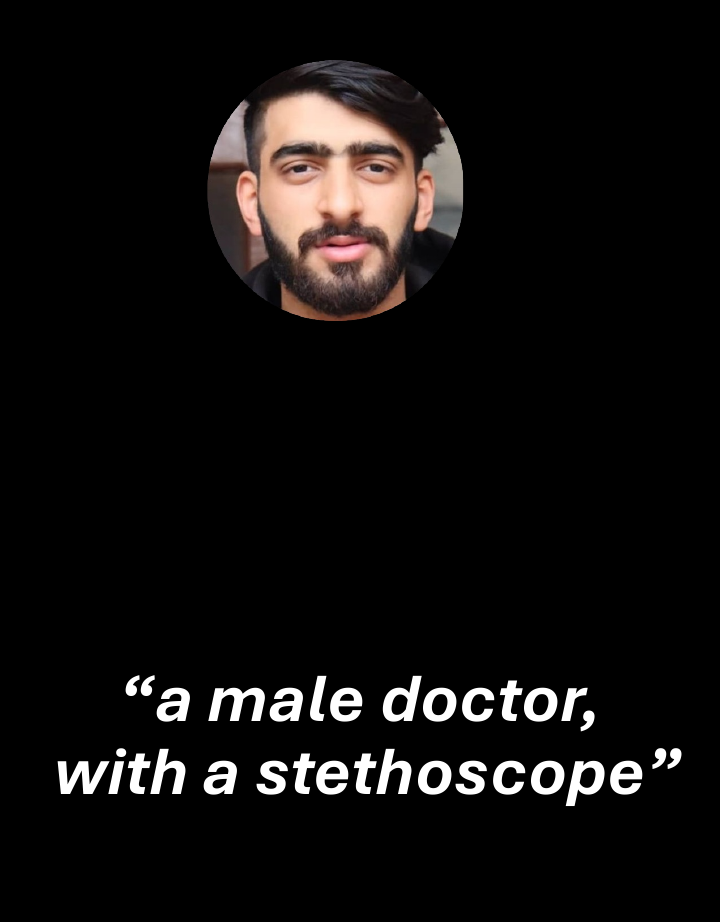}\end{minipage} 
    \vspace*{1mm} 
     &
    \begin{minipage}{0.18\textwidth}\centering\includegraphics[width=\linewidth]{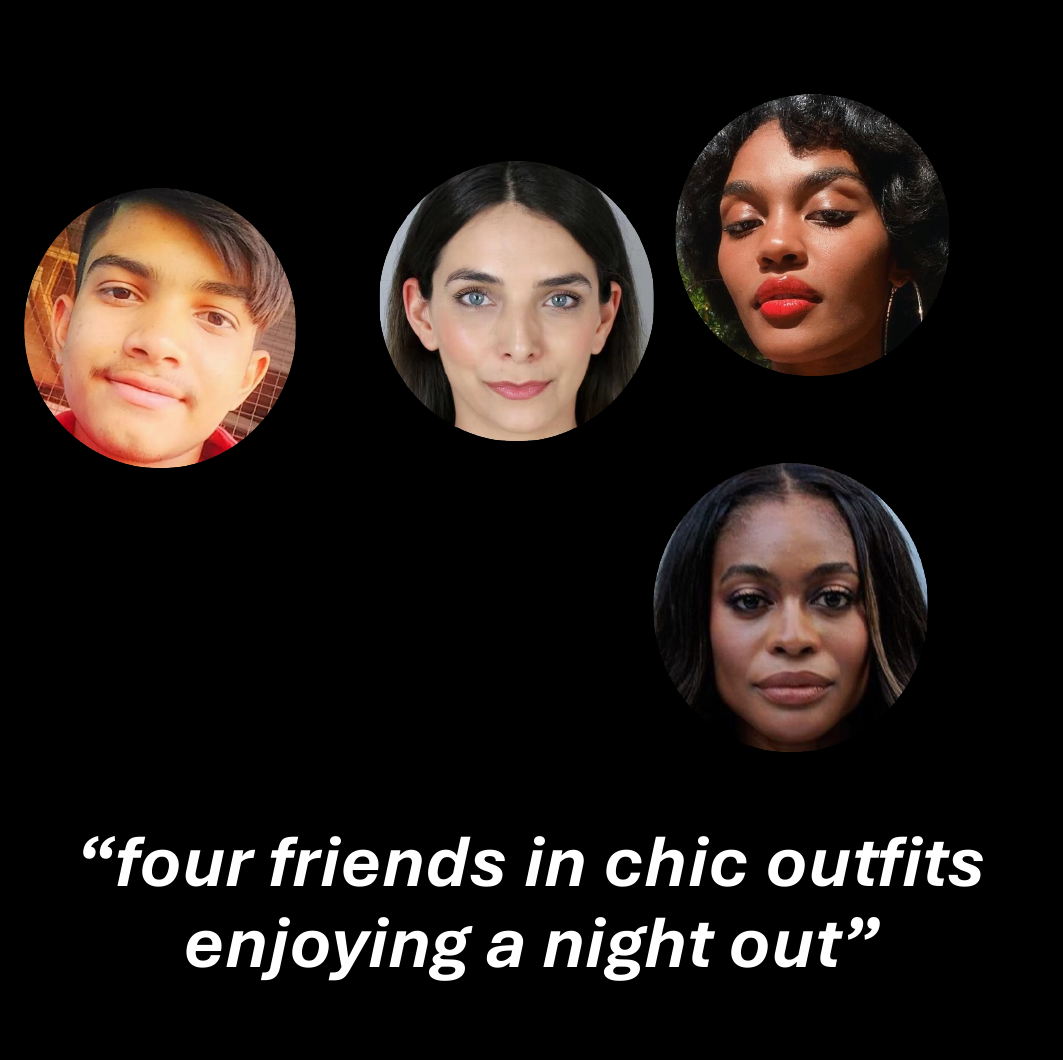}\end{minipage}
    
    \\
\midrule
\multirow{1}[4]{*}[0.8cm]{
\rotatebox{0}{ 
 \scriptsize{\textbf{Generation Result} } }
    } 
    &
    \begin{minipage}{0.14\textwidth}\centering\includegraphics[width=\linewidth]{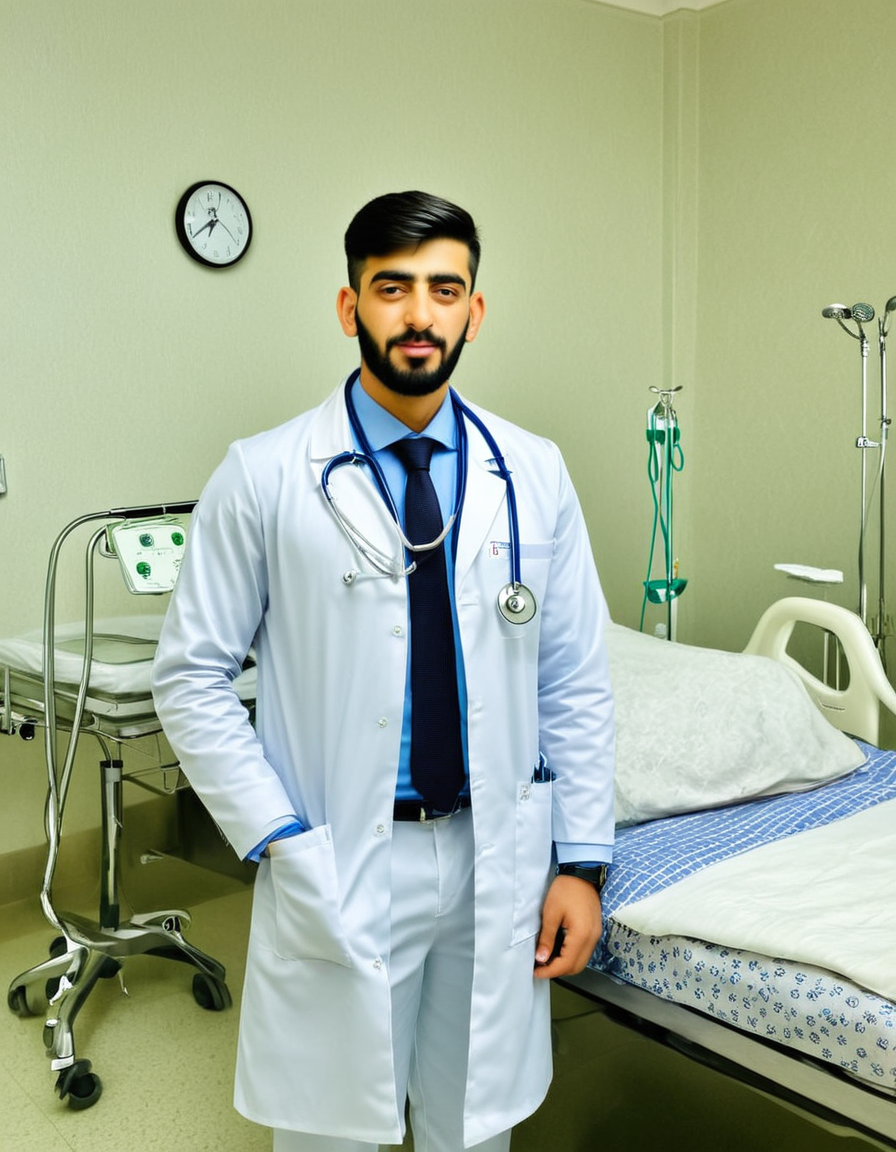}\end{minipage} 
    \vspace*{1mm} 
     &
    \begin{minipage}{0.18\textwidth}\centering\includegraphics[width=\linewidth]{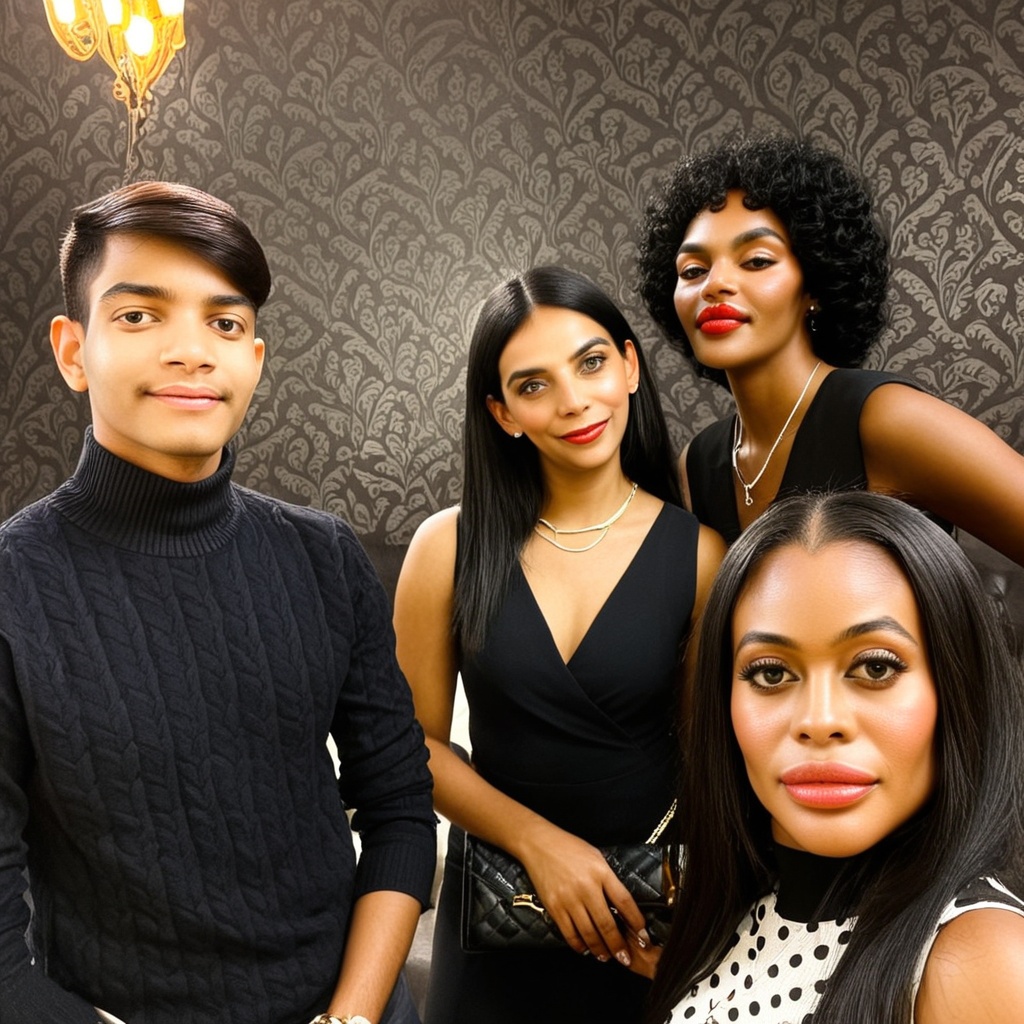}\end{minipage}
    \\
\midrule
      \multirow{1}[4]{*}[0.8cm]{
\rotatebox{0}{ 
 \scriptsize{\textbf{ID + ID Location}} }
    } 
    &
    \begin{minipage}{0.18\textwidth}\centering\includegraphics[width=\linewidth]{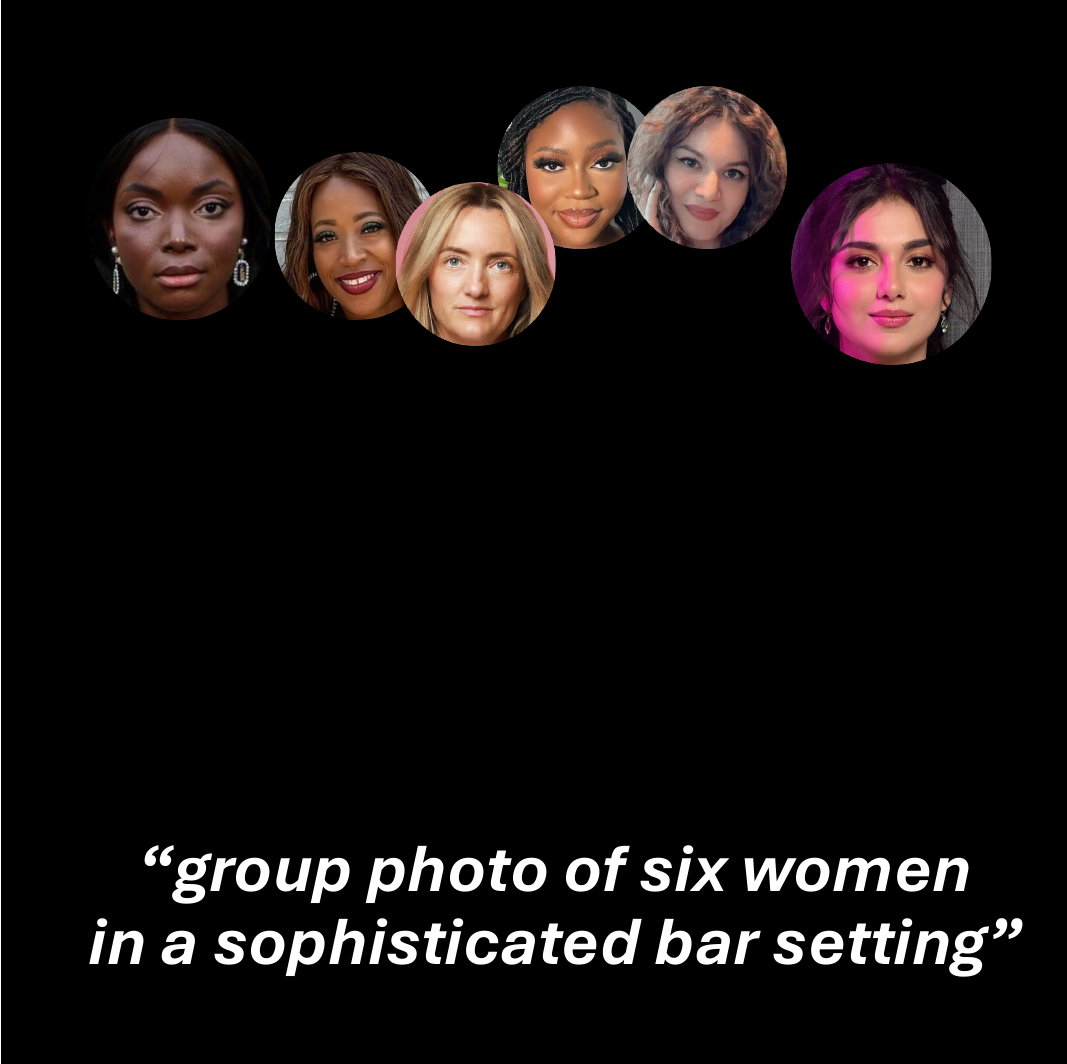}\end{minipage}
    &
    \begin{minipage}{0.19\textwidth}\centering\includegraphics[width=\linewidth]{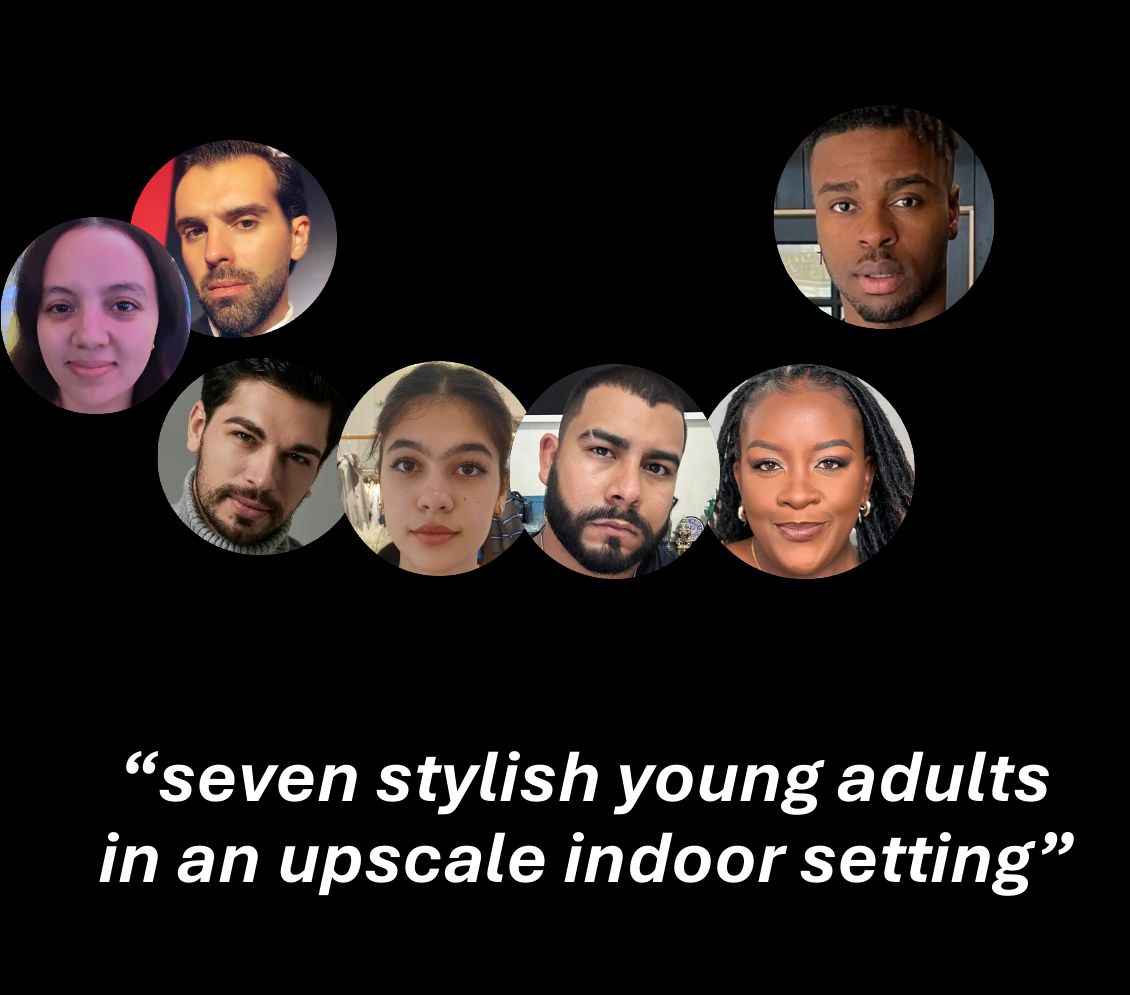}\end{minipage}
    \\
    \midrule
      \multirow{1}[4]{*}[0.8cm]{
\rotatebox{0}{ 
 \scriptsize{\textbf{Generation Result}} }
    } 
    &
     \begin{minipage}{0.18\textwidth}\centering\includegraphics[width=\linewidth]{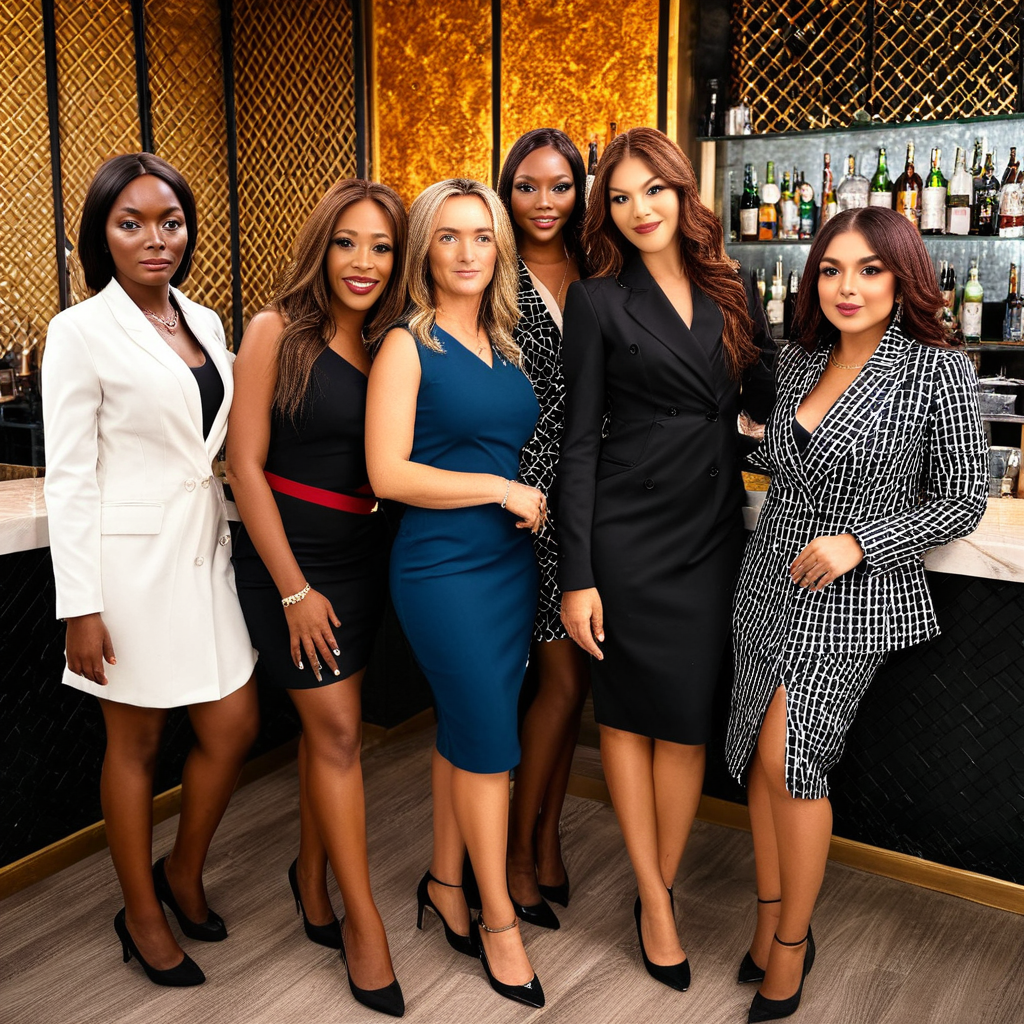}\end{minipage}
    &
    \begin{minipage}{0.19\textwidth}\centering\includegraphics[width=\linewidth]{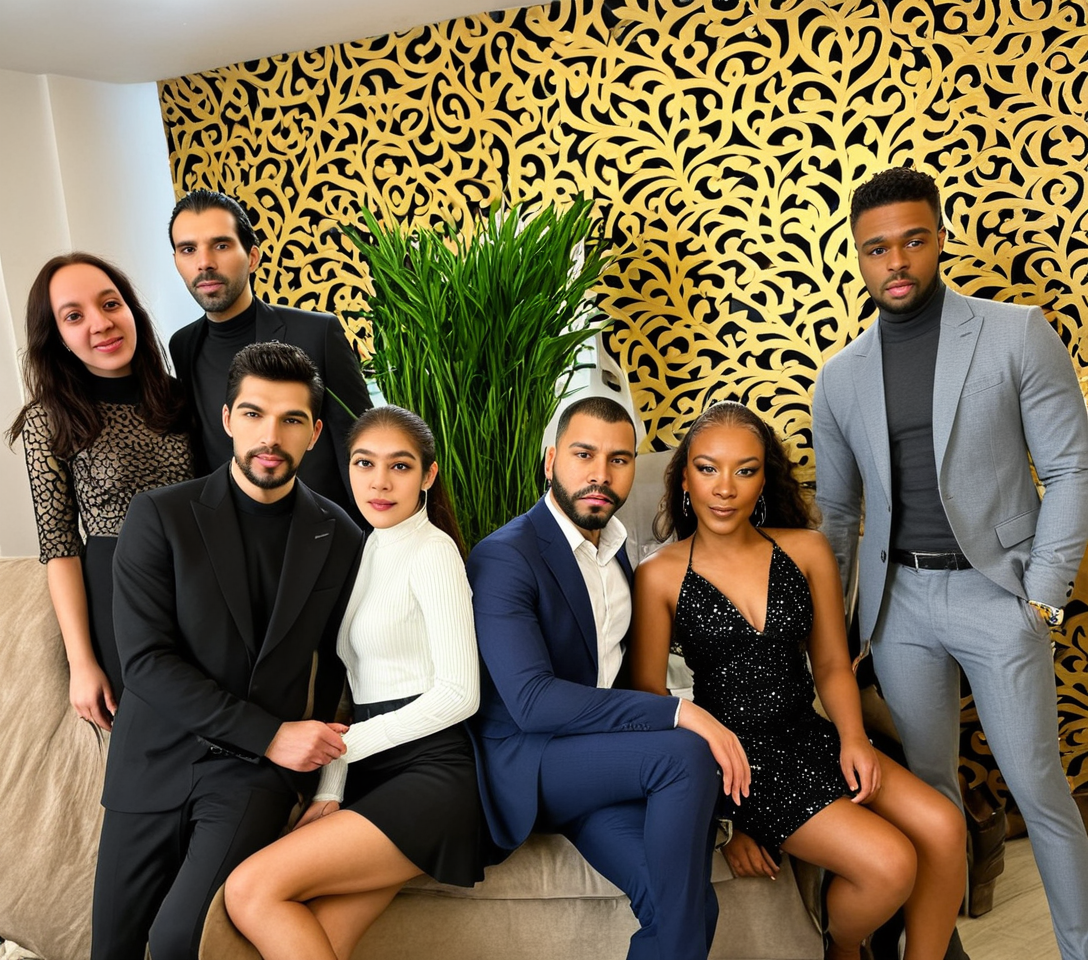}\end{minipage}
    \\
  \bottomrule[1pt]
\end{tabular} 

  }
  \vspace{-1.5mm}
  \caption{Additional visualizations of ID-Patch pose-free generation. Here the condition image is only ID patches rendered onto a black canvas without any pose condition. 
  }
  \label{fig:nopose_supp}
  \vspace*{-3mm}
\end{figure*}

Additional pose-free generated images of our proposed ID-Patch can be found in \textbf{Fig.\,\ref{fig:nopose_supp}}, where the generation is conditioned on individual face locations while corresponding head sizes and body poses are inferred implicitly. Our method can accommodate a large number of people with diverse ethnic backgrounds, yet generating visually appealing group photo results. 

\begin{figure*}[t]
  \centering
  \resizebox{\textwidth}{!}{
  \begin{tabular}{c|c|c|c}
  \toprule[1pt]
  \midrule
  \multicolumn{1}{c|}{\scriptsize{\textbf{ID + Pose}}}  & \multicolumn{1}{c|}{\scriptsize{\textbf{OMG}}}
 & \multicolumn{1}{c|}{\scriptsize{\textbf{InstantFamily}}}
 & \multicolumn{1}{c}{\scriptsize{\textbf{ID-Patch (Ours)}}}
   \\
\midrule

    \begin{minipage}{0.18\textwidth}\centering\includegraphics[width=\linewidth]{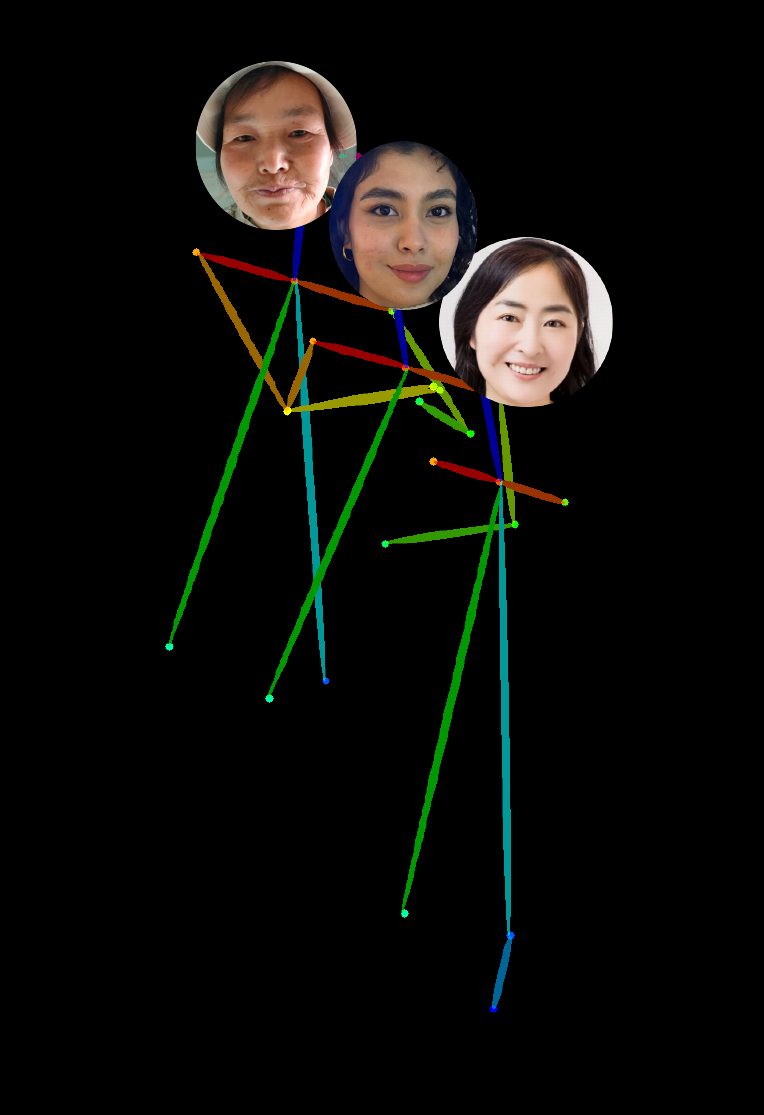}\end{minipage} 
    \vspace*{1mm} 
     &
    \begin{minipage}{0.18\textwidth}\centering\includegraphics[width=\linewidth]{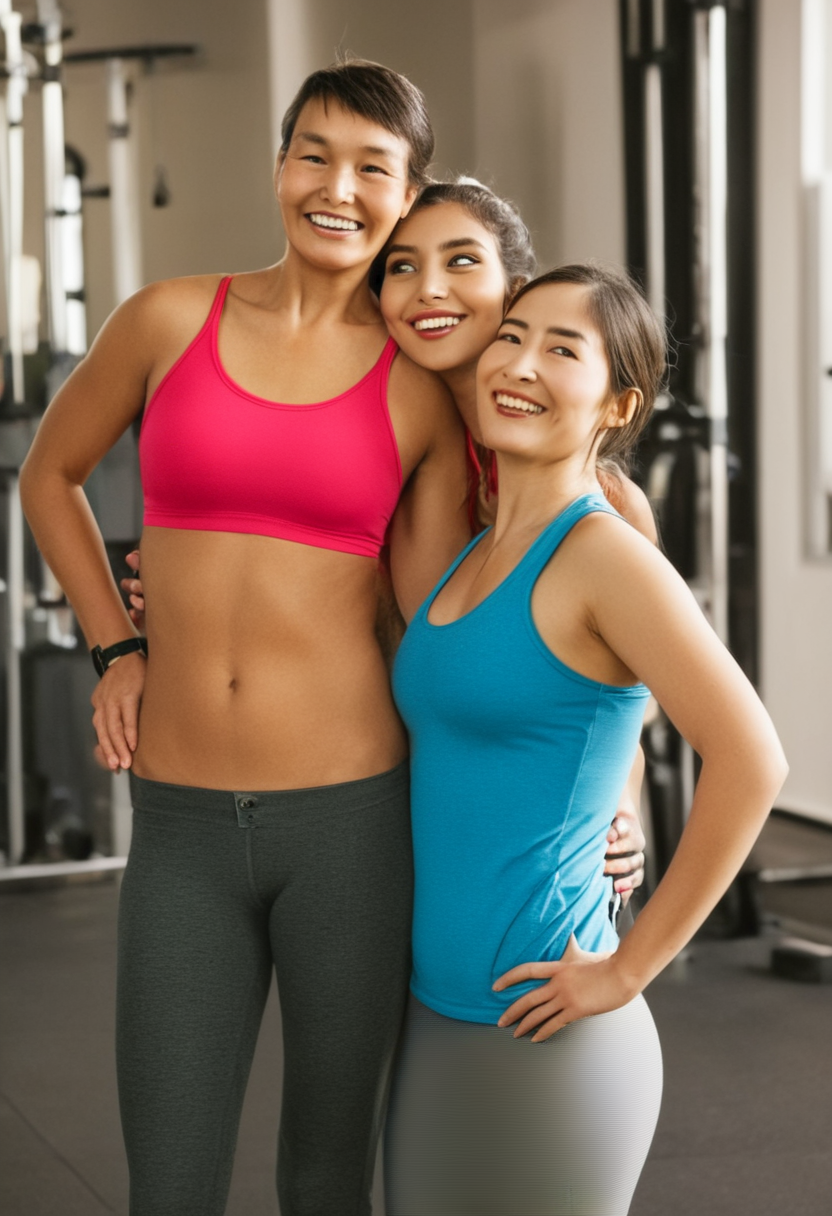}\end{minipage}
    &
    \begin{minipage}{0.18\textwidth}\centering\includegraphics[width=\linewidth]{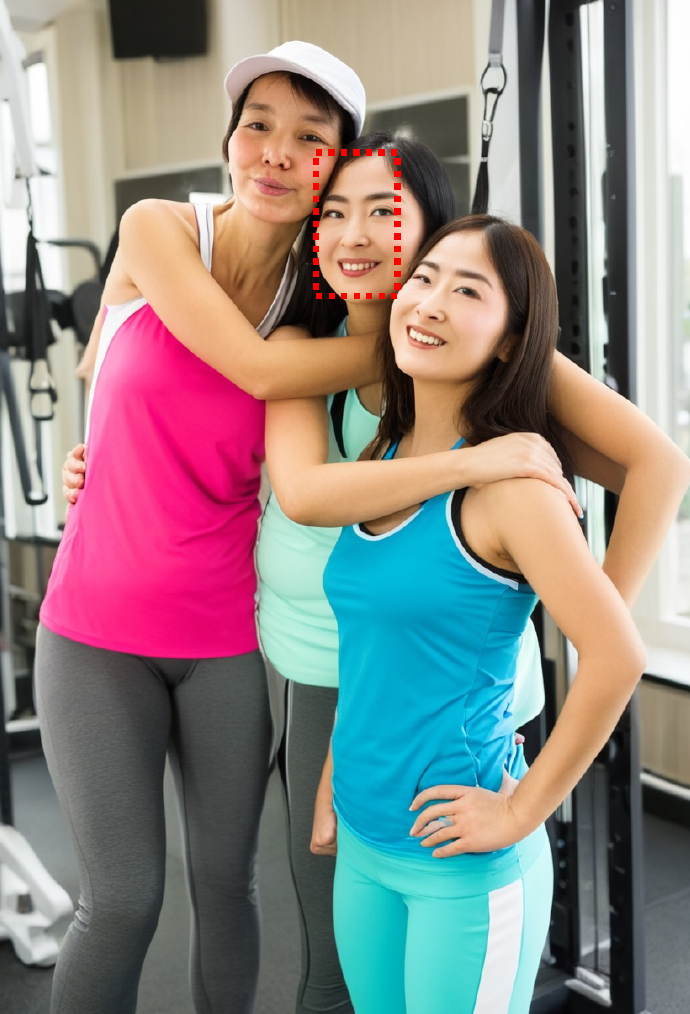}\end{minipage}
    &
    \begin{minipage}{0.18\textwidth}\centering\includegraphics[width=\linewidth]{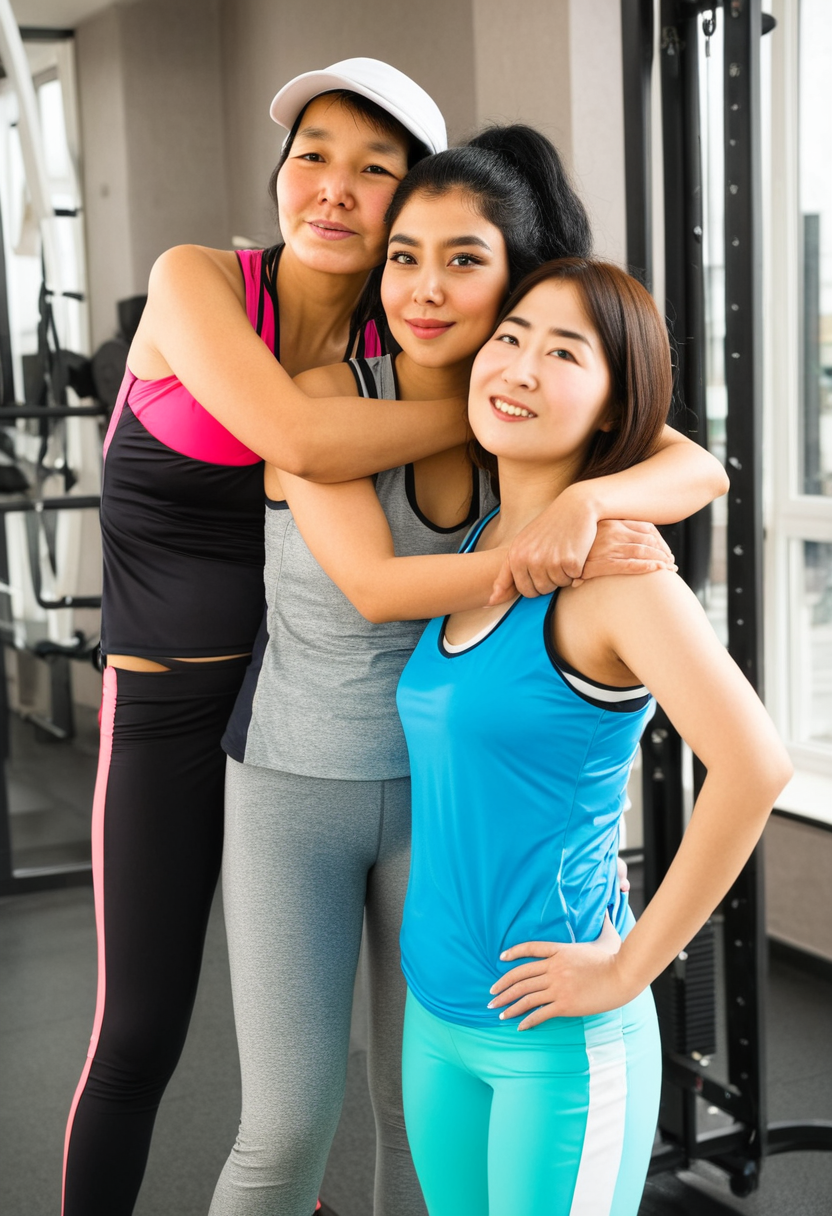}\end{minipage}
    
    \\
\midrule

    \begin{minipage}{0.18\textwidth}\centering\includegraphics[width=\linewidth]{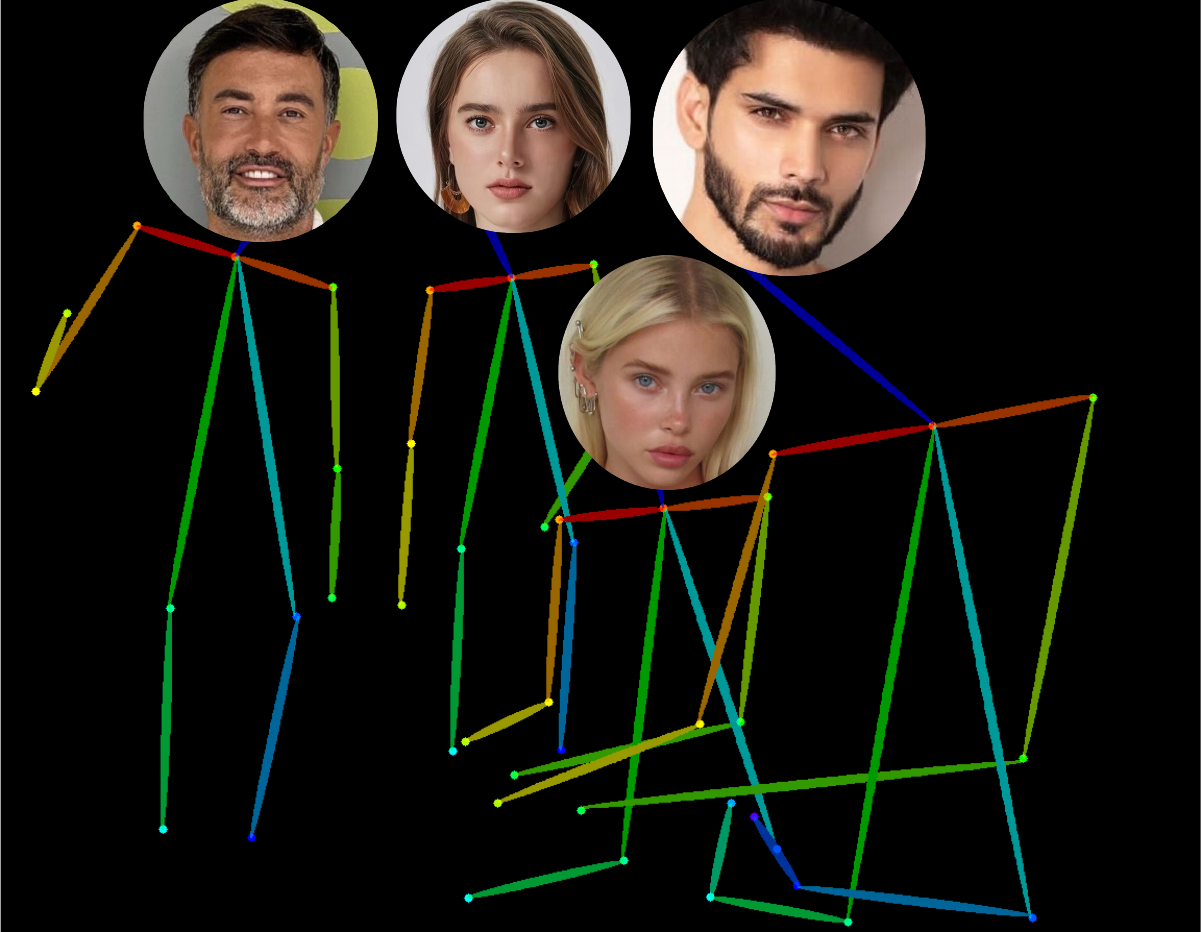}\end{minipage} 
    \vspace*{1mm} 
     &
    \begin{minipage}{0.18\textwidth}\centering\includegraphics[width=\linewidth]{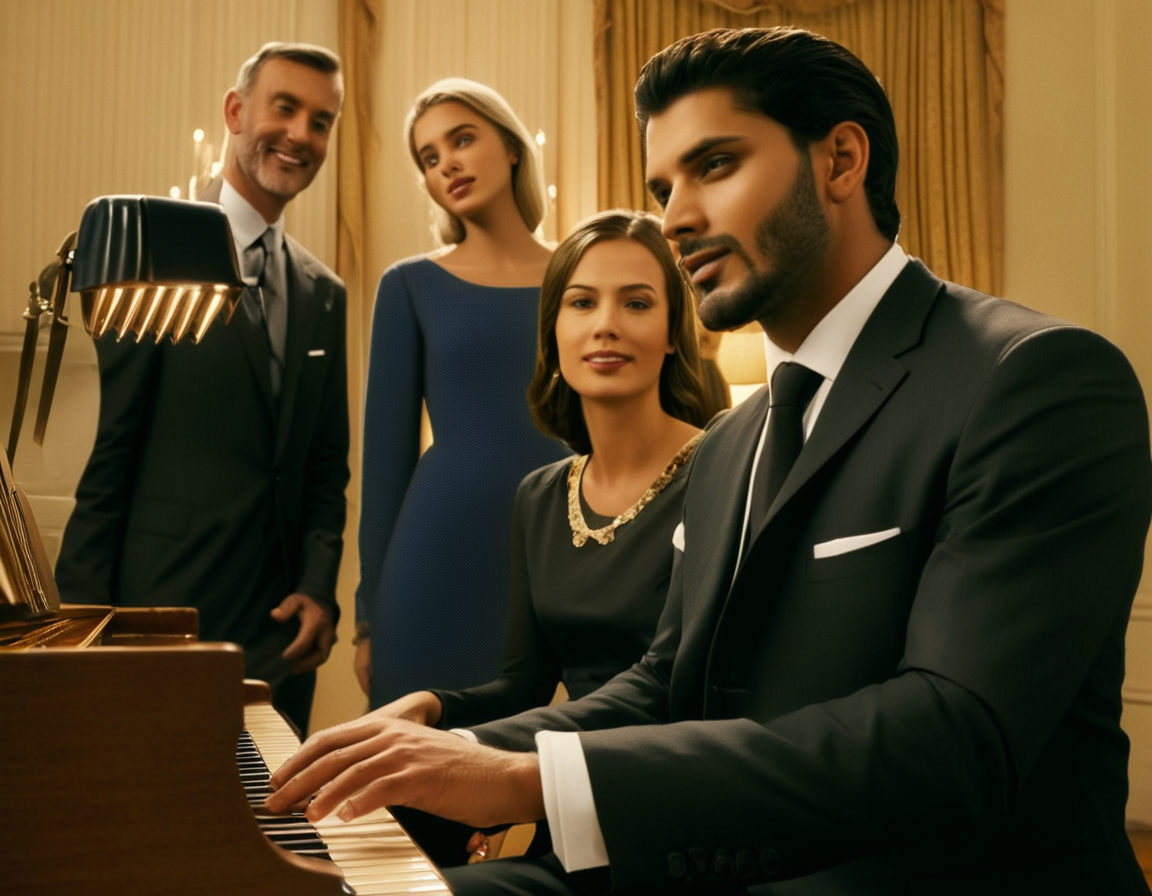}\end{minipage}
    &
    \begin{minipage}{0.18\textwidth}\centering\includegraphics[width=\linewidth]{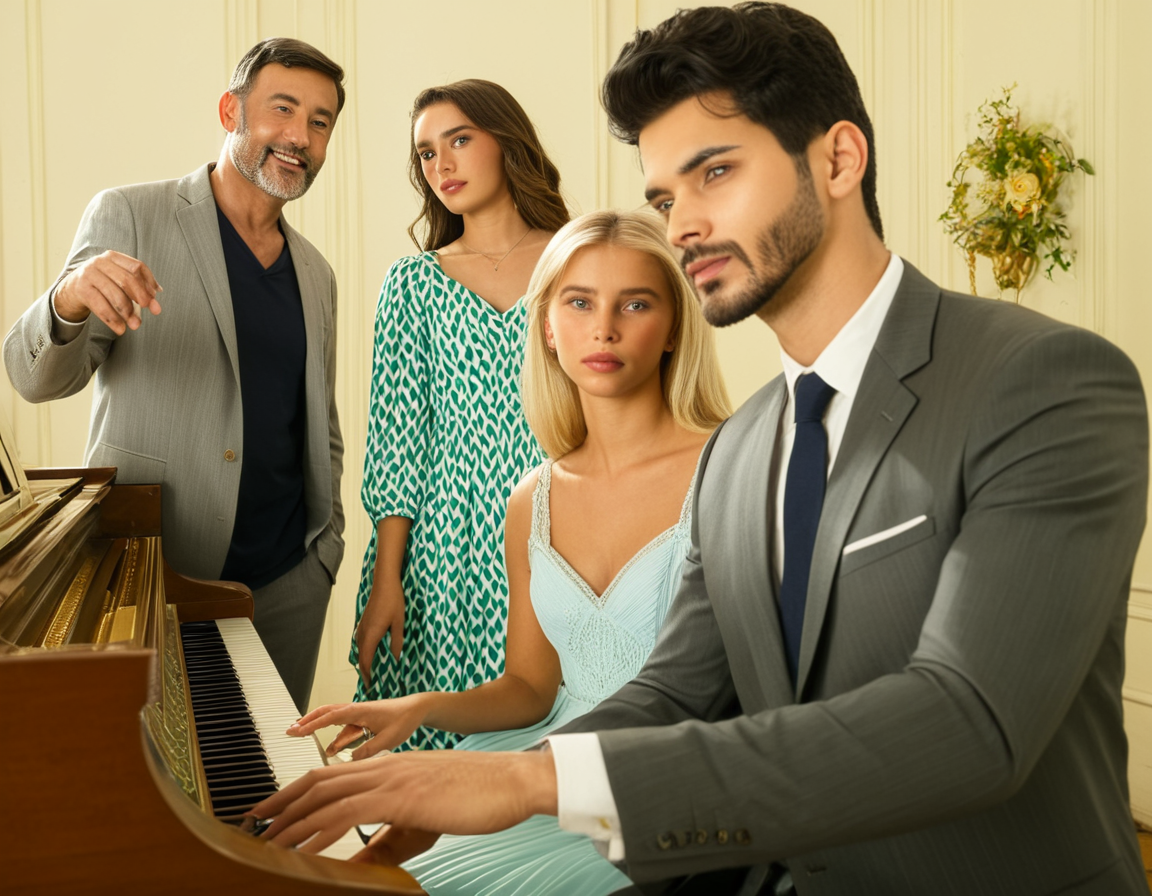}\end{minipage}
    &
    \begin{minipage}{0.18\textwidth}\centering\includegraphics[width=\linewidth]{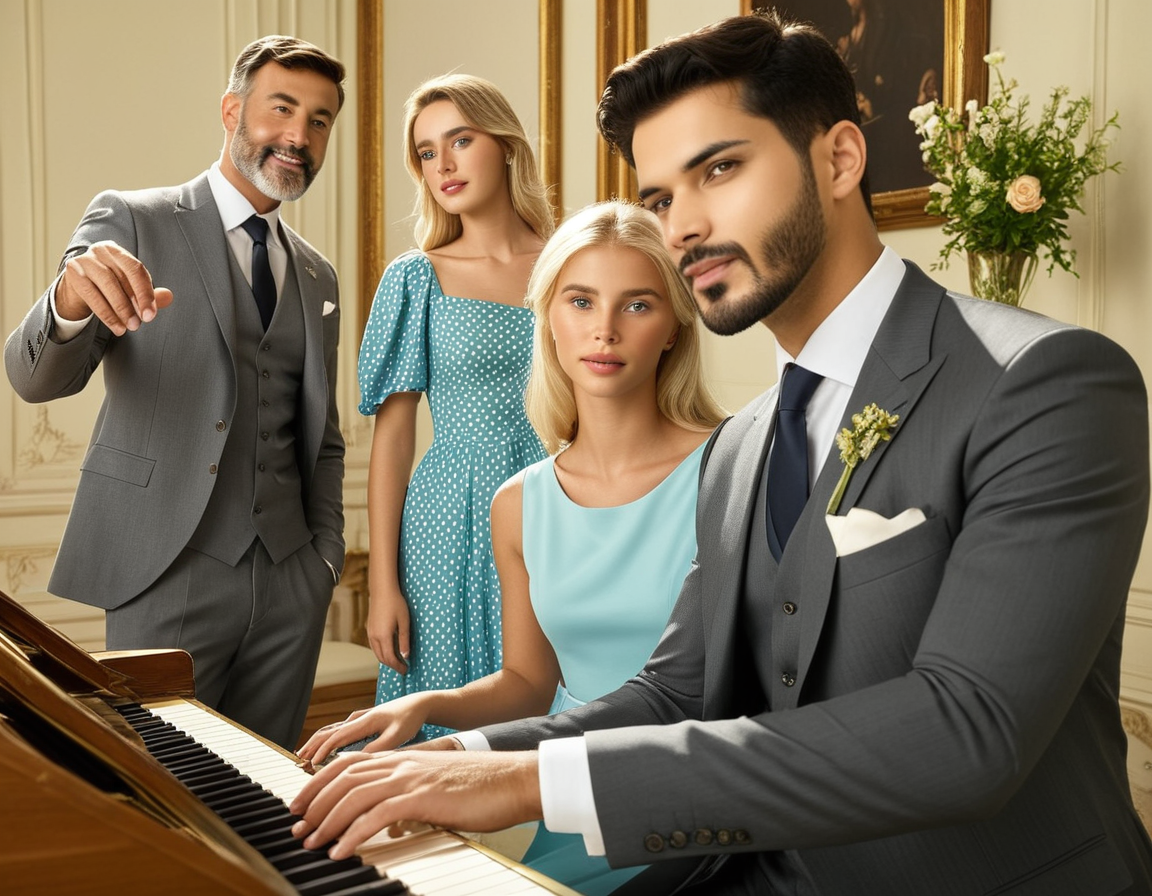}\end{minipage}
    \\
    \midrule

    \begin{minipage}{0.18\textwidth}\centering\includegraphics[width=\linewidth]{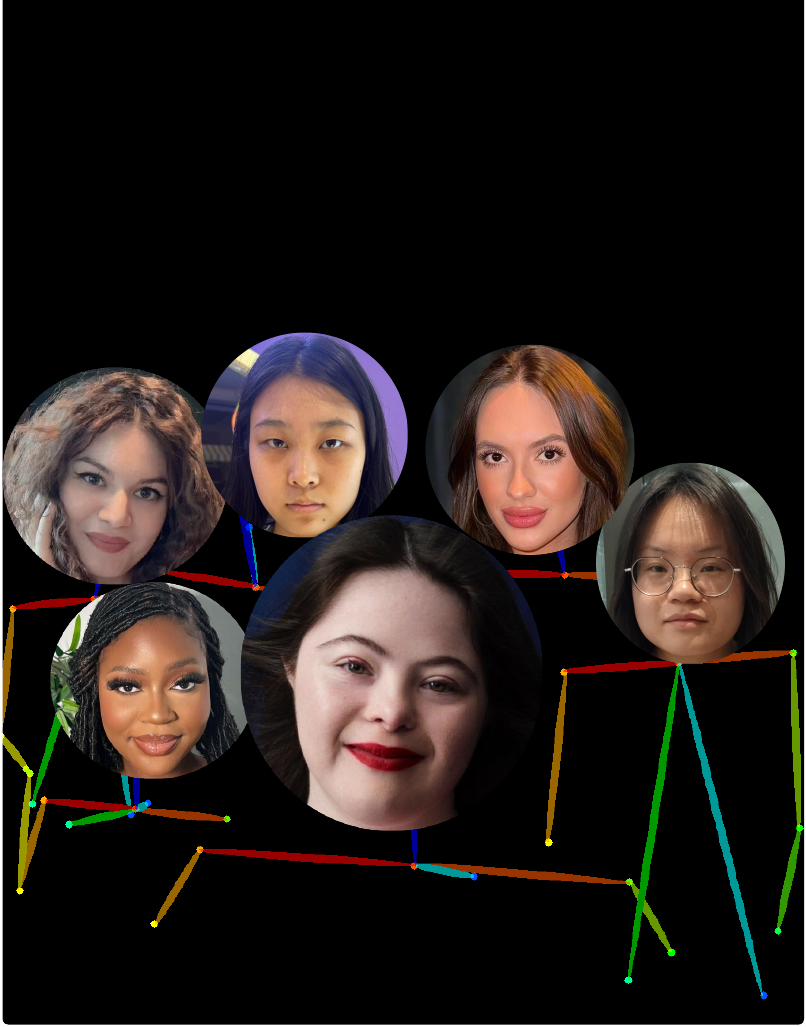}\end{minipage} 
    \vspace*{1mm} 
     &
    \begin{minipage}{0.18\textwidth}\centering\includegraphics[width=\linewidth]{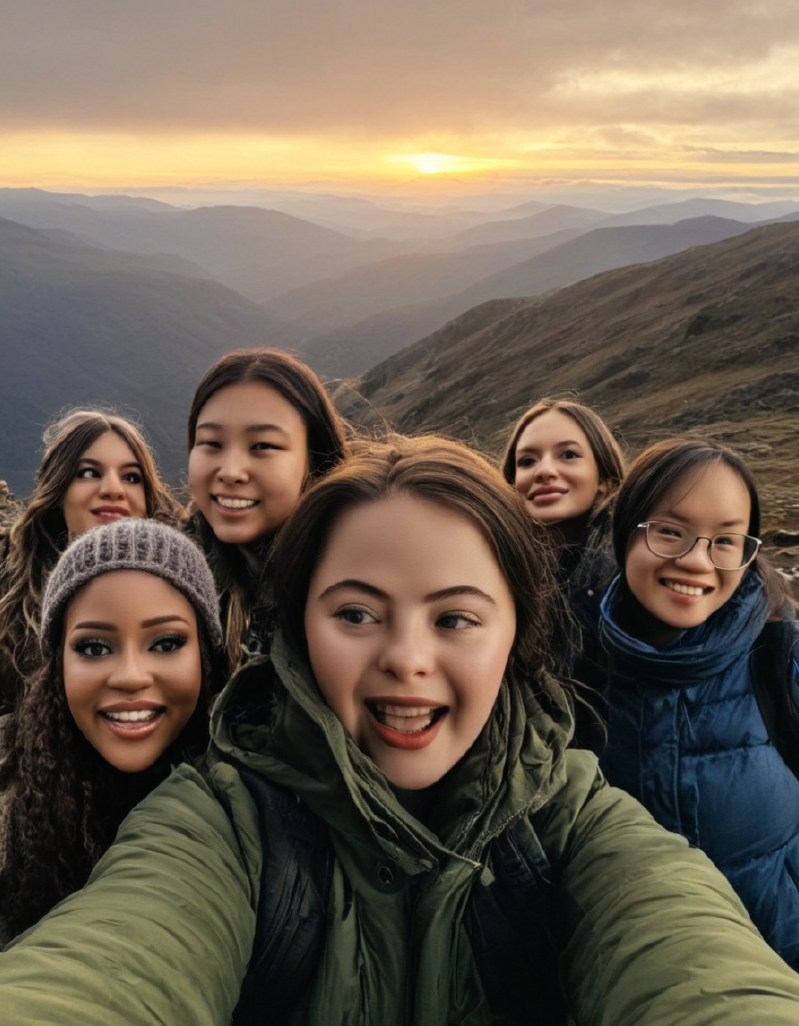}\end{minipage}
    &
    \begin{minipage}{0.18\textwidth}\centering\includegraphics[width=\linewidth]{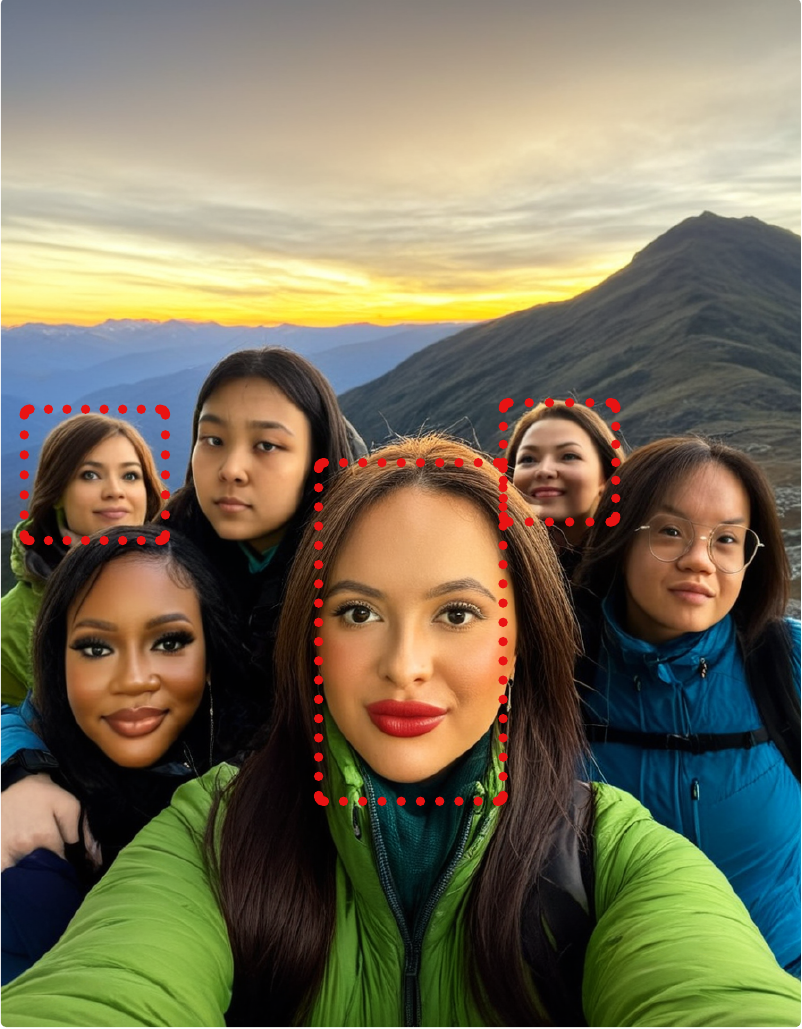}\end{minipage}
    &
    \begin{minipage}{0.18\textwidth}\centering\includegraphics[width=\linewidth]{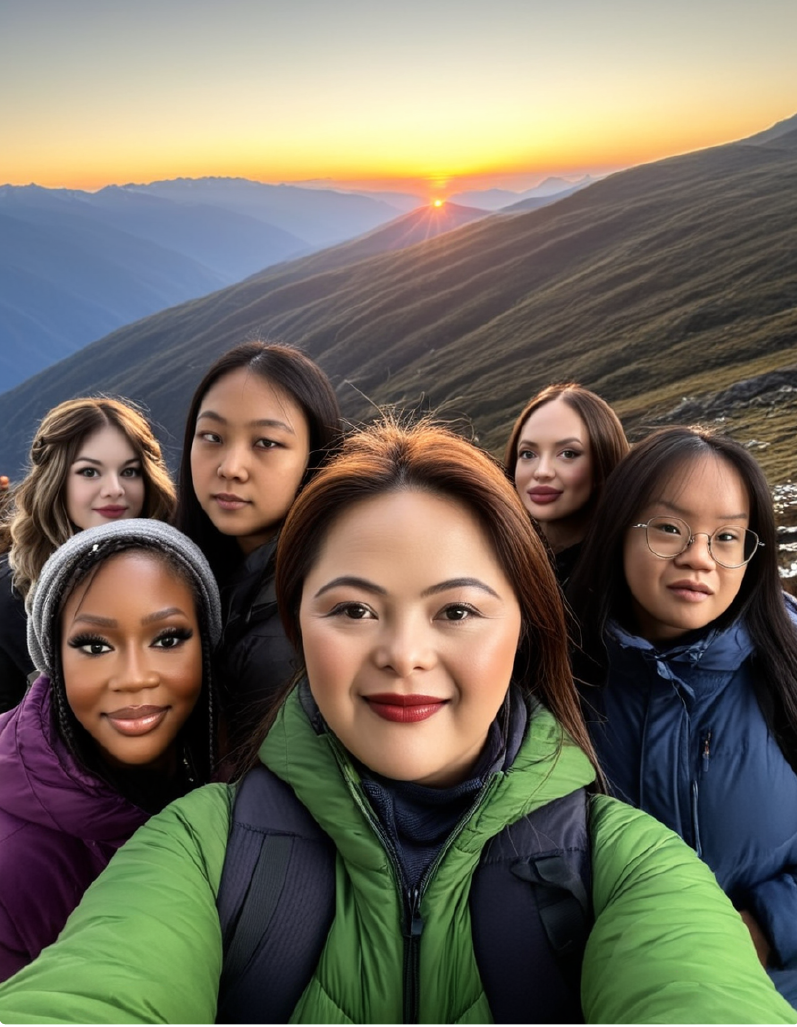}\end{minipage}
    \\
    \midrule

    \begin{minipage}{0.18\textwidth}\centering\includegraphics[width=\linewidth]{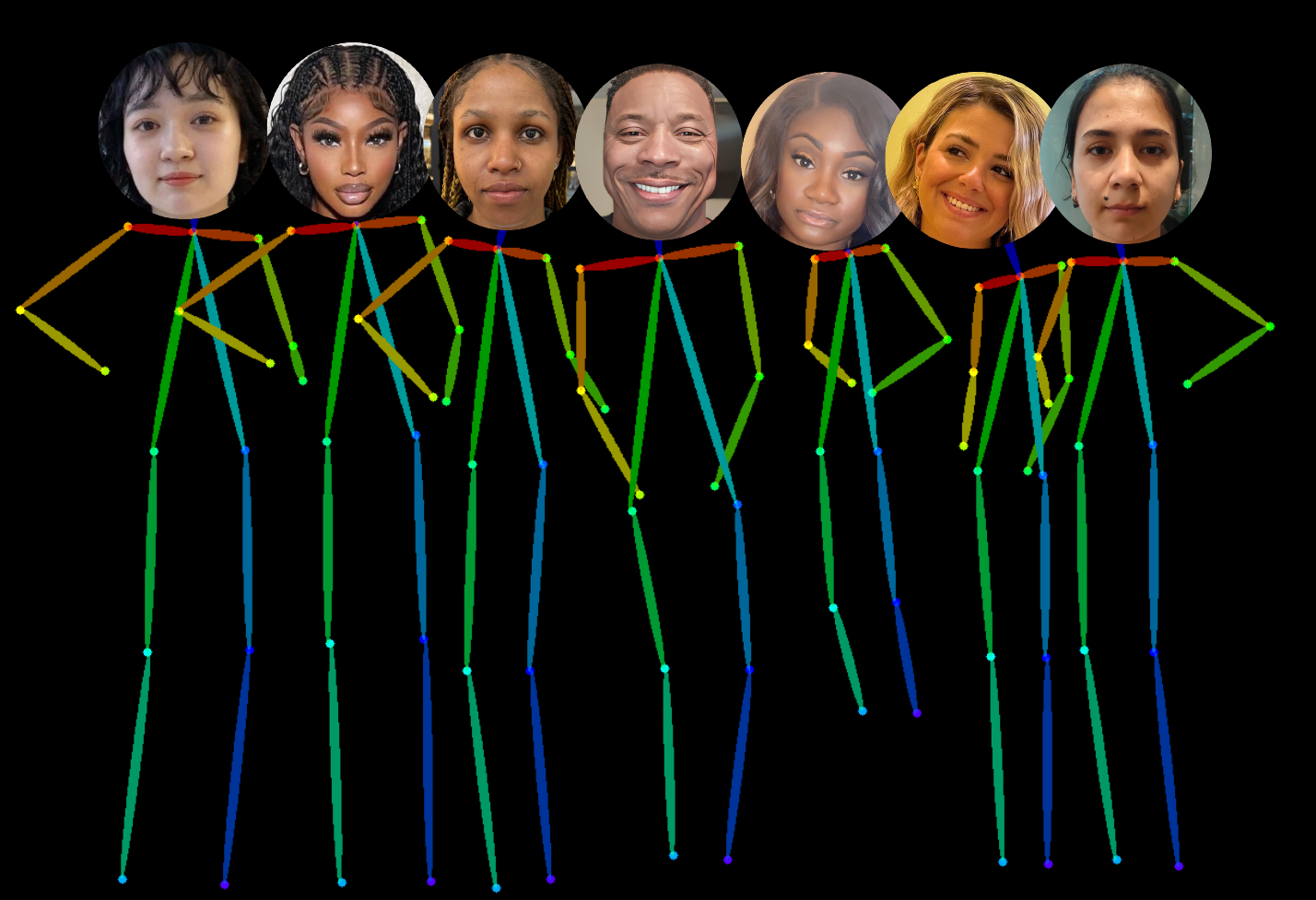}\end{minipage} 
    \vspace*{1mm} 
     &
    \begin{minipage}{0.18\textwidth}\centering\includegraphics[width=\linewidth]{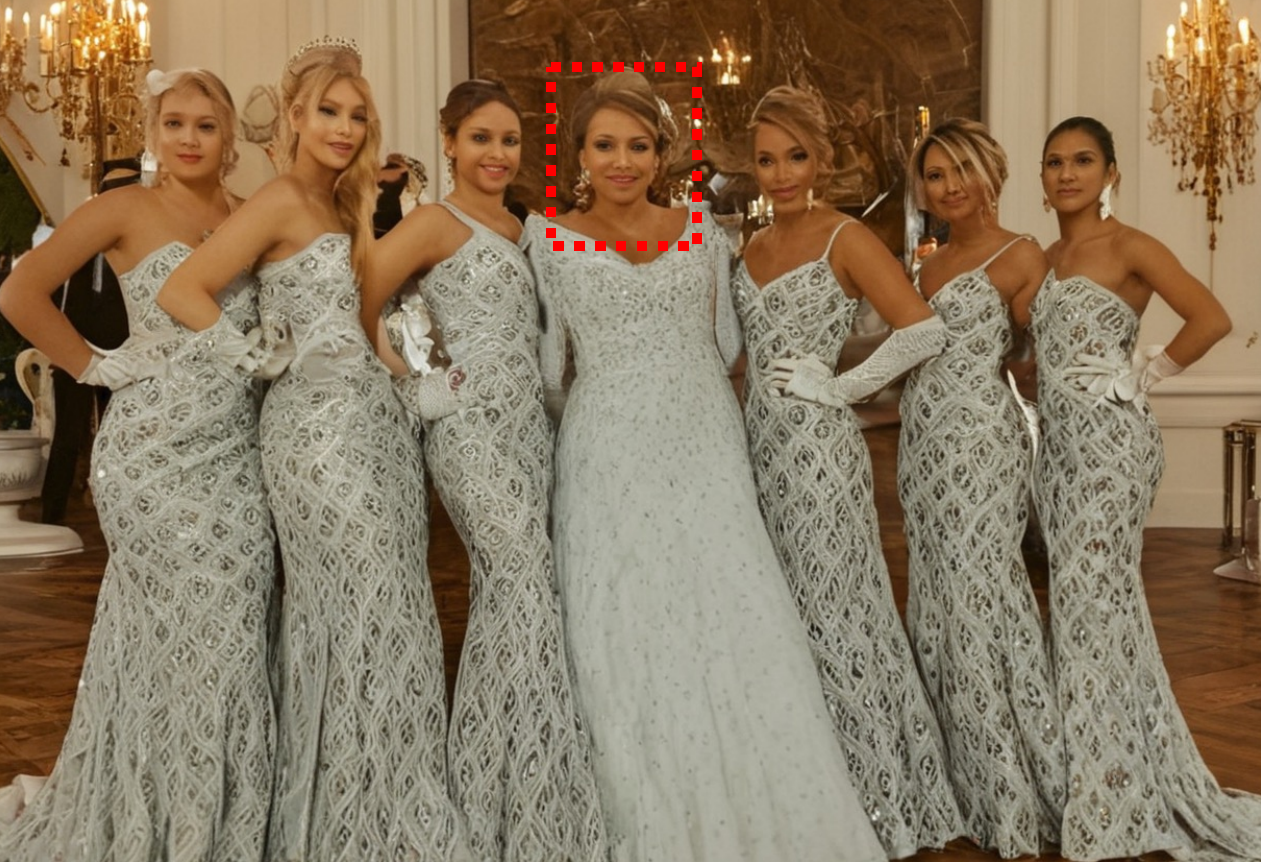}\end{minipage}
    &
    \begin{minipage}{0.18\textwidth}\centering\includegraphics[width=\linewidth]{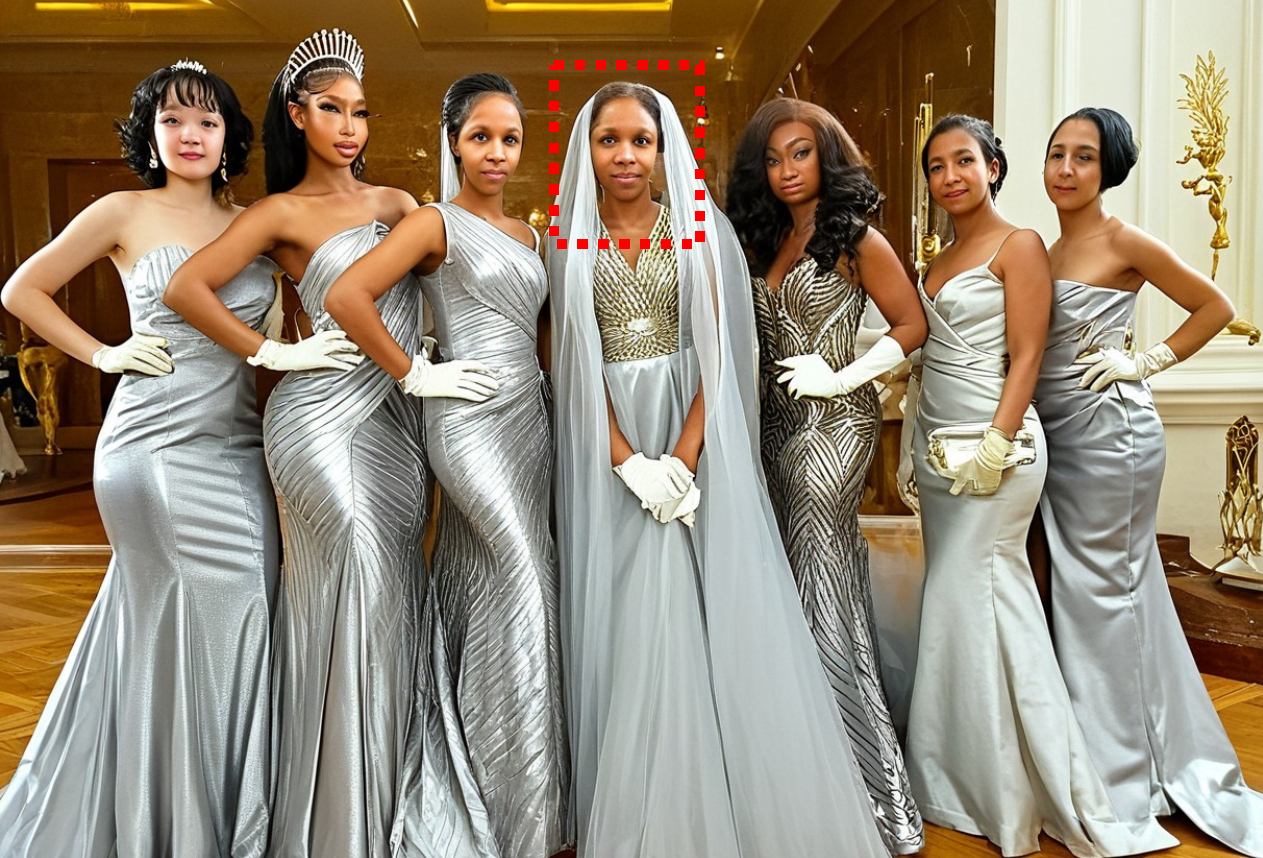}\end{minipage}
    &
    \begin{minipage}{0.18\textwidth}\centering\includegraphics[width=\linewidth]{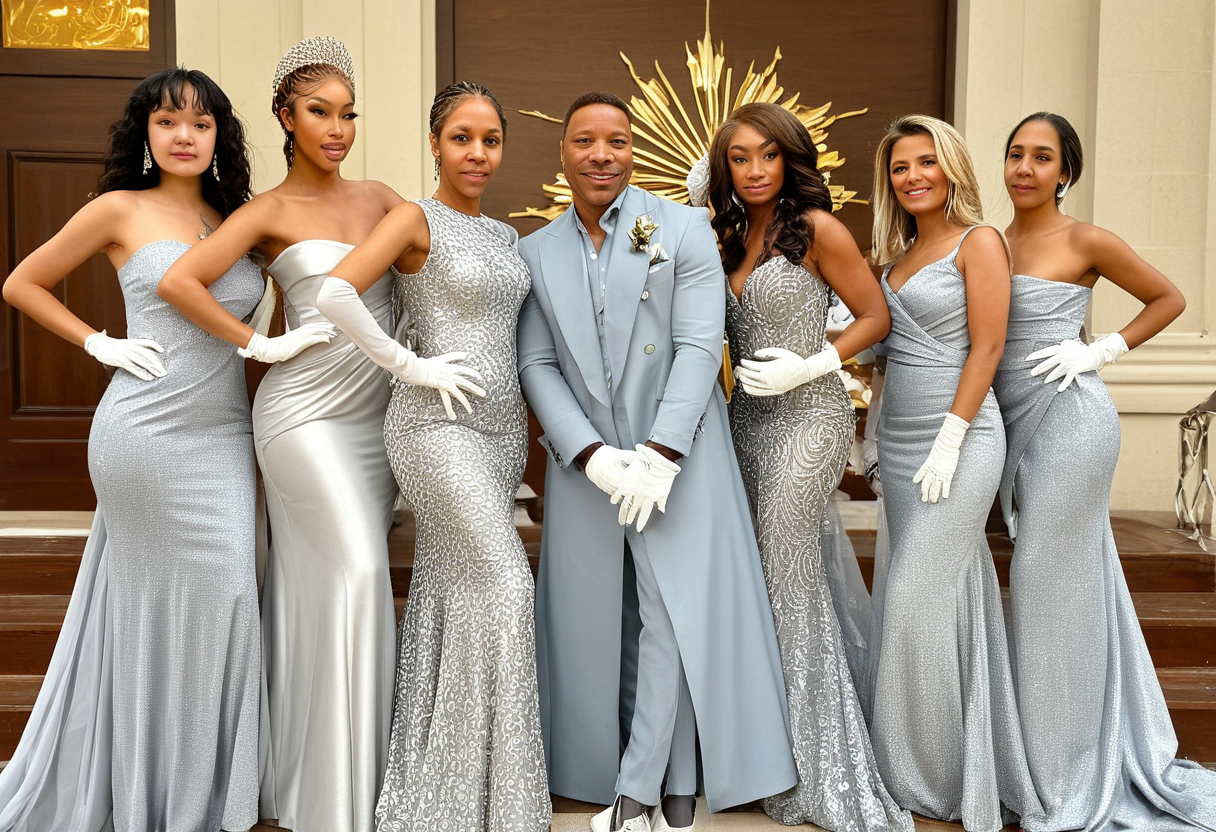}\end{minipage}
    \\
    \midrule

    \begin{minipage}{0.18\textwidth}\centering\includegraphics[width=\linewidth]{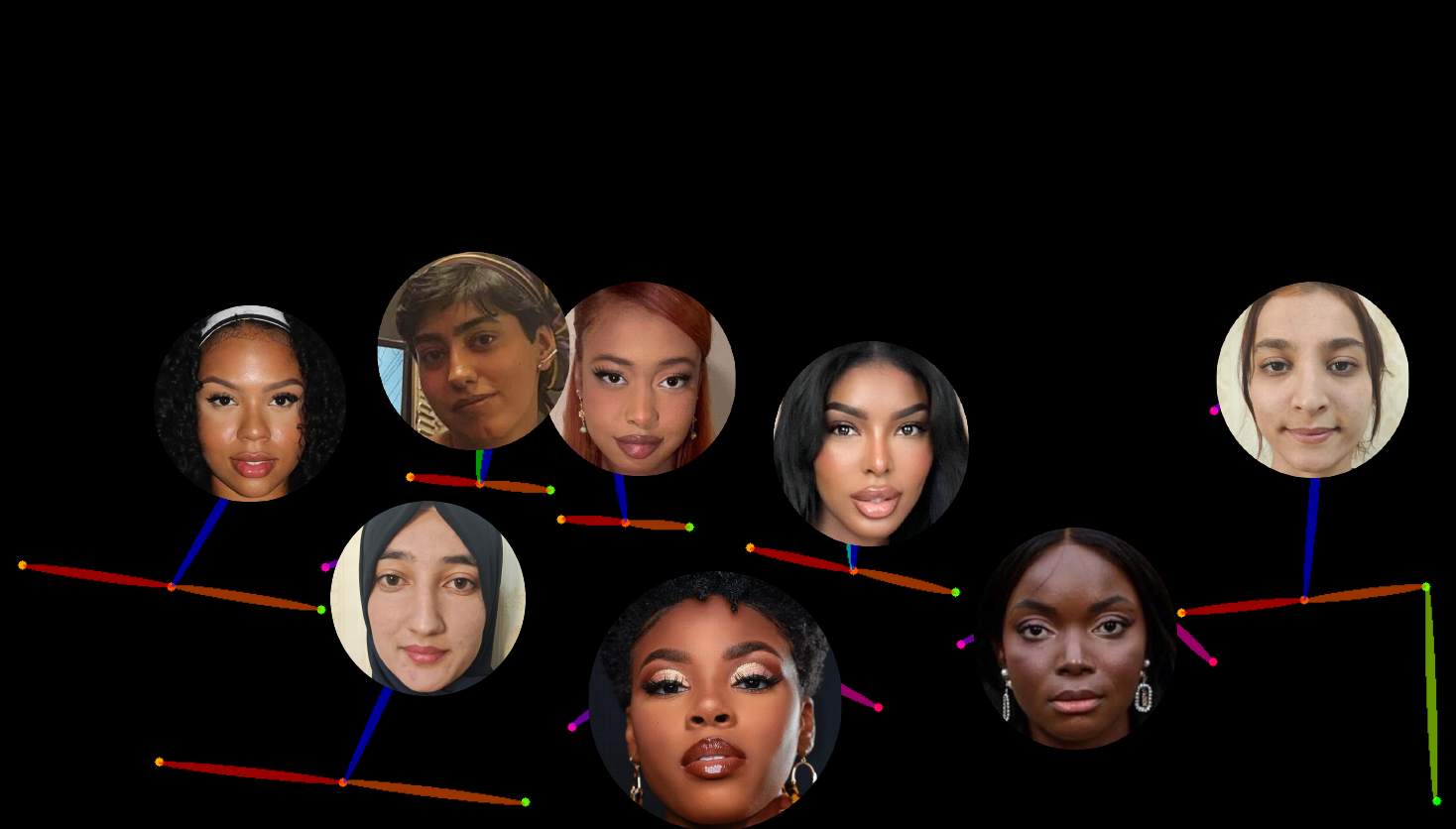}\end{minipage} 
    \vspace*{1mm} 
     &
    \begin{minipage}{0.18\textwidth}\centering\includegraphics[width=\linewidth]{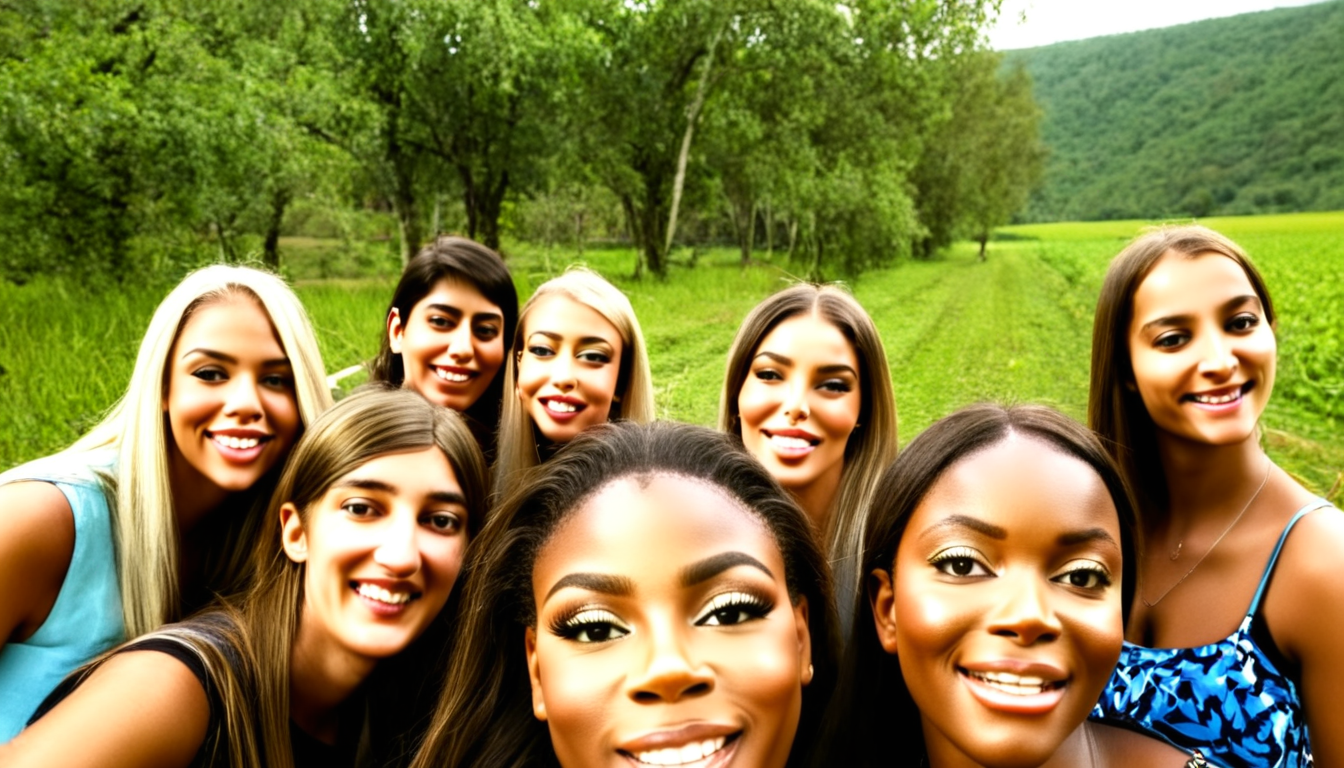}\end{minipage}
    &
    \begin{minipage}{0.18\textwidth}\centering\includegraphics[width=\linewidth]{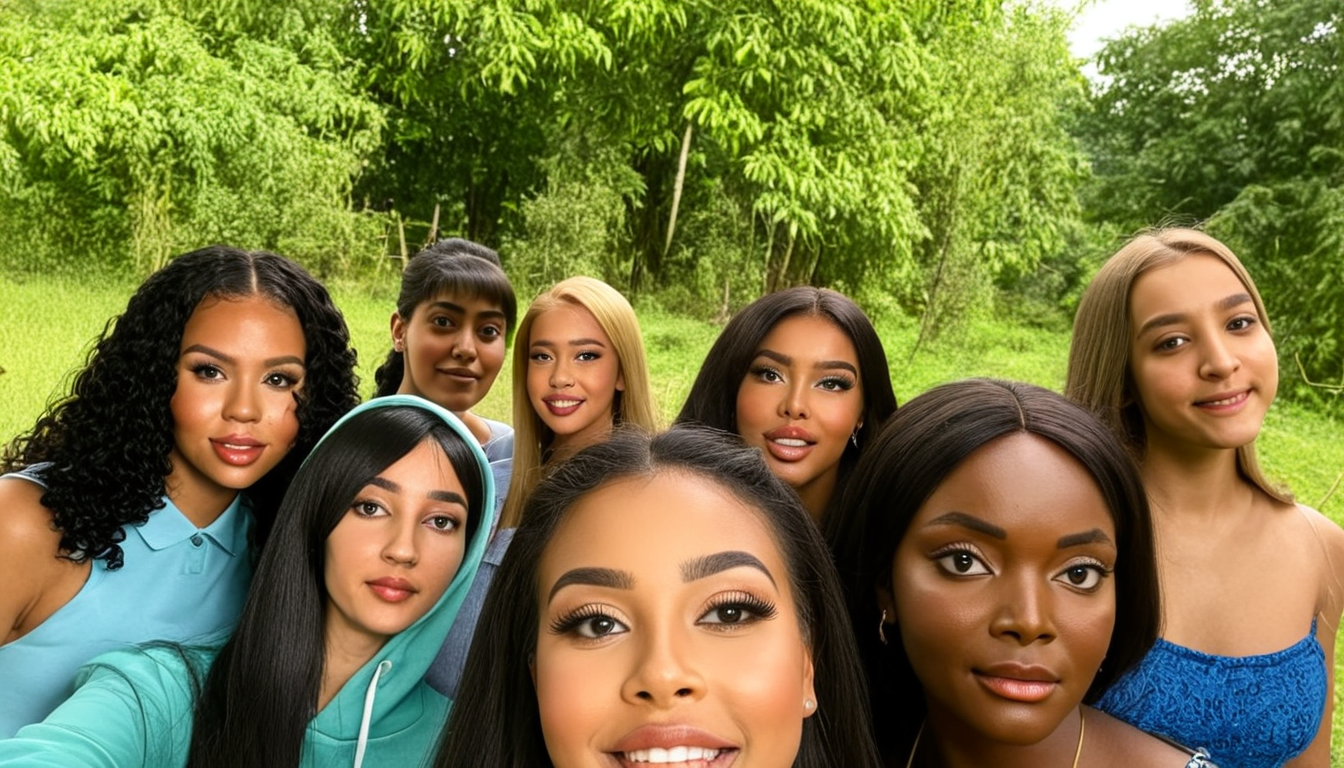}\end{minipage}
    &
    \begin{minipage}{0.18\textwidth}\centering\includegraphics[width=\linewidth]{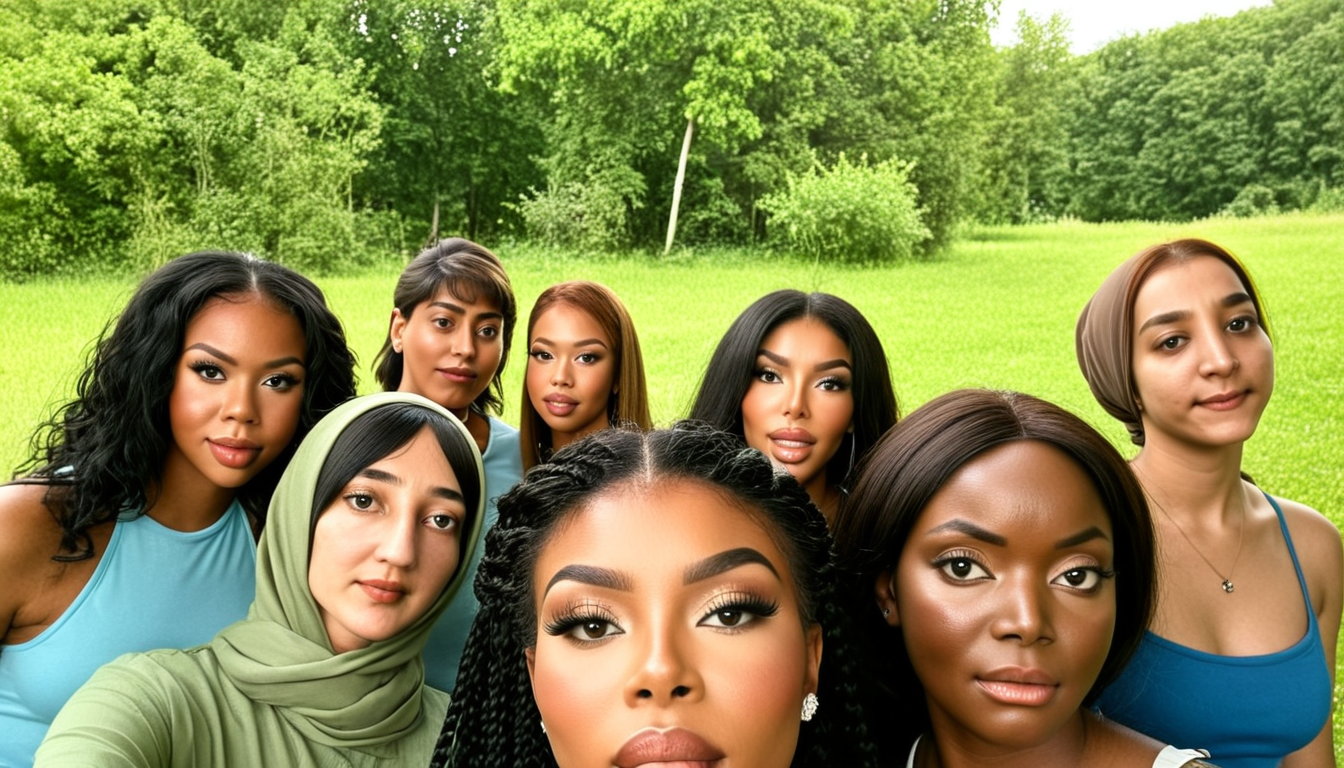}\end{minipage}
    \\
  \midrule
  \bottomrule[1pt]
\end{tabular} 

  }
  \caption{Part I: Additional comparison with baselines on pose-conditioned generation, where red dashed boxes highlight instances with low identity resemblance. 
  }
  \label{fig:add_compare_1}
  \vspace*{-3mm}
\end{figure*}

\begin{figure*}[t]
  \centering
  \resizebox{0.99\textwidth}{!}{
  \begin{tabular}{c|c|c|c}
  \toprule[1pt]
  \midrule
  \multicolumn{1}{c|}{\scriptsize{\textbf{ID + Pose}}}  & \multicolumn{1}{c|}{\scriptsize{\textbf{OMG}}}
 & \multicolumn{1}{c|}{\scriptsize{\textbf{InstantFamily}}}
 & \multicolumn{1}{c}{\scriptsize{\textbf{ID-Patch (Ours)}}}
    \\
\midrule

    \begin{minipage}{0.18\textwidth}\centering\includegraphics[width=\linewidth]{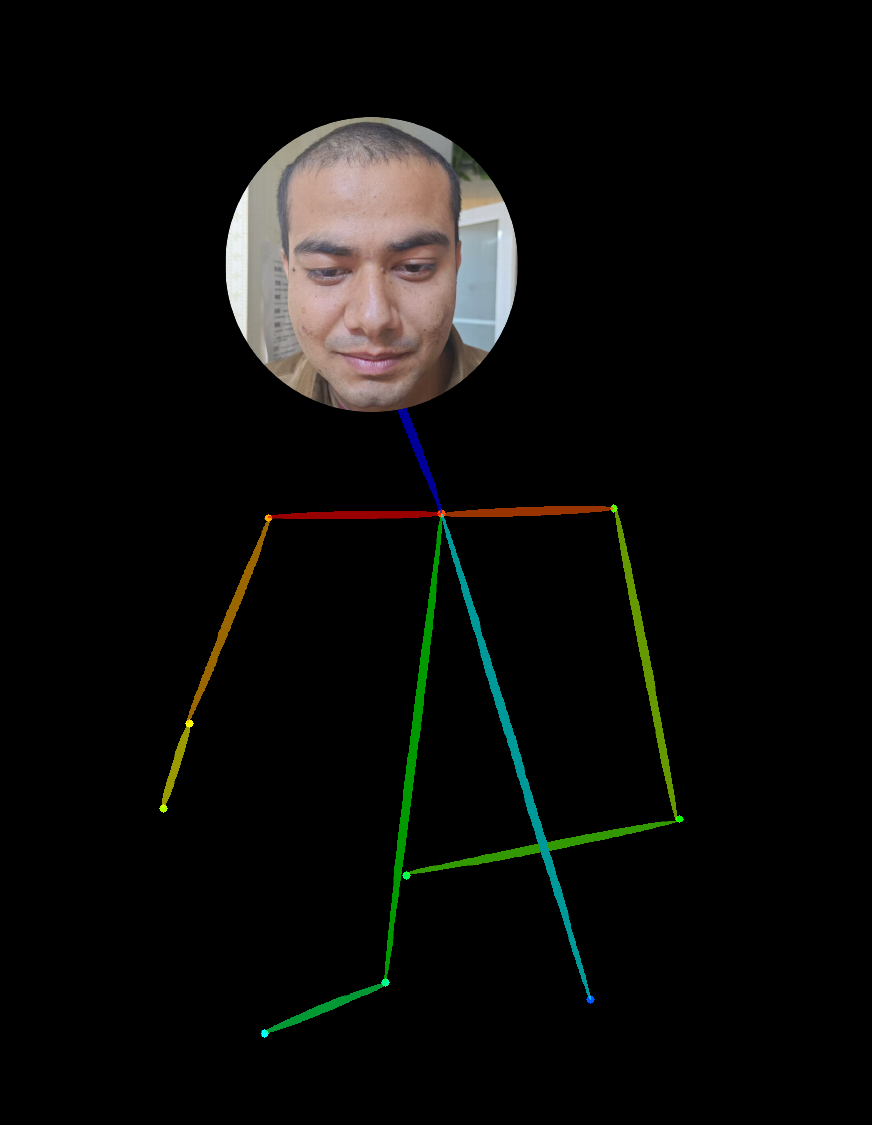}\end{minipage} 
    \vspace*{1mm} 
     &
    \begin{minipage}{0.18\textwidth}\centering\includegraphics[width=
    \linewidth]{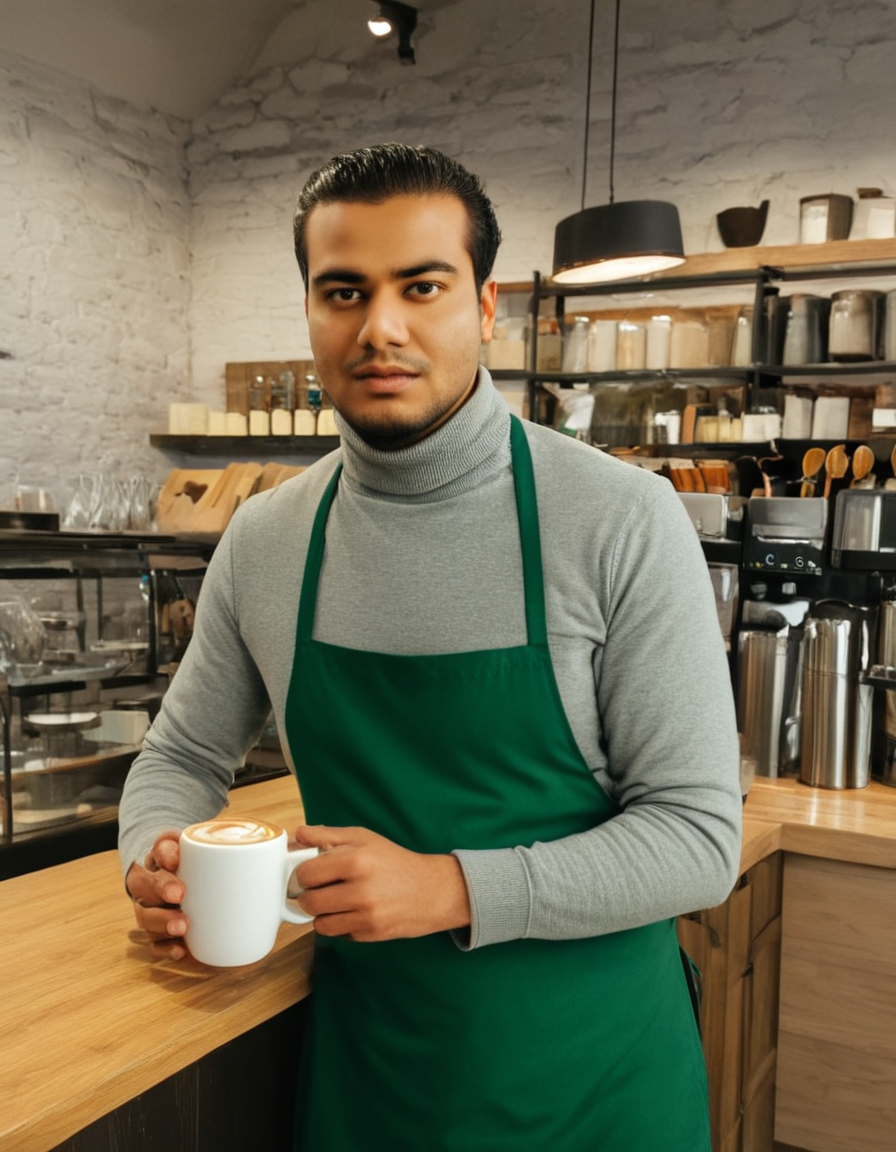}\end{minipage}
    &
    \begin{minipage}{0.18\textwidth}\centering\includegraphics[width=\linewidth]{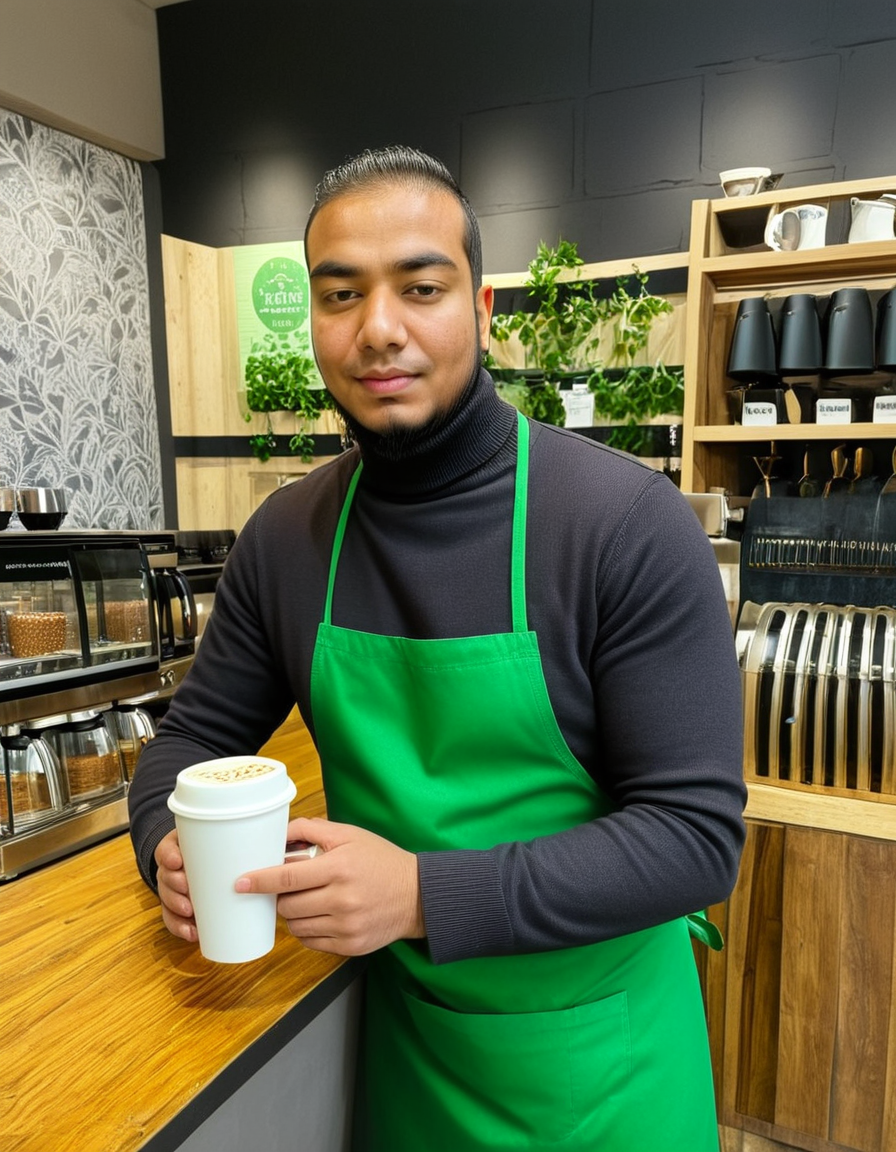}\end{minipage}
    &
    \begin{minipage}{0.18\textwidth}\centering\includegraphics[width=\linewidth]{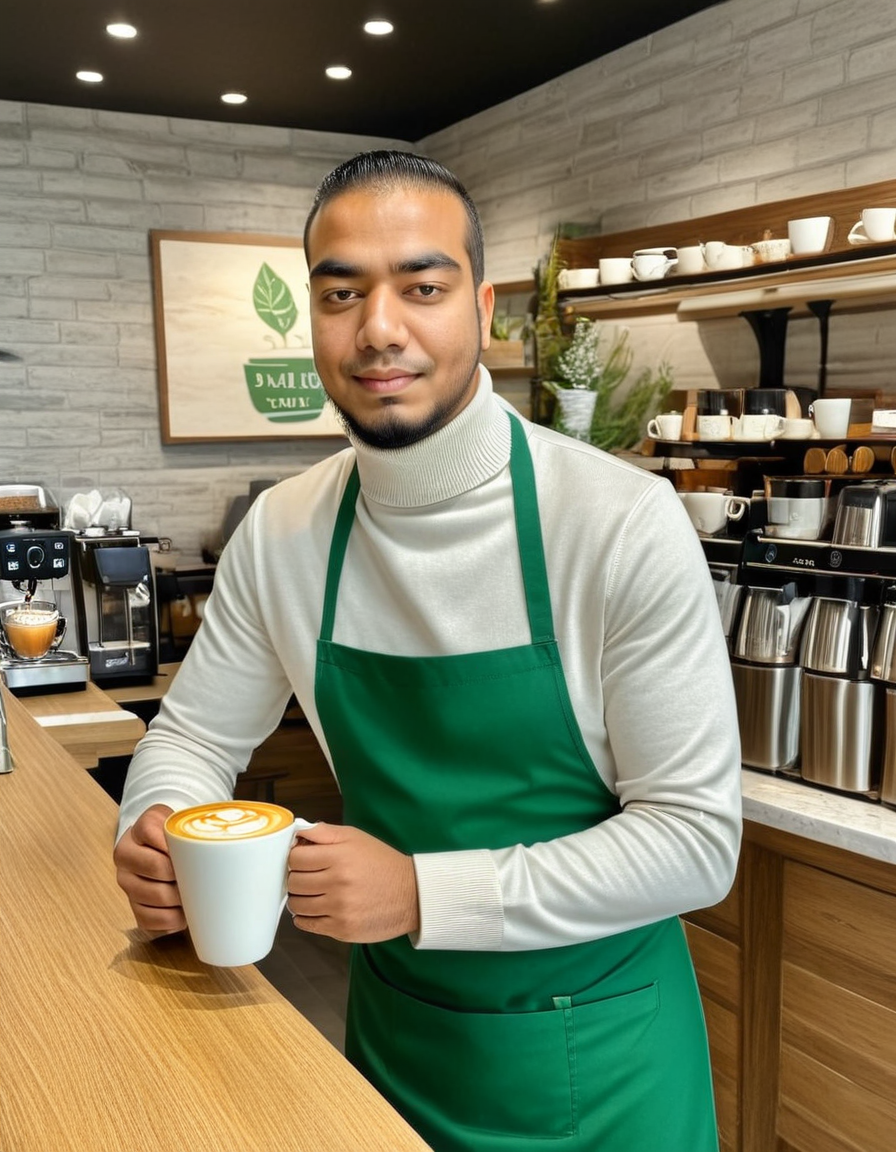}\end{minipage}
    \\
\midrule

    \begin{minipage}{0.18\textwidth}\centering\includegraphics[width=\linewidth]{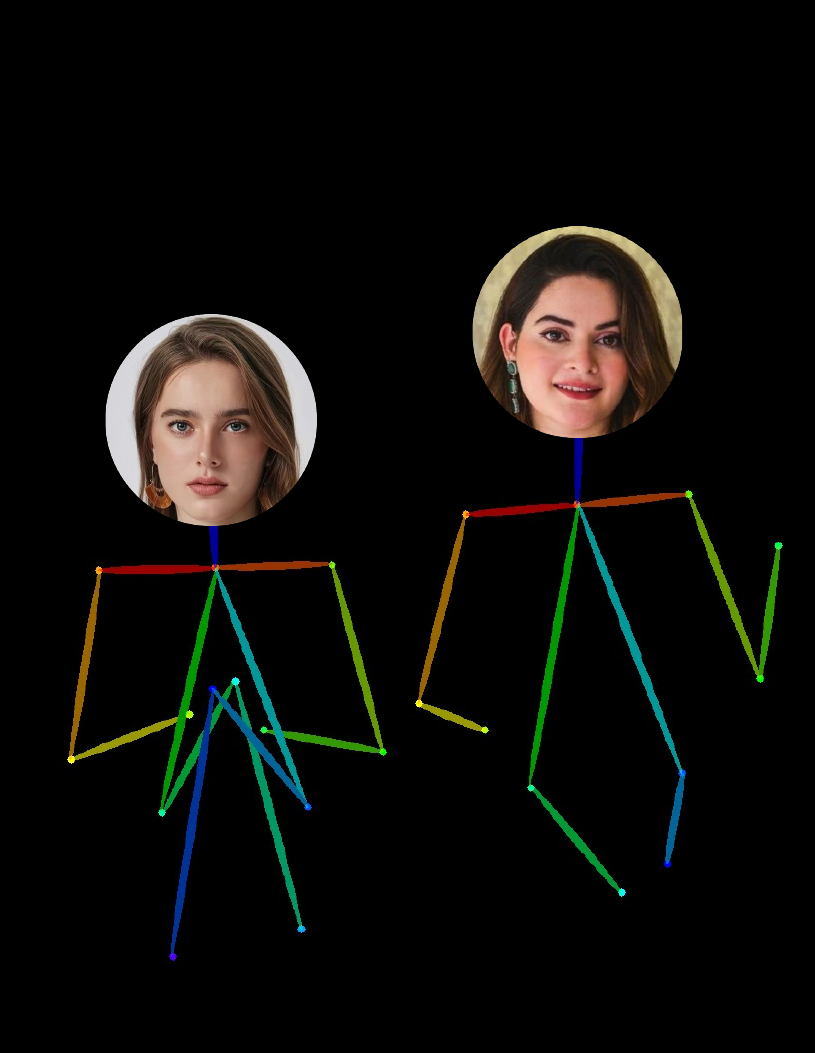}\end{minipage} 
    \vspace*{1mm} 
     &
    \begin{minipage}{0.18\textwidth}\centering\includegraphics[width=\linewidth]{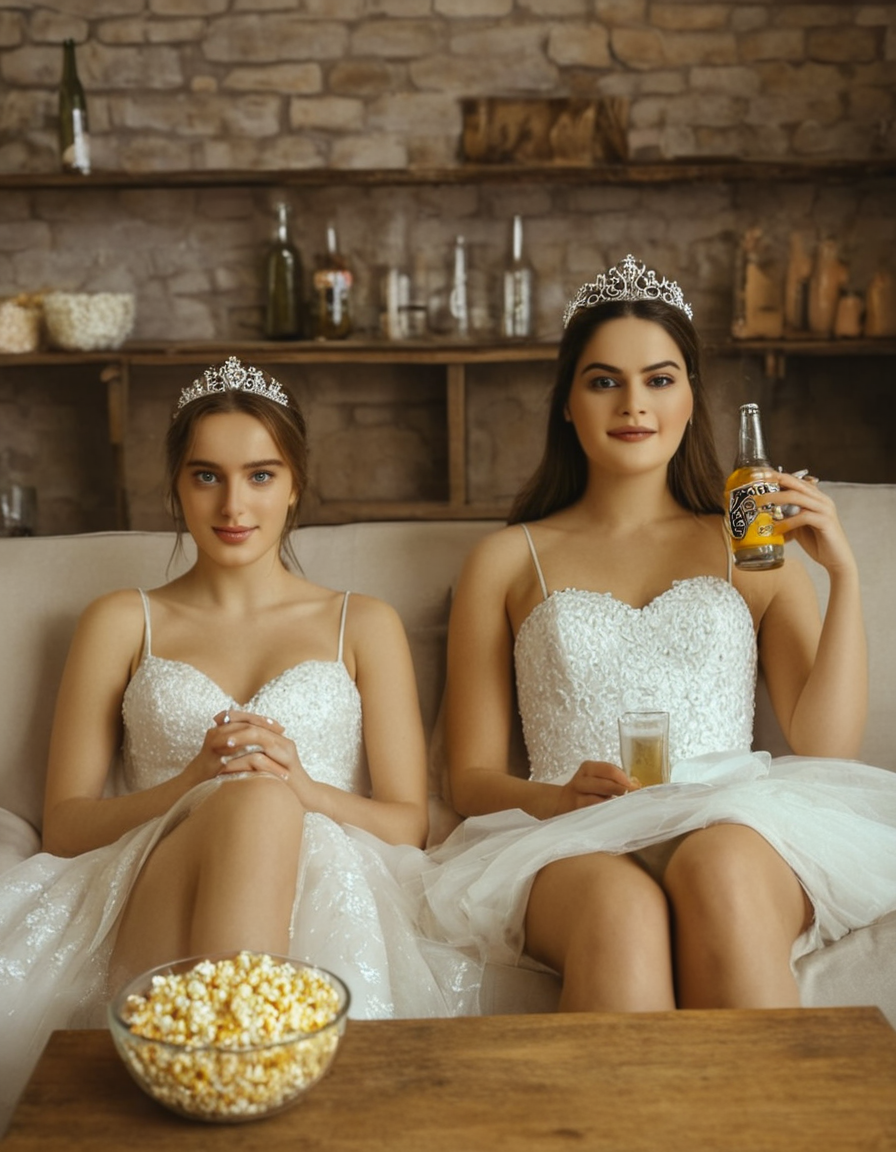}\end{minipage}
    &
    \begin{minipage}{0.18\textwidth}\centering\includegraphics[width=\linewidth]{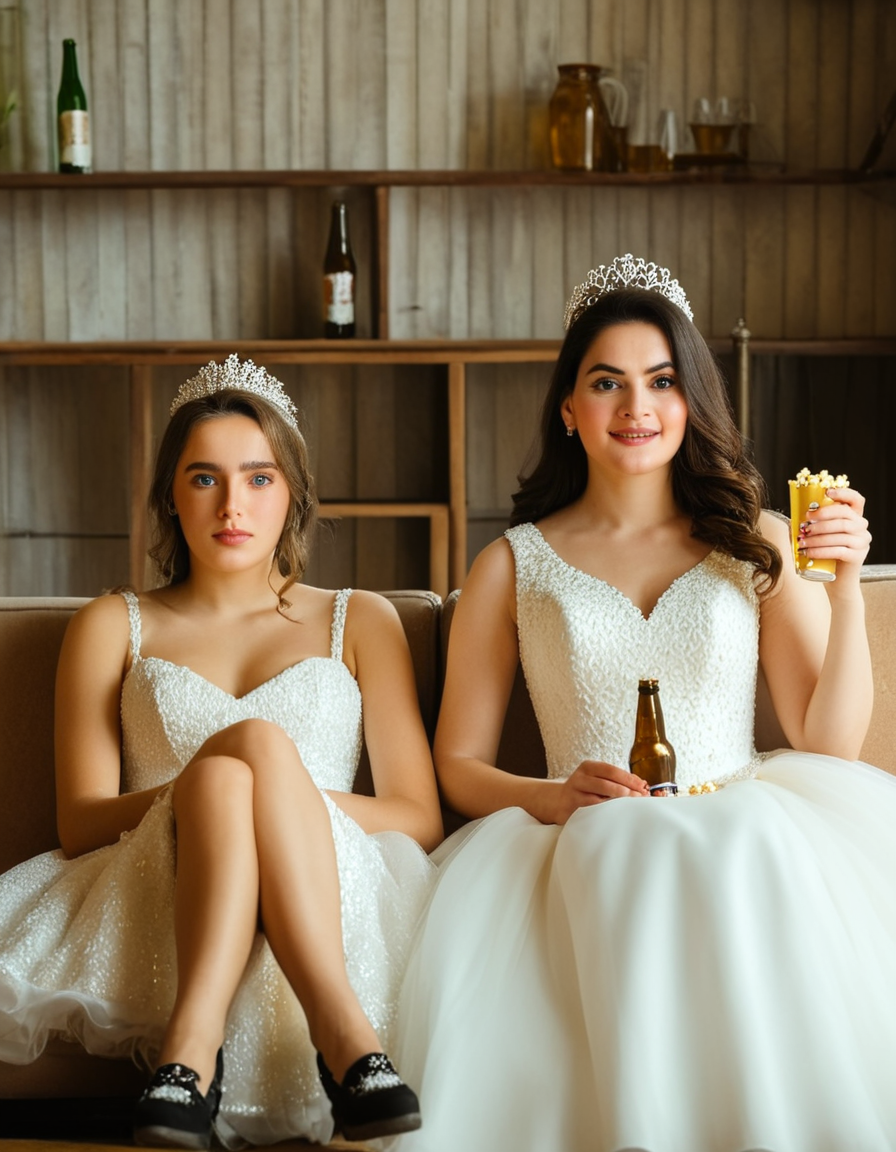}\end{minipage}
    &
    \begin{minipage}{0.18\textwidth}\centering\includegraphics[width=\linewidth]{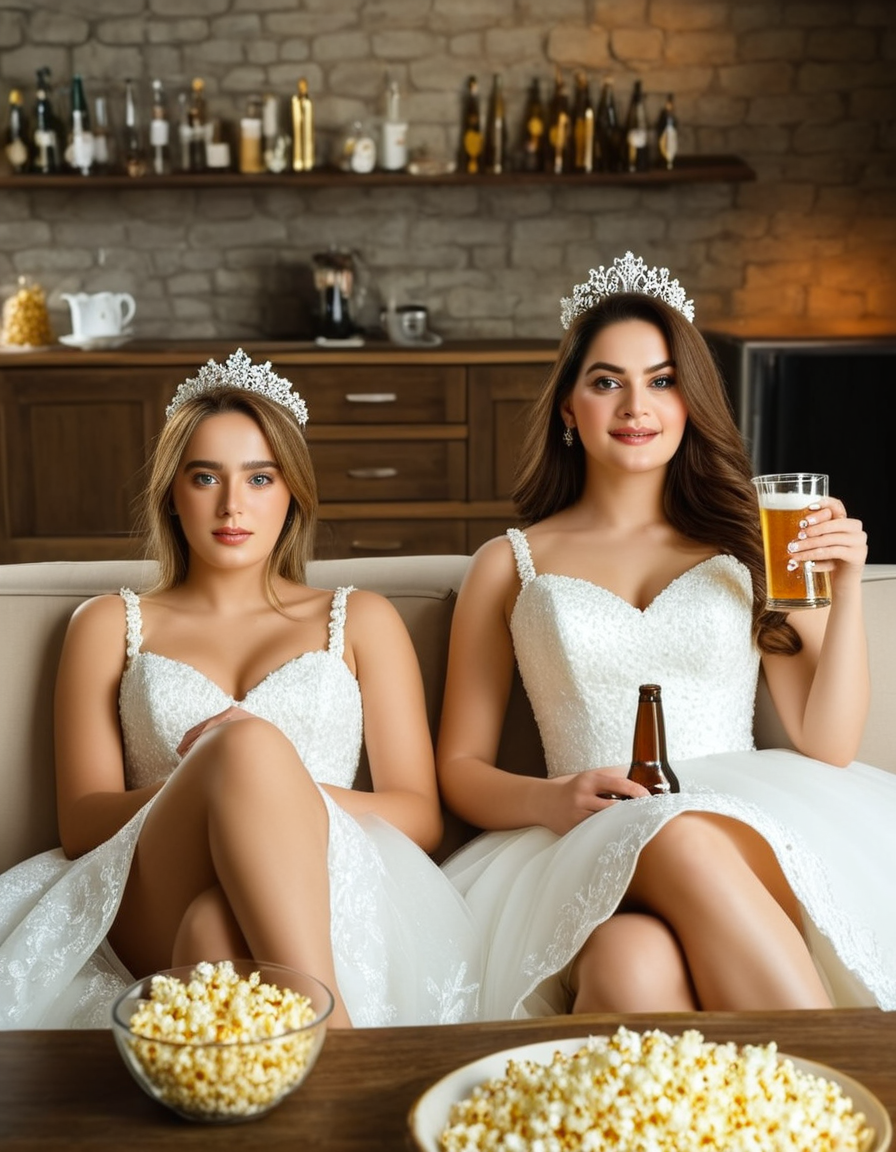}\end{minipage}
    \\

    \midrule

    \begin{minipage}{0.18\textwidth}\centering\includegraphics[width=\linewidth]{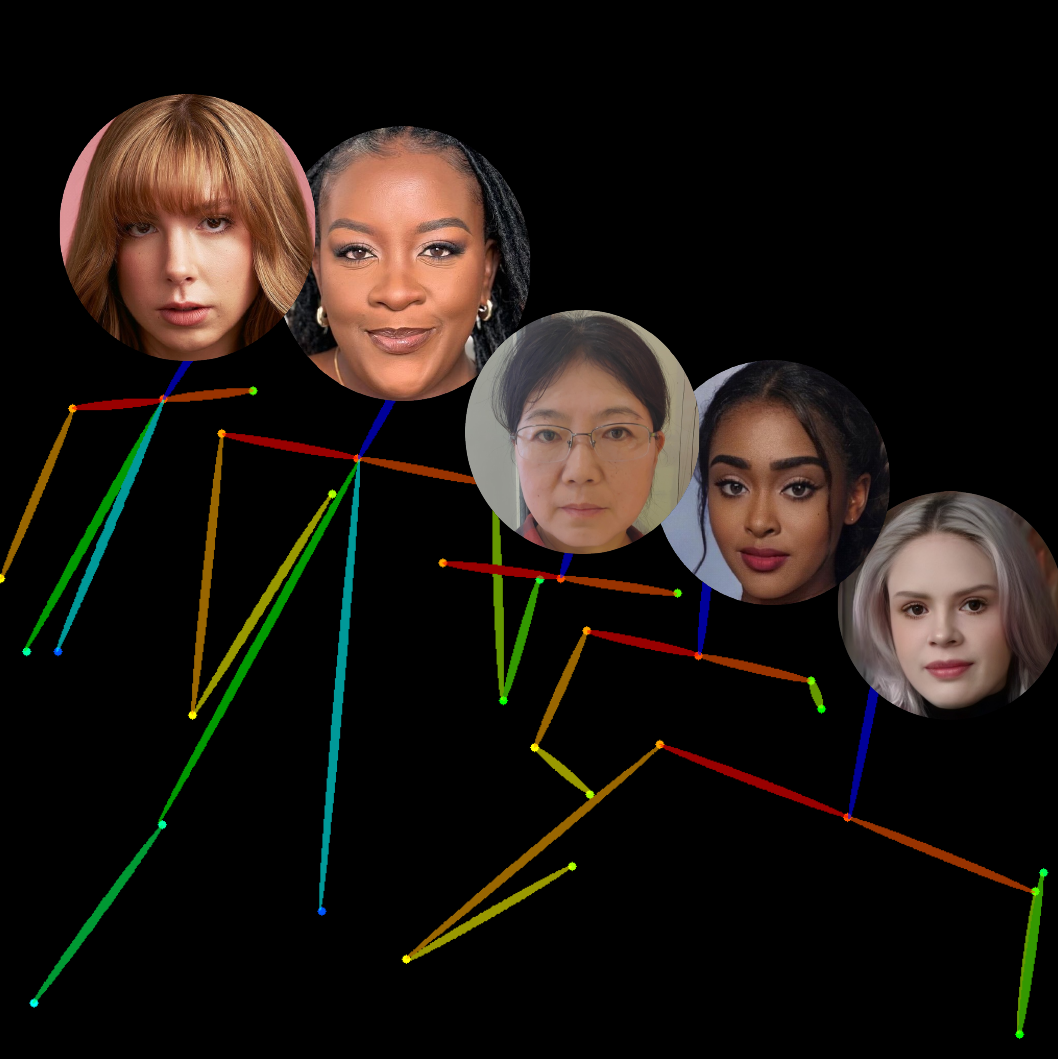}\end{minipage} 
    \vspace*{1mm} 
     &
    \begin{minipage}{0.18\textwidth}\centering\includegraphics[width=\linewidth]{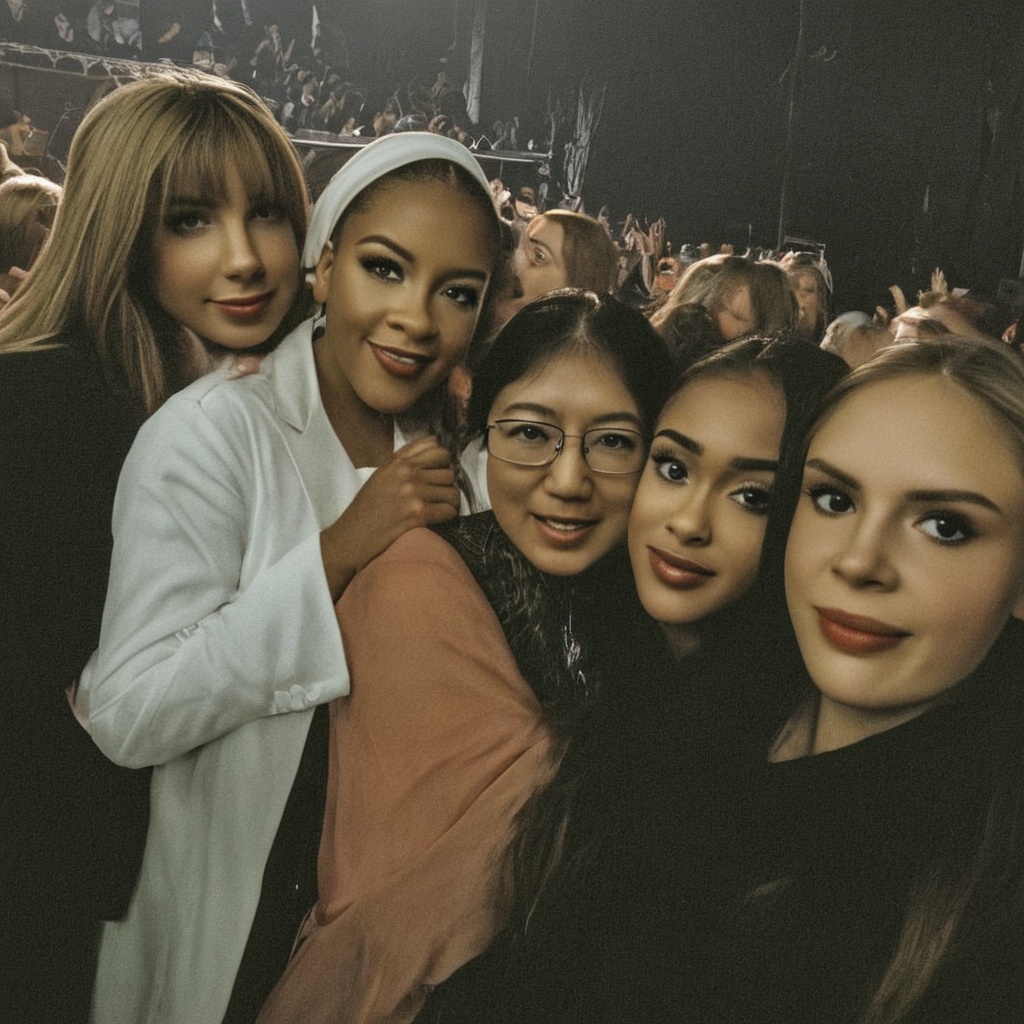}\end{minipage}
    &
    \begin{minipage}{0.18\textwidth}\centering\includegraphics[width=\linewidth]{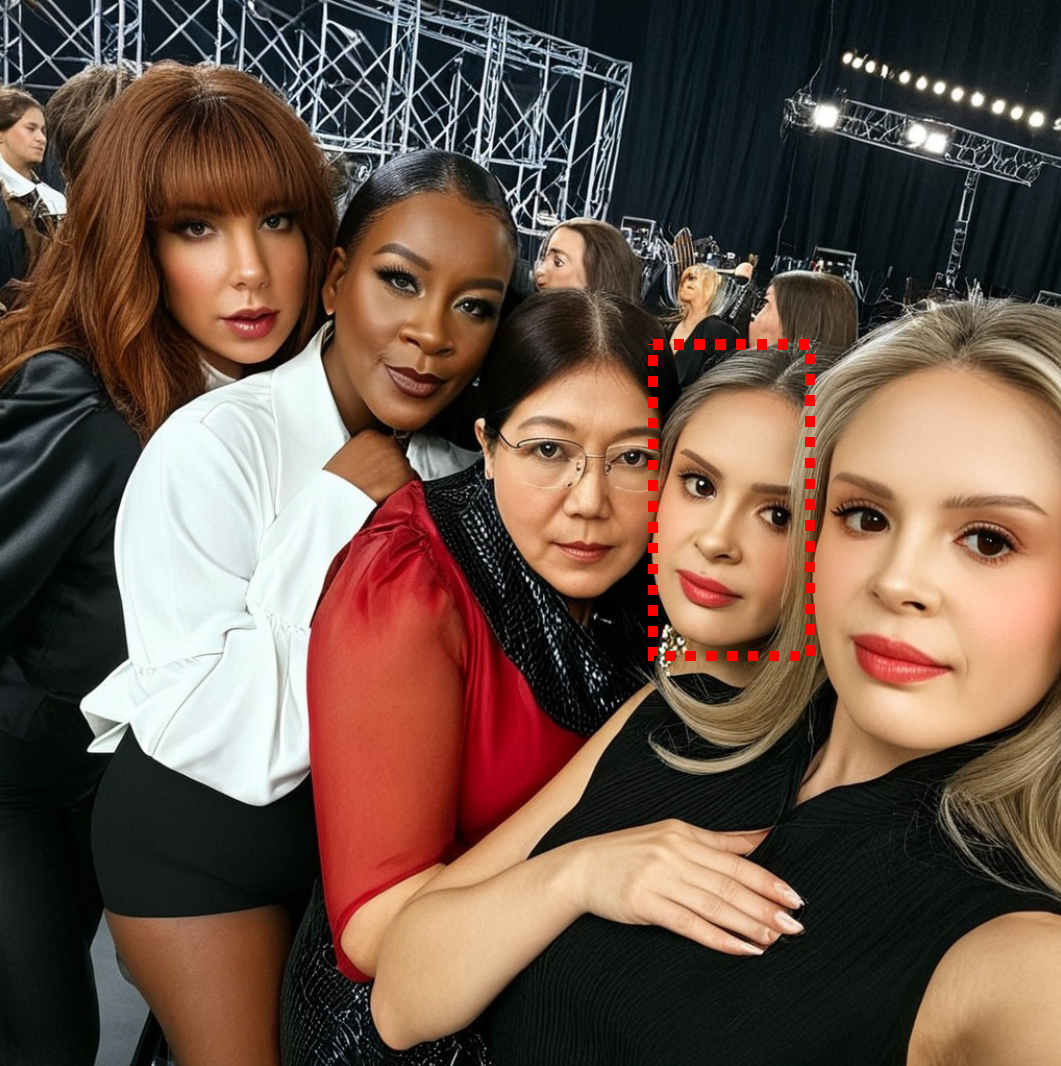}\end{minipage}
    &
    \begin{minipage}{0.18\textwidth}\centering\includegraphics[width=\linewidth]{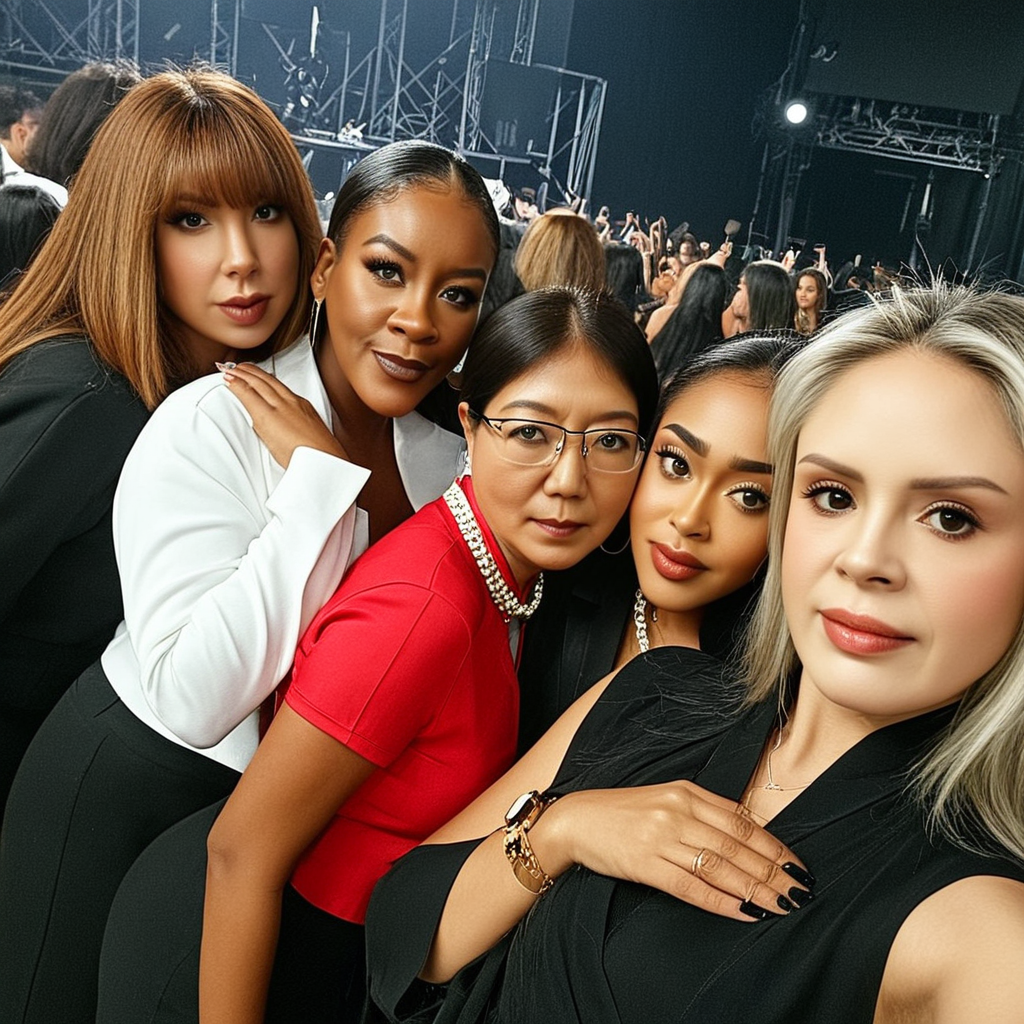}\end{minipage}
    \\
    \midrule

    \begin{minipage}{0.18\textwidth}\centering\includegraphics[width=\linewidth]{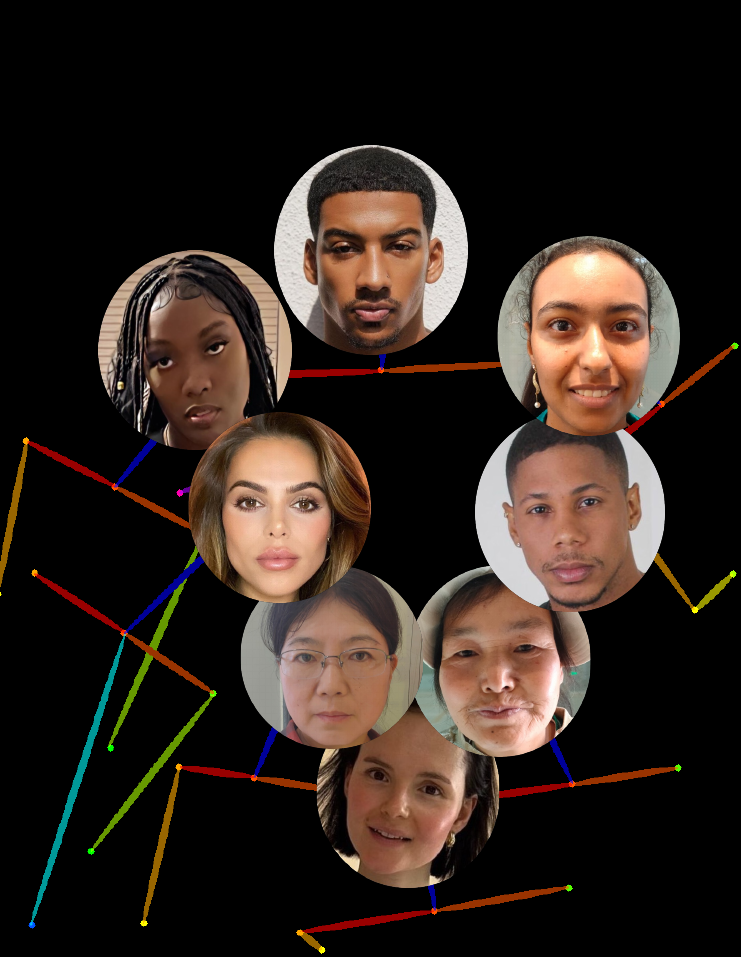}\end{minipage} 
    \vspace*{1mm} 
     &
    \begin{minipage}{0.18\textwidth}\centering\includegraphics[width=\linewidth]{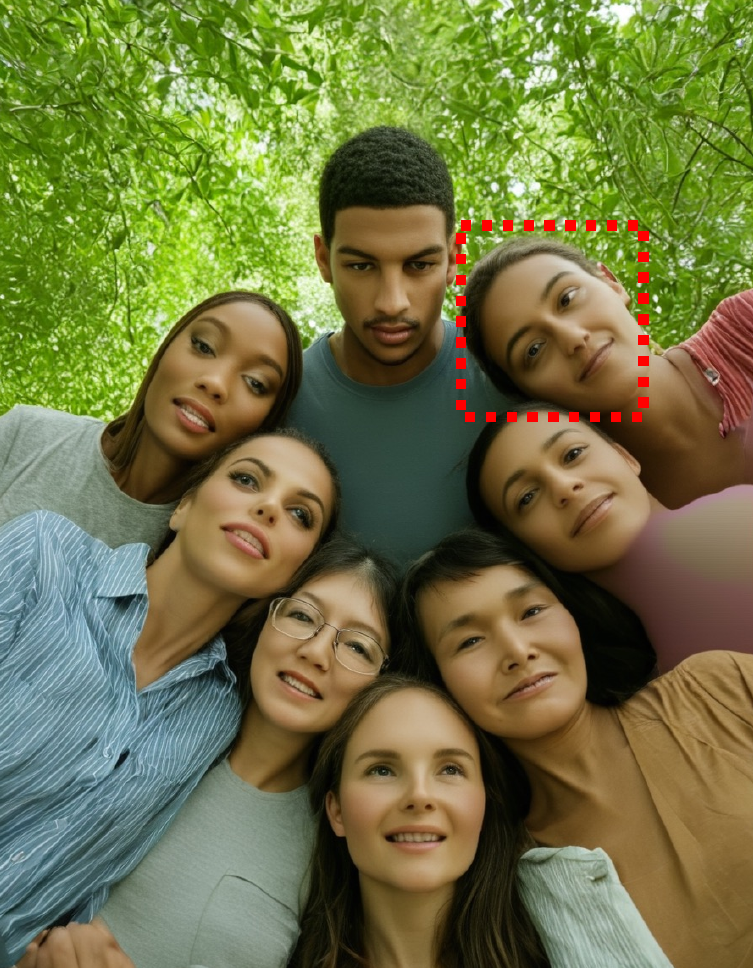}\end{minipage}
    &
    \begin{minipage}{0.18\textwidth}\centering\includegraphics[width=\linewidth]{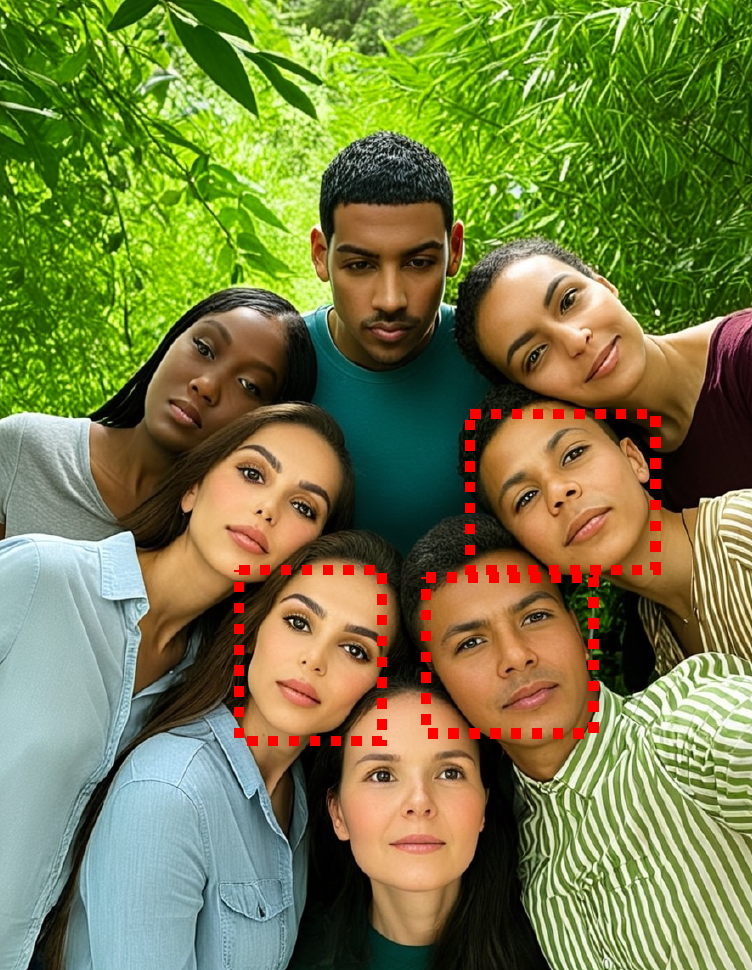}\end{minipage}
    &
    \begin{minipage}{0.18\textwidth}\centering\includegraphics[width=\linewidth]{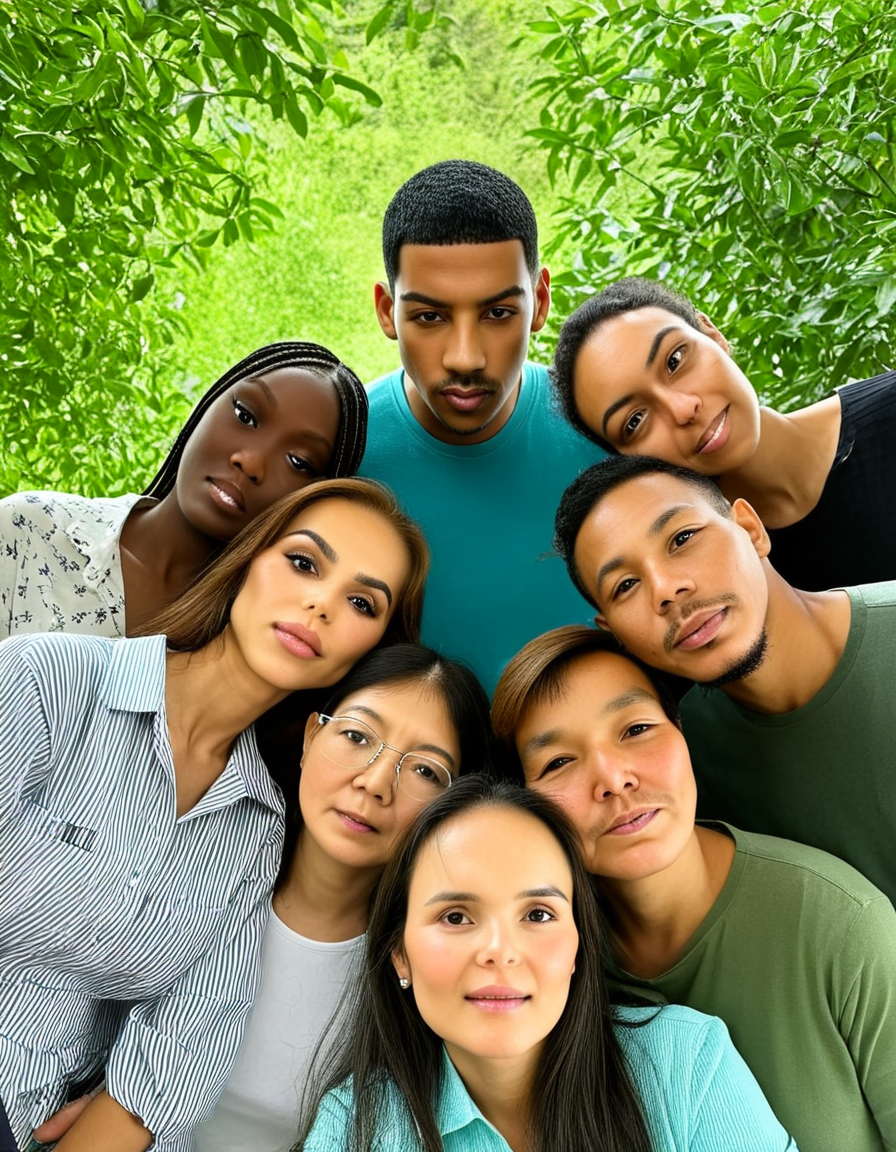}\end{minipage}
    \\
  \midrule
  \bottomrule[1pt]
\end{tabular} 

  }
  \caption{Part II: Additional comparison with baselines on pose-conditioned generation, where red dashed boxes highlight instances with low identity resemblance. 
  }
  \label{fig:add_compare_2}
  \vspace*{-3mm}
\end{figure*}

Additional comparisons of pose-conditioned generated images are available in \textbf{Fig.\,\ref{fig:add_compare_1}} and \textbf{Fig.\,\ref{fig:add_compare_2}}. In these visualizations, our proposed method, ID-Patch, consistently achieves robust ID association without ID leakage. In contrast, the other two methods experience significant ID leakage as the total number of faces in the generated images increases.

\end{document}